\documentclass[sn-mathphys-num]{sn-jnl}

\usepackage{color}
\usepackage{latexsym}
\usepackage{graphicx}
\usepackage{caption}
\usepackage{subcaption}
\usepackage{enumitem}
\usepackage{bbm}
\usepackage{mathrsfs}
\usepackage{amsmath, amssymb}  
\usepackage{mathtools}         
\usepackage{booktabs}          
\usepackage{amsfonts}

\usepackage{amsmath}
\numberwithin{equation}{section}
\usepackage{amsthm}
\usepackage{amssymb}

\usepackage[dvipsnames]{xcolor}
\newcommand{\redblock}[1]{{\color{black}#1}}

\definecolor{darkblue}{rgb}{0.0,0,0.5}
\usepackage{algorithm,algorithmic}
\usepackage{caption}

\newcommand{\R}{\mathbb{R}}
\newcommand{\E}{\mathbb{E}}

\renewcommand{\H}{\mathcal{H}}

\newcommand{\kernel}{K}

\newcommand{\norm}[1]{\left\|{#1}\right\|} 

\newcommand{\normal}{\mathsf{N}}  

\renewcommand{\P}{\mathbb{P}}

\newtheorem{lemma}{Lemma}
\newtheorem{corollary}{Corollary}
\newtheorem{assumption}{Assumption}[section]
\newtheorem{innercustomgeneric}{\customgenericname}
\providecommand{\customgenericname}{}

\renewcommand{\H}{\mathbb{H}}

\newcommand{\Cov}{{\rm Cov}}
\newcommand{\Var}{{\rm Var}}

\renewcommand{\H}{\mathcal{H}}

\newcommand{\eps}{\epsilon}

\newcommand{\half}{\frac{1}{2}}

\newcommand{\profile}{\psi}

\newcommand{\C}{\mathbb{C}} 
\newcommand{\supp}{\mathop{\rm supp}}

\theoremstyle{thmstyleone}%
\newtheorem{theorem}{Theorem}
\newtheorem{proposition}[theorem]{Proposition}%

\theoremstyle{thmstyletwo}%
\newtheorem{example}{Example}%
\newtheorem{remark}{Remark}%

\theoremstyle{thmstylethree}%
\newtheorem{definition}{Definition}%

\begin{document}

\title[A Compositional Kernel Model for Feature Learning]{A Compositional Kernel Model for Feature Learning}


\author*{\fnm{Feng} \sur{Ruan}}\email{fengruan@northwestern.edu}
\author*{\fnm{Keli} \sur{Liu}}\email{keli.liu25@gmail.com}
\author*{\fnm{Michael} \sur{Jordan}}\email{jordan@cs.berkeley.edu}

\affil*{\orgdiv{Department of Statistics and Data Science}, \orgname{Northwestern University}}
\affil*{\orgname{Company E}}
\affil*{\orgdiv{Department of Statistics and Computer Science}, \orgname{University of California, Berkeley}}


\abstract{
\phantom{.....}
We study a compositional variant of kernel ridge regression in which the predictor is applied to a coordinate-wise reweighting of the inputs. Formulated as a variational problem, this model provides a tractable setting for studying feature learning in compositional architectures. From the perspective of variable selection, we show how relevant variables are recovered while noise variables are eliminated.  We prove that both global minimizers and stationary points discard noise coordinates when the noise variables are Gaussian distributed. A central finding is that $\ell_1$-type kernels, such as the Laplace kernel, succeed in recovering features contributing to nonlinear effects at stationary points, whereas Gaussian kernels recover only linear ones.
}

\maketitle

\section{Introduction}
Deep learning has achieved remarkable success across domains such as vision, language, and science. A widely believed explanation for this success is representation learning --- also called feature learning --- the empirically observed ability of deep models to automatically extract task-relevant features from raw data, \emph{without manual engineering}, to support downstream prediction~\cite{BengioCoVi13}. 

This ability is generally attributed to two fundamental ingredients of deep models: (i) their \emph{compositional architecture} and (ii) the use of \emph{optimization}. The \emph{compositionality} of the architecture endows the model with the ability to form intermediate representations of the data via composition of simple transformations. These representations are not manually defined but are learned from data by \emph{optimizing} a loss function designed to minimize prediction error.

However, despite the empirical success of this paradigm, our theoretical understanding of how and why such representations emerge remains fundamentally limited. In particular, it remains unclear how the interplay between \emph{compositional structure} and \emph{optimization} gives rise to task-aligned features --- and under what conditions this mechanism succeeds or fails. To address this gap, we study a stylized compositional model that preserves these two core ingredients of feature learning --- while remaining simple enough to enable analysis of how features are learnt during training.

Our starting point is the classical kernel ridge regression (KRR) framework, which aims to learn a nonlinear function $f$ that predicts the response $Y \in \R$ from $X \in \R^d$, where $(X, Y) \sim \P$~\cite{CuckerSm02}. The function $f$ is chosen from a reproducing kernel Hilbert space $H$ (e.g., Sobelev space). In KRR, the learning problem is formulated as the solution to the variational problem
\begin{equation*}
	\min_f \E[(Y- f(X))^2] + \lambda \norm{f}_H^2.
\end{equation*} 
Here, $\norm{f}_H$ denotes the RKHS norm. The expectation term measures the accuracy of approximation, while the regularization term $\lambda \norm{f}_H^2$ imposes a certain amount of smoothness,  which helps to achieve good 
generalization to unseen test data.

To incorporate feature learning into this framework, we introduce a \emph{compositional variant} of KRR, in which the function $f \in H$ is applied to a learned coordinate-wise reweighting of the input
\begin{equation}
\label{eqn:model-f-beta}
	\min_{\beta} \min_f \mathbb{E}\left[(Y - f(\beta \circ X))^2\right] + \lambda \|f\|_H^2.
\end{equation}
Here, $\beta \in \R^d$ defines a learnable input transformation (coordinate-wise reweighting), where $\beta \circ X \in \R^d$ denotes the Hadamard product, defined by $(\beta \circ X)_j = \beta_j X_j$ for $j = 1, 2, \ldots, d$. In this setup, \emph{feature learning} corresponds to learning $\beta$: the model adaptively up-weights and down-weights coordinates before applying the predictor $f$. 

Building on this formulation, we use the model as a simple testbed to study feature learning. In this paper, we formalize this question through the perspective of \emph{variable selection}: 
\begin{quote}
	If $Y$ depends only on a few coordinates of $X$, can the compositional model~\eqref{eqn:model-f-beta} recover these relevant features and eliminate irrelevant ones through joint optimization of $f$ and $\beta$?
\end{quote} 
Formally, we assume $Y$ depends on $X$ only through a subset of coordinates
$X_{S_*}:= (X_j)_{j \in {S_*}}$, i.e., 
\begin{equation}
\label{eqn:sparse-setup}
	\E[Y|X] = \E[Y|X_{S_*}]
\end{equation}
 for some unknown index subset $S_* \subset \{1, 2, \ldots, d\}$. For identifiability, 
we define $S_*$ as the \emph{minimal} subset of coordinates of $X$ that suffices to predict $Y$ 
(see Definition~\ref{definition:minimal-sufficient-feature-set}). 
From the perspective of \emph{variable selection}, a central goal of feature learning is to discard as 
many irrelevant features in $X_{S_*^c}$ as possible---and a more ambitious one---recovering the exact 
support, so that $\operatorname{supp}(\beta)$ matches $S_*$, the index of the true relevant features $X_{S_*}$. 


\subsection{Overview of Main Results} 
\label{sec:overview-of-main-results}
Feature learning in this model is governed by the double minimization in~\eqref{eqn:model-f-beta}. 
Since for each fixed $\beta$, the inner minimization reduces to a standard KRR solution with weighted 
input $\beta \circ X$, this leads us to the reduced objective 
\begin{equation}
\mathcal{J}(\beta,\lambda) := \min_{f \in H} \E\!\left[(Y - f(\beta \circ X))^2\right] + \lambda \|f\|_H^2.
\end{equation}
The learning problem thus reduces to minimizing $\mathcal{J}(\beta,\lambda)$ over $\beta$. 
From the perspective of \emph{variable selection}, two fundamental questions drive our analysis:

\vspace{.5em} 
\begin{itemize}
\item[$\mathsf{(P1)}$] Under what conditions does minimizing $\mathcal{J}(\beta, \lambda)$ yield a sparse $\beta$ that discards all the noise variables, i.e., $\supp(\beta) \subset S_*$?
\item[$\mathsf{(P2)}$] More stringently, when does $\beta$ exactly recover all relevant features, i.e., $\operatorname{supp}(\beta) = S_*$?
\end{itemize} 
\vspace{.5em}


Answering these questions is far from straightforward: the objective $\mathcal{J}$ is nonconvex in $\beta$ and depends on it nonlinearly. This motivates our study of the \emph{landscape} of $\mathcal{J}$, including its first variation, global minimizers, and stationary points. We analyze these objects under various data distributions $(X,Y)\sim\P$ and translation-invariant RKHS $H$, with particular focus on when the landscape promotes variable selection as a form of feature learning. Methodologically, we use tools from probability, Fourier analysis, and the calculus of variations to uncover how the geometry of $\mathcal{J}$ reflects the interplay between $\P$ and $H$, and when this variational objective identifies predictive features.


Our results are purely \emph{static} in nature: we characterize the landscape of $\mathcal{J}$ and the statistical properties of its global minimizers and critical points, which serves as a foundation for studying feature learning in $\mathcal{J}$. Questions about the \emph{dynamics} of optimization algorithms, or about \emph{finite sample effects} arising from replacing $\E$ by its empirical counterpart $\E_{{\rm emp}}$, are clearly important in practice. However, these investigations should build on the present foundation and are best pursued as separate lines of work beyond the scope of this paper.


With the goal of characterizing the landscape of $\mathcal{J}$ in mind, we now summarize our main results, showing how $\mathcal{J}$ induces variable selection at its stationary points and global minimizers. 

\redblock{
\vspace{.5em}
\begin{itemize}
\item (\emph{Global Minimizer Recovers Features Perfectly},~Section~\ref{sec:global-minimizers}). 
Assuming (a) the relevant features $X_{S_*}$ and irrelevant ones $X_{S_*^c}$ are statistically independent, (b) $X$ has nontrivial predictive power of $Y$ ($\E[Y|X] \neq 0$), and (c) the noise features have nonzero variance ($\Var(X_j) \neq 0$ for $j \notin S_*$), then for small $\lambda$, every global minimizer of $\mathcal{J}$ exactly recovers the true support $S_*$ (Theorem~\ref{theorem:global-minimizer-three}).  

The independence condition (a) avoids statistical confounding from correlations between relevant and irrelevant features, and is comparable to those in the nonparametric feature selection literature~\cite{LaffertyWasserman08, CommingesDa12, CandesFaJaLv18}. The condition (b) $\E[Y|X] \neq 0$ is necessary: otherwise, $\mathcal{J}(\beta, \lambda) \equiv \E[Y^2]$ for all $\beta$, and there is no feature selection. The condition (c) is mild but necessary: if $\Var(X_i) = 0$ for some $i \in S_*^c$, then $\mathcal{J}$ becomes constant with respect to that noise feature $i$, making it impossible to eliminate it. Finally, the condition that $\lambda$ is small is necessary: for large $\lambda$, noise features are still eliminated at every global minimizer (Theorem~\ref{theorem:global-minimizer}), but relevant features may also be eliminated if the regularization term $\lambda \norm{f}_H^2$ dominates.
\end{itemize}
}
\vspace{.5em}
However, the objective $\mathcal{J}$ is nonconvex, and global minimization is rarely tractable in nonconvex settings.  This shifts 
our focus to understanding the behavior of local minima and stationary points.

\redblock{
\vspace{.5em}
\begin{itemize}
\item (\emph{Stationary Points Eliminate Noise Variables},~Section~\ref{sec:stationary-points-and-irrelevant-feature-elimination}).
A striking feature of the objective $\mathcal{J}$ is its ability to filter out irrelevant variables---even at \emph{stationary points}, not just global minima. Under assumptions (a) and (c) above, we show that if the irrelevant features follow a Gaussian distribution (a canonical distribution for modeling noise), then any stationary point $\beta_\dagger$ with nontrivial predictive power---formally, $\E[Y|\beta_\dagger \circ X] \neq 0$---must assign zero weight to all irrelevant coordinates (Corollary~\ref{corollary:elimination-at-stationary-points}). This denoising effect holds regardless of the distribution of relevant features and holds for all $\lambda > 0$. 
\end{itemize}
}

\vspace{.3em}
These results ensure that stationary points with nontrivial prediction is automatically noise-free---but can they also recover the  relevant features? We provide some answers to this in Section~\ref{sec:stationary-points-and-relevant-feature-recovery}. 

\vspace{.3em}
\begin{itemize}
\item (\emph{Advantage of Laplace over Gaussian Kernels})
A key theme in our analysis is that recovery of relevant features at stationary points depends critically on the choice of $H$. We say a feature $X_i$ has a main effect if $\mathbb{E}[Y | X_i] \ne \E[Y]$. Among common kernels, those based on the $\ell_1$ norm---such as the Laplace kernel---enjoy a distinctive advantage: they are sensitive to main effects of any form. 
As an example, the point $\beta = 0$ can never be stationary if any feature exhibits a main effect. In contrast, Gaussian kernels respond only to linear effects: if the features are nonlinearly predictive but linearly uncorrelated with $Y$, then $\beta=0$ is stationary despite $\E[Y|X_i]\ne \E[Y]$.

\vspace{.3em}
\item (\emph{Laplace Kernel Detects All Main Effects})
This advantage of Laplace kernels extends beyond the origin $\beta = 0$. Assuming mutual independence between the relevant features, we show that within any bounded region, every stationary point $\beta_\dagger$ must assign nonzero weights to features that carry a main effect for small enough $\lambda$ (Corollary~\ref{corollary:extraction-of-main-effect}). The independence assumption ensures that variation in $\mathbb{E}[Y | X]$ can be attributed to individual features $X_i$ in an unambiguous way. In short, $\ell_1$-type kernels enable local optima---not just global optima---to recover all relevant features that exhibit main effects. 

\vspace{.3em}
\item (\emph{Recovering Interactions Requires Coordination})
In contrast, recovering features involved only in interaction effects requires coordination among variables. Assume mutual independence between the relevant features. We say that a feature $X_i$ participates in an interaction if $\E[Y|X_{S \backslash \{i\}}] \ne \E[Y|X_S]$ for some index set $S \subset \{1, 2, \ldots, d\}$ where $i \in S$. In words, $X_i$ contributes information about $Y$ beyond what is already contained in the other variables of $S$.  We further assume that $X_i$ has no main effect, meaning $\mathbb{E}[Y | X_i] = \mathbb{E}[Y]$. 

If all interaction partners are inactive (i.e., $\beta_j = 0$ for all $j \in S \backslash \{i\}$), then the partial derivative with respect to $\beta_i$ can vanish---implying $X_i$ is undetectable in isolation. This occurs, for example, at $\beta = 0$, which is a stationary point under such conditions. However, once the partner features $(X_j)_{j \in S \backslash \{i\}}$ become active---that is, their coefficients $\beta_j$ are nonzero---the partial derivative with respect to $\beta_i$ becomes nonzero. As a result, any stationary point with sufficiently active partners (meaning that $(\beta_j)_{j \in S \backslash \{i\}}$ are bounded away from $0$) must also activate $X_i$ (Corollary~\ref{corollary:extraction-of-interaction-effect}).
%
\end{itemize} 

\vspace{.5em} 

\redblock{
The main conceptual insight of Section~\ref{sec:stationary-points-and-relevant-feature-recovery} is the following. The Laplace and Gaussian RKHSs are both universal RKHSs, and hence are as rich as nonparametric model classes from the viewpoint of function approximation~\cite{SriperumbudurFuLa11}. Nevertheless, when embedded in the compositional kernel model~\eqref{eqn:model-f-beta}, they exhibit qualitatively different feature-recovery behavior at stationary points $\beta_\dagger$. This shows that approximation power alone does not determine feature recovery. Rather, \emph{feature recovery is governed by the variational landscape of $\mathcal J$}, which depends on finer structural properties of the RKHS $H$. We provide a technical preview of these properties in Section~\ref{sec:technical-contributions}, with full theory deferred to Section~\ref{sec:stationary-points-and-relevant-feature-recovery}.
}

\subsection{Related Work} 
This paper contributes to the growing line of work on understanding \emph{feature learning}, broadly defined as the process by which compositional models automatically extract predictive representations from raw inputs by minimizing data-driven objectives~\cite{BengioCoVi13}. A key conceptual distinction of \emph{feature learning} from classical variable selection is that feature learning is inherently \emph{continuous} and \emph{optimization-driven}: instead of making discrete keep-or-discard decisions over coordinates, the representation is parametrized by real-valued weights or transformations that are optimized jointly with the predictor, with the goal of achieving better generalization on unseen data.

Our focus in this paper is on the structure of the solutions---specifically, the support of the global minimizer and stationary points. By studying these supports, we view feature learning through the lens of variable selection, thereby providing a theoretical foundation that connects the continuous landscape of feature learning with the discrete problem of selecting relevant variables.

That said, there are two lines of related work for this piece of manuscript. 
\vspace{.5em} 
\begin{itemize}
\item (\emph{Variable selection in kernel methods}) 
Early work by Vapnik and collaborators considered variable selection in kernel-based models (e.g., support vector machines), using a variational formulation similar to~\eqref{eqn:model-f-beta} but with a binary restriction $\beta \in \{0,1\}^d$~\cite{WestonMuChPoPoVa00, Guyonetal02}. Subsequent studies introduced continuous relaxations---conceptually close to our proposed model~\eqref{eqn:model-f-beta}---mainly for applications in bioinformatics and related areas~\cite{GrandvaletCa02, CaoShSuYaCh07, Allen13, ChenStWaJo17}. Our paper can be viewed as a continuation of this line: we treat $\beta$ as continuous and learn it through optimization. The crucial difference is that previous work did not analyze the statistical variable selection properties of the resulting solutions of $\beta$ in any continuous relaxations---not even for global minimizers, let alone stationary points. As a result, our work provides the missing statistical foundation for this line of research, with guarantees for both global minimizers and stationary points, under various distributional assumptions $\P$ and choices of RKHS $H$.

\vspace{.5em} 
\item (\emph{Feature learning in kernel methods})
A recent line of work---motivated by the empirical success of deep learning---has studied feature learning in the context of kernel methods. The key idea is to extend predictors from the form $f(X)$ to two-layer such as $f(\beta \circ X)$, as studied in this paper, or more generally $f(UX)$ with a learnable matrix $U$~\cite{FukumizuBaJo09, JordanLiRu21, ChenLiLiRu23, Adit24A, FollainBa24, LiRu25, LiRub25, ZhuDaDrFa25}. These formulations are compositional in nature and aim to endow kernel methods with the ability to adaptively discover low-dimensional, task-relevant representations. Empirically, these two parameterizations often offer complementary advantages in practice: the diagonal form $f(\beta \circ X)$ acts directly on raw features, making it suited to datasets where individual features carry interpretable meaning (e.g., tabular datasets), while the full matrix form $f(UX)$ allows learning linear combinations of features and is conceptually closer to two-layer neural networks~\cite{ChenLiLiRu23, Adit24A, Beagleholeetal25, ZhuDaDrFa25, DanhoferDaDuPo25}.

The main contribution of this work is to provide a statistical and variational foundation for learning $f(\beta \circ X)$: we characterize when the continuous, optimization-driven formulation of feature learning in kernel methods~\eqref{eqn:model-f-beta} yields statistically valid variable selection. This perspective is crucial to feature learning because, in regimes where the coordinates of $X$ are independent, correct variable selection in $\beta$ is a necessary prerequisite for generalization. A major theoretical result in this paper reveals that the choice of kernel---particularly the Laplace kernel---plays a decisive role in enabling such feature recovery. At the same time, we establish a strong denoising property of $\mathcal{J}$, showing how irrelevant features can be  removed through continuous optimization. Taken together, these results clarify when optimization in compositional kernel methods can be relied upon to achieve genuine feature learning rather than spurious fits.



Another recent line of work centers on the so-called AGOP (Average Gradient Outer Product) method~\cite{Adit24A}. This line of work also adopts the parameterization $f(\beta \circ X)$ (and $f(UX)$) within kernel ridge regression and can be viewed as operating on the same objective $\mathcal{J}$. AGOP relies on an iterative update scheme $\beta^{(t+1)} = {\rm AGOP}(\beta^{(t)})$ (and $U^{(t+1)} = {\rm AGOP}(U^{(t)})$ resp.) designed to extract a low-dimensional set of relevant features, and has been studied both theoretically---with generalization error bounds that outperform that of standard kernel methods---and empirically, with remarkable success on tasks such as tabular dataset predictions~\cite{Adit24A, ZhuDaDrFa25}. Our study takes a complementary perspective. Instead of analyzing a specific iterative scheme, we focus on the optimization landscape of $\mathcal{J}$ itself and establish population-level guarantees on feature recovery for its global minimizers and stationary points. These results hold under more relaxed distributional assumptions and highlight how the geometry of $\mathcal{J}$ shapes variable selection independently of any algorithmic choices. A central insight of this analysis is that the choice of kernel is crucial: whereas existing guarantees primarily focus on smooth kernels, we show that nonsmooth $\ell_1$-type kernels such as the Laplace kernel enables the recovery of nonlinear main effects that Gaussian kernels miss. Interestingly, a recent algorithmic development on AGOP---namely xRFM~\cite{Beagleholeetal25}, a state-of-the-art method for tabular datasets---demonstrates empirically that adopting the Laplace kernel improves generalization over other kernels (e.g., Gaussian), aligning with the theoretical insights developed in this paper.
\end{itemize}

\redblock{
\subsection{Technical Contributions} 
\label{sec:technical-contributions}
\indent\indent
From a technical standpoint, the most related work is a recent foundational study of compositional models of the form $f(UX)$, which shares one author with the present paper~\cite{LiRu25, LiRub25}. There, the matrix $U \in \R^{d \times d}$ and the function $f \in H$ are learned jointly. The results in Sections~\ref{sec:global-minimizers} and~\ref{sec:stationary-points-and-irrelevant-feature-elimination} of the present paper borrow several ideas from this line of work, although their development in our setting requires additional care due to the different structure of the model. Our setting differs in two important respects. First, we restrict to the diagonal case $U=\operatorname{diag}(\beta)$ and study variable selection, so a coordinate system is fixed in advance. Second—and more significantly—we relax the assumptions on the RKHS $H$: the earlier work focuses on $H$ induced by smooth, translation- and rotation-invariant kernels (e.g., Gaussian), whereas here we work with general translation-invariant RKHS $H$, including the Laplace kernel and other $\ell_1$-type kernels, which is not smooth nor rotation-invariant.

This relaxation is essential to the main conceptual insight developed in Section~\ref{sec:stationary-points-and-relevant-feature-recovery}. Our analysis shows that the choice of kernel fundamentally shapes the \emph{first variation of $\mathcal{J}$}, and thereby determines the feature-recovery properties \emph{at its stationary points}. Intuitively, the derivative of $\mathcal{J}$ with respect to a coordinate $\beta_i$ acts as a \emph{statistical test} for whether a feature $i$ carries marginal predictive signal (see Theorem~\ref{theorem:gradient-expression}). Under the Laplace kernel, and more generally $\ell_1$-type kernels, this derivative with respect to $\beta_i$ involves the map $(x,x') \mapsto |x-x'|$ on $\mathbb{R}^2$, which is \emph{conditionally negative definite}. This property yields a \emph{nonparametric} test statistics, allowing detection of both linear and nonlinear marginal signals (Section~\ref{sec:laplace-kernel-why}). In contrast, for the Gaussian kernel, and more generally $\ell_2$-type radial kernels, the corresponding derivative involves $(x,x') \mapsto |x-x'|^2$, which does not possess this property. As a result, these kernels respond only to linear main effects (Section~\ref{sec:laplace-kernel-why}). Fourier analysis makes this distinction precise and shows that this \emph{conditional negative definiteness} property underlies the contrasting feature recovery behaviors at stationary points (Section~\ref{sec:decomposition-of-derivative}---Section~\ref{sec:relevant-feature-recovery-at-stationary-points}).


To support this analysis, Section~\ref{sec:preliminaries} develops the variational foundations of $\mathcal{J}$, including continuity, directional differentiability, and stationarity conditions, with special care devoted to the nonsmooth behavior of the Laplace kernel. Sections~\ref{sec:global-minimizers} and~\ref{sec:stationary-points-and-irrelevant-feature-elimination} establish initial variable-selection results for global minimizers and stationary points. Remarkably, we show that any stationary point with nontrivial predictive power eliminates Gaussian noise features. Finally, Section~\ref{sec:stationary-points-and-relevant-feature-recovery} develops the paper's central message: $\ell_1$-type kernels such as the Laplace kernel enables stronger guarantees for feature recovery at stationary-points than the $\ell_2$ type kernels such as the Gaussian kernel. We conclude with several open questions in Section~\ref{sec:open-questions}.
}

\subsection{Notation} 
Let \( x \in \R^d \) be a vector. We write \( x = (x_1, \ldots, x_d) \), where \( x_i \) denotes the \(i\)-th coordinate of \(x\), and define its support as \( \supp(x) := \{ i \in \{1, \ldots, d\} : x_i \ne 0 \} \). For a subset \( S \subset \{1, \ldots, d\} \), we write \( x_S \in \R^{|S|} \) for the subvector of \( x \) containing coordinates \( \{x_i : i \in S\} \). We use \( \| \cdot \|_p \) to denote the \( \ell_p \) norm on \( \R^d \).  We use \( e_i \in \R^d \) to denote the \(i\)-th standard basis vector.  We write $\langle x, y \rangle:= \sum_{i=1}^d x_i y_i$ for the standard Euclidean inner product between vectors $x, y \in \R^d$. We use $\bar{z}$ to denote the complex conjugate of a complex number $z \in \C$. For complex-valued random variables $Z_1, Z_2$, we define the covariance as $\Cov(Z_1, Z_2) := \E[Z_1 \bar{Z_2}] -\E[Z_1] \E[\bar{Z_2}]$, and write $\Cov(Z_1):=\Cov(Z_1, Z_1)$. 
Let $C(\R^d)$ denote the space of continuous functions equipped with $L_\infty$ norm. Let $L^p(\R^d)$ denote the Lebesgue space for $1 \le p \le \infty$, and $L^p(\mu)$ denote the Lebesgue space with respect to a measure $\mu$ on $\R^d$.

The Fourier transform of a function \( f : \R^d \to \C \) is denoted by \( \hat{f} \). We use the convention:
\begin{equation*}
	\hat{f}(\omega) = \int_{\R^d} f(x) e^{-2\pi i\langle x, \omega\rangle} dx,~~~f(x) = \int_{\R^d} \hat{f}(\omega) e^{2\pi i \langle x, \omega \rangle} d\omega. 
\end{equation*} 
 The reason
for the $2\pi i$ choice is that it makes the least appearance of dimensional
constants.

\section{Preliminaries} 
\label{sec:preliminaries} 
\subsection{Translation Invariant RKHS} 
\label{sec:translation-invariant-RKHS}
We now give a quick review of translation invariant reproducing kernel
Hilbert spaces (RKHS); more general references on RKHS can be found in~\cite{Aronszajn50, CuckerSm02}. 

Let $H$ be a Hilbert space of real-valued continuous function on $\R^d$ associated to a 
translation-invariant kernel. The space $H$ is defined in terms of the Fourier transform by 
\begin{equation*}
	H := \left\{f: f \in  L^2(\R^d), 
		\norm{f}_H^2 = \int_{\R^d} \frac{|\hat{f}|^2(\omega)}{k(\omega)} d\omega < \infty\right\}
\end{equation*}
where we require $k \in L^1(\R^d)$ to be even, strictly positive on $\R^d$, satisfying the normalizing condition: 
\begin{equation*}
	\int_{\R^d} k(\omega) d\omega = 1.
\end{equation*}

\begin{lemma}[Continuous embedding of $H$ in $C(\R^d)$]
\label{lemma:continuous-embedding}
For any $f \in H$, we have $\norm{f}_{L_\infty} \le \norm{f}_H$, and $f$ has a continuous representative, and $\lim_{x \to \infty} f(x) = 0$.
\end{lemma} 
The proof of Lemma~\ref{lemma:continuous-embedding} is standard and given in Appendix~\ref{sec:proof-lemma-continuous-embedding}. Throughout this paper, we identify $H$ with its image in $C(\R^d)$ and work with continuous representatives. Given that $H$ is continuously embedded in $C(\R^d)$, $H$ is a reproducing kernel Hilbert space. We then define the kernel function
\begin{equation*}
	\kernel(x) = \int_{\R^d} k(\omega) e^{2\pi i \langle x, \omega\rangle} d\omega.
\end{equation*} 
Note that $\kernel$ is real-valued as $k$ is even, and continuous as $k \in L^1(\R^d)$. 
Let $\kernel(x, x') = \kernel(x-x')$.
The continuous function $\kernel(x, \cdot)$ lies in $H$, and has Fourier transform 
$k(\omega) e^{-2\pi i\langle \omega, x\rangle}$. These functions $\kernel(x, \cdot)$ 
are known as \emph{reproducing kernel}. Each $f \in H$ can be represented as 
\begin{equation*}
	f(x) = \langle f, \kernel(x, \cdot)\rangle_H,~~~\forall x\in \R^d.
\end{equation*}

\redblock{
\begin{lemma}[Universality]
\label{lemma:denseness-of-H-in-L-2}
For any probability measure $\nu$ on $\R^d$, the space $H$ is dense in $L^2(\nu)$. 
\end{lemma} 
\vspace{.4em}

This property is often referred to as $L^2$-universality of $H$~\cite[Section 3]{SriperumbudurFuLa11}; in words, $H$ is sufficiently rich to approximate any function in $L^2(\nu)$. It follows from~\cite[Theorem 4.1, Proposition 5.6]{CarmeliDeToUm10}, using that $k$ has full support on $\R^d$. We provide a self-contained proof in Appendix~\ref{sec:proof-of-lemma-denseness-of-H-in-L-2}.
}

\subsubsection{Examples} 
We focus on two canonical types of translation-invariant RKHSs: $\ell_2$-type (radial) and $\ell_1$-type kernels.

\begin{example} [Gaussian RKHS]
The reproducing kernel is $\kernel(x,x') = e^{-\|x-x'\|_2^2}$ on $\R^d$, with
\begin{equation*}
	\kernel(x) = e^{-\norm{x}_2^2},\quad~\text{and}~\quad k(\omega) = \pi^{d/2} e^{- \pi^2 \norm{\omega}_2^2}.
\end{equation*} 
\end{example}

\begin{example} [Laplace RKHS]
The reproducing kernel is $\kernel(x, x') = e^{-\|x-x'\|_1}$ on $\R^d$, with 
\begin{equation*}
	\kernel(x) = e^{-\norm{x}_1},\quad~\text{and}~\quad k(\omega) =  \frac{2^d}{\Pi_{i=1}^d (1+4\pi^2 \omega_i^2)}.
\end{equation*} 
\end{example} 

These two examples extend to broader classes. 
By a classical result of Schoenberg~\cite{Schoenberg38}, $\kernel(x)=\profile(\|x\|_2^2)$ defines 
a translation-invariant RKHS on $\R^d$ for all $d$ if and only if $\profile$ is non-constant and 
\begin{equation*}
	\profile(z) = \int_0^\infty e^{-tz} \mu(dt)
\end{equation*}
for some finite measure $\mu$ on $[0,\infty)$.
Under our normalization, $\kernel(0) = \int k(\omega) d\omega = \int_0^\infty \mu(dt) = 1$, so 
in what follows we take $\mu$ to be a probability measure. 

\vspace{.5em} 
\begin{example}[$\ell_2$-type RKHS (radial RKHS)]
\label{example:radial-kernel-RKHS}
Let $\mu$ be a probability measure with compact support in $(0, \infty)$. Define 
\begin{equation*}
\begin{split} 
	\kernel(x) = \profile(\norm{x}_2^2) = \int_0^\infty e^{-t\norm{x}_2^2}\mu(dt),\quad\quad
	k(\omega) = \int_0^\infty 
		(\pi t^{-1})^{d/2} e^{- \pi^2 t^{-1} \norm{\omega}_2^2}\mu(dt).
\end{split} 
\end{equation*}
Then $\kernel$ and $k$ form a Fourier transform pair, and define a translation-invariant RKHS as above. 
\end{example} 


\vspace{.5em} 
\begin{example}[$\ell_1$ type RKHS]
\label{example:ell-1-type-RKHS}
Let $\mu$ be a probability measure with compact support in $(0, \infty)$. Define 
\begin{equation*}
\begin{split} 
	\kernel(x) =\profile(\norm{x}_1) =\int_0^\infty e^{-t\norm{x}_1}\mu(dt),\quad\quad 
	k(\omega) = \int_0^\infty  \frac{(2t)^d}{\Pi_{i=1}^d (t^2+4\pi^2 \omega_i^2)} \mu(dt).
\end{split} 
\end{equation*}
Then $\kernel$ and $k$ are a Fourier transform pair, and define a translation-invariant RKHS as above. 
\end{example}

\subsection{Kernel Ridge Regression} 
\label{sec:kernel-ridge-regression}
Let $X \in \R^d$ be a random vector, and $Y \in \R$ be a random variable with $\E[Y^2] < \infty$. 
We define 
\begin{equation*}
	\mathcal{I}(f, \beta, \lambda) :=  \E[(Y- f(\beta \circ X))^2] + \lambda \norm{f}_H^2.
\end{equation*}
Consider the following minimization problem, which we will refer to as \emph{kernel ridge regression}: 
\begin{equation*}
	\mathcal{J}(\beta, \lambda) = \min_{f \in \H} \mathcal{I}(f, \beta, \lambda).
\end{equation*}
The existence and uniqueness of the minimizer is well known~\cite{CuckerSm02}, and we denote it by
$f_{\beta, \lambda}$. 
We define 
\begin{equation*}
	r_{\beta, \lambda}(x, y) := y - f_{\beta, \lambda}(\beta \circ x)
\end{equation*}
which we refer to as the \emph{residual function} at $\beta$. It captures the part of $y$ that is not explained by 
the regression model $f_{\beta, \lambda}$ evaluated on the transformed input $\beta \circ x$. 


The following lemmas~\ref{lemma:euler-lagrange-identity} and~\ref{lemma:trivial-bound-on-r-f} are standard; 
proofs are deferred to Appendix~\ref{sec:proof-of-lemma-euler-lagrange-identity} and~\ref{sec:proof-of-trivial-bound-on-r-f}.


\vspace{.5em}
\begin{lemma}[Euler-Lagrangian]
\label{lemma:euler-lagrange-identity}
The minimizer $f_{\beta, \lambda}$ to this problem satisfies 
\begin{equation*}
	\E[(Y- f_{\beta, \lambda}(\beta \circ X)) g(\beta \circ X)] = \lambda \langle f_{\beta, \lambda}, g\rangle_H,~~~\forall g\in H.
\end{equation*}
In particular, by taking $g(\cdot) = \kernel(\cdot, z)$,
\begin{equation*}
	\E[(Y- f_{\beta, \lambda}(\beta \circ X)) \kernel(\beta \circ X, z)] = \lambda f_{\beta, \lambda}(z),~~~\forall z \in \R^d.~~
\end{equation*}
\end{lemma} 

\vspace{.5em}
\begin{lemma}
\label{lemma:trivial-bound-on-r-f}
We always have 
\begin{equation*}
	\E[Y^2] \ge \mathcal{J}(\beta, \lambda) = \E[r_{\beta, \lambda}(X, Y)^2] + \lambda \norm{f_{\beta, \lambda}}_H^2.
\end{equation*} 
The inequality becomes an equality if and only if $\E[Y|\beta \circ X] = 0$.
\end{lemma}


\vspace{.5em}

Finally, we note the following identity relating $f_{\beta,\lambda}$, $r_{\beta,\lambda}$, and $\kernel$; a proof is given in Appendix~\ref{sec:proof-of-lemma-r-r'-K-bound}.

\vspace{.5em}
\begin{lemma}
\label{lemma:r-r'-K-bound}
We always have
	\begin{equation*}	
		\E[r_{\beta, \lambda}(X, Y) r_{\beta, \lambda}(X', Y') \kernel(\beta \circ X, \beta \circ X')] = \lambda^2 \norm{f_{\beta, \lambda}}_H^2.
	\end{equation*}
In the above, $(X', Y')$ denote an independent copy of $(X, Y)$. 
\end{lemma} 

 \vspace{.5em}

\subsection{Continuity in Parameter} 
A fundamental property is that the minimizer $f$ and the objective $\mathcal{J}$ depend 
continuously on the parameter $\beta$. 
This fact essentially follows from the reinterpretation of the $f(UX)$ model in related work~\cite{LiRu25}; 
for completeness, we include a self-contained proof n Appendix~\ref{sec:proof-of-lemma-continuity-of-J}.

\vspace{.5em} 
\begin{lemma}
\label{lemma:continuity-of-J}
For every $\lambda > 0$, the mapping $\beta \mapsto f_{\beta, \lambda}(\beta \circ X)$ is continuous under the $L^2(\P)$ norm, i.e., 
\begin{equation*}
	\lim_{\beta' \to \beta} \E[(f_{\beta', \lambda}(\beta'\circ X) - f_{\beta, \lambda}(\beta \circ X))^2] = 0.
\end{equation*}
Also, the function $\beta \mapsto \mathcal{J}(\beta, \lambda)$ is continuous in $\R^d$.  
\end{lemma} 

\subsection{First Variation} 
\label{sec:first-variation}
We study the first variation of $\mathcal{J}(\beta, \lambda)$ with respect to the parameter $\beta$. 
To this end, we define for every $\beta, x, x' \in \R^d$
\begin{equation*}
	\kernel_\beta(x, x') := \kernel(\beta \circ x, \beta \circ x') = \kernel(\beta \circ (x- x')).
\end{equation*}
We use the notation 
\begin{equation*}
\mathrm{D}\kernel_\beta(x, x')[v] := \lim_{s \to 0^+} \frac{1}{s} (\kernel_{\beta + sv}(x, x') - \kernel_{\beta} (x, x'))
\end{equation*}
to refer to its directional derivative (also called \emph{G\^{a}teaux differential}) at $\beta$ along the direction $v \in \R^d$, whenever the limit exists.

Theorem~\ref{theorem:first-variation-formula} provides an explicit expression for the directional derivative 
of $\mathcal{J}(\beta, \lambda)$. Given any direction $v$ and $\beta$, we write
\begin{equation*}
\mathrm{D} \mathcal{J}(\beta, \lambda)[v] := \lim_{s \to 0^+} \frac{1}{s} \left( \mathcal{J}(\beta + s v, \lambda) - \mathcal{J}(\beta, \lambda) \right)
\end{equation*}
if the limit exists. 

\vspace{.5em} 
\begin{theorem}
\label{theorem:first-variation-formula}
Let $\beta \in \R^d$ and $v \in \R^d$. Suppose there exists $\epsilon > 0$ such that, for every $x, x' \in \R^d$, 
\begin{equation*}
	\mathrm{D}\kernel_{\beta + tv}(x, x')[v]
\end{equation*}
exists for $t \in [0, \eps)$ and is continuous in $t$. In addition, 
assume the uniform integrability condition: 
\begin{equation*}
	 \E\left[\sup_{t \in [0, \eps)} (\mathrm{D} \kernel_{\beta + tv}(X, X')[v])^2\right] < \infty. 
\end{equation*}
Then the directional derivative $\mathrm{D} \mathcal{J}(\beta, \lambda)[v]$ exists and is given by 
\begin{equation*}
	\mathrm{D} \mathcal{J}(\beta, \lambda)[v] =
		- \frac{1}{\lambda} \E[r_{\beta, \lambda}(X, Y) r_{\beta, \lambda}(X', Y') \mathrm{D} \kernel_\beta(X, X')[v]].
\end{equation*}
In the above, $(X', Y')$ denote an independent copy of $(X, Y)$. 
\end{theorem} 

\vspace{.5em} 
Related formulas appeared in the author's prior work on $f(UX)$ model~\cite[Proposition 4.4]{LiRu25}, though under stronger smoothness assumptions, namely Fréchet differentiability on $\beta \mapsto \kernel_\beta(x, x')$ for fixed $x,x'$, and in more specialized settings where $\kernel$ is assumed to be a radial basis kernel.

In contrast, our analysis establishes the directional differentiability of $\mathcal{J}$ under a minimal assumption: that the map $\beta \mapsto \kernel_\beta(x, x')$ admits directional derivatives that satisfy a uniform integrability condition. This relaxation is crucial for handling kernels such as the Laplace kernel, where the map $\beta \mapsto \kernel_\beta(x, x')$ is not Fr\'{e}chet differentiable, yet directional derivatives (i.e., G\^{a}teaux differentials) remain well-defined. The proof of Theorem~\ref{theorem:first-variation-formula} is given in Appendix~\ref{sec:proof-of-theorem-first-variation-formula}.

\redblock{
Conceptually, the argument combines a population-level representer theorem for kernel ridge regression with a differentiation principle in the spirit of Danskin's theorem (see Appendix~\ref{sec:proof-architecture} for the proof architecture). Our proof is inspired by prior work~\cite{LiRu25} but departs from it in essential ways.
}

\subsubsection{Examples} 
We illustrate Theorem~\ref{theorem:first-variation-formula} for two representative RKHSs introduced earlier: 
one induced by a radial kernel and one by an $\ell_1$-type kernel. Both are defined through a function $\profile$ of the form
\begin{equation*}
	\profile(z) = \int_0^\infty e^{-tz} \mu(dt),~~~\text{for}~~~z \ge 0
\end{equation*}
where $\mu$ is a probability measure with compact support that is contained in $(0, \infty)$. The function $\profile$ is 
differentiable on $[0, \infty)$ with 
\begin{equation*}
	\profile^{\prime}(z) = -\int_0^\infty  t e^{-tz} \mu(dt),~~~\text{for}~~~z \ge 0
\end{equation*}
where $\profile'(0)$ is understood as the right-hand derivative of $\psi$ at $0$. 

\vspace{.5em} 
\begin{example}[First Variation for Radial Kernel RKHS]
\emph{
 Let $\mu$ be a probability measure with support in a compact subset of $(0, \infty)$, and define 
$\profile$ as above. In example~\ref{example:radial-kernel-RKHS}, we showed that 
\begin{equation*}
	\kernel(x) = \profile(\|x\|_2^2) 
\end{equation*}
defines a radial kernel and induces an RKHS $H$. 
For such kernels, $\kernel_\beta(x, x') = \profile(\|\beta \circ (x - x')\|_2^2)$ is continuously differentiable in $\beta$, with directional derivative
\begin{equation*}
	\mathrm{D} \kernel_\beta(x, x')[v] = 2 \profile^\prime(\|\beta \circ (x - x')\|_2^2) \cdot \sum_i \beta_i v_i (x_i - x_i')^2.
\end{equation*}
Since $\profile^\prime$ is bounded by $|\profile^\prime(0)| = \int t \mu(dt)$, the uniform integrability condition in Theorem~\ref{theorem:first-variation-formula} holds whenever $X$ has finite fourth moment. In this case, the directional derivative $\mathrm{D} \mathcal{J}(\beta, \lambda)[v]$ exists for all $\beta$ and $v$, and is given by
\begin{equation*}
\begin{split} 
	\mathrm{D} \mathcal{J}(\beta, \lambda)[v] 
	&= -\frac{2}{\lambda} 
		\E[r_{\beta, \lambda}(X, Y) r_{\beta, \lambda}(X', Y') \cdot \profile^\prime(\|\beta \circ (X - X')\|_2^2) \cdot \sum_i \beta_i v_i (X_i - X_i')^2].
\end{split} 
\end{equation*}
This gives the first variation formula to radial kernel RKHSs. In particular, the directional derivative $\mathrm{D} \mathcal{J}(\beta, \lambda)[v]$ is linear in the direction $v$. 
}  $\clubsuit$
\end{example}

\vspace{.5em} 

We can analogously compute the directional derivative for $\ell_1$ type kernel RKHS. 

\vspace{.5em} 
\begin{example} [First Variation for $\ell_1$-type RKHS]
\label{example:first-variation-for-ell-one-type-RKHS}
\emph{
 Let $\mu$ be a probability measure with support in a compact subset of $(0, \infty)$, and define 
$\profile$ as above. In example~\ref{example:ell-1-type-RKHS}, we showed that 
\begin{equation*}
	\kernel(x) = \profile(\|x\|_1) 
\end{equation*}
defines an $\ell_1$ type kernel and induces an RKHS $H$. 
For such kernels, $\kernel_\beta(x, x') = \profile(\|\beta \circ (x - x')\|_1)$ is directionally differentiable for every $\beta$ and direction $v$, with directional derivative
\begin{equation*}
	\mathrm{D} \kernel_\beta(x, x')[v] = \profile^\prime(\|\beta \circ (x - x')\|_1) \cdot \sum_i w(\beta_i; v_i)|x_i - x_i'|,
\end{equation*}
where 
\begin{equation*}
	w(\beta_i; v_i) := \begin{cases}
		{\rm sgn}(\beta_i) v_i~~~&\beta_i \ne 0\\
		|v_i|~~~&\beta_i = 0
	\end{cases}.
\end{equation*}
Since $\profile^\prime$ is bounded, the uniform integrability condition in Theorem~\ref{theorem:first-variation-formula} holds whenever $X$ has finite second moment. In this case, the directional derivative $\mathrm{D} \mathcal{J}(\beta, \lambda)[v]$ exists for all $\beta$ and $v$, with 
\begin{equation*}
\begin{split} 
	\mathrm{D} \mathcal{J}(\beta, \lambda)[v] 
	&= -\frac{1}{\lambda} 
		\E[r_{\beta, \lambda}(X, Y) r_{\beta, \lambda}(X', Y') \cdot \profile^\prime(\|\beta \circ (X - X')\|_1) \cdot \sum_i w(\beta_i; v_i)|X_i - X_i'|].
\end{split} 
\end{equation*}
Unlike the radial kernel RKHSs, the directional derivative 
$\mathrm{D} \mathcal{J}(\beta, \lambda)[v]$ is \emph{not} linear in $v$. 
Thus, for $\ell_1$ type kernels, the objective $\mathcal{J}$ is not differentiable in the classical (Fr\'echet) sense.
}  $\clubsuit$
\end{example}
Theorem~\ref{theorem:first-variation-formula} and the above examples illustrate that, in many scenarios of interest, 
the directional derivative $\mathrm{D} \mathcal{J}(\beta, \lambda)[v]$ exists for all $\beta \in \R^d$ and all $v \in \R^d$. 
Accordingly, throughout the paper we assume this derivative exists for all $\beta$ and $v$.

This directional regularity is fundamental---it enables a principled notion of stationarity, which play a central role in understanding the landscape of $\mathcal{J}$.

\subsection{Directional Stationarity and Local Optimality}
This paper is concerned with the minimization problem of $\mathcal{J}(\beta, \lambda)$ over $\beta$: 
\[
	\inf_{\beta \in \R^d}~ \mathcal{J}(\beta, \lambda).
\]
In general, the objective $\mathcal{J}(\beta, \lambda)$ is nonconvex in the parameter $\beta$, so global minimization is typically intractable for numerical algorithms. It is therefore more natural to study local notions of optimality, which better reflect the types of solutions produced by numerical optimization methods.

\vspace{.5em} 
\begin{definition}
We say that $\beta_* \in \R^d$ is a \emph{global minimizer} of $\mathcal{J}(\beta, \lambda)$ if 
$
	\mathcal{J}(\beta_*, \lambda) = \inf_{\beta \in \R^d} \mathcal{J}(\beta, \lambda)$. 
We say that $\beta_{\dagger} \in \R^d$ is a \emph{local minimizer} of $\mathcal{J}(\beta, \lambda)$ if there exists $\delta > 0$ such that
\[
	\mathcal{J}(\beta_{\dagger}, \lambda) \le \mathcal{J}(\beta, \lambda) \quad \text{for all } \beta \in \R^d \text{ with } \|\beta - \beta_{\dagger}\| \le \delta.
\]
\end{definition}

When $\mathcal{J}$ admits directional derivatives, a necessary condition for local optimality is the nonnegativity of all directional derivatives~\cite[Chapter 6]{RockafellarTyWe98}. 
This motivates the notion of \emph{directional stationary points} (also known as \emph{D-stationary points} in the variational analysis literature~\cite{RockafellarTyWe98}).


\vspace{.5em} 
\begin{definition} 
We say that $\beta_{\dagger} \in \R^d$ is a directional stationary point of $\mathcal{J}$ if 
\begin{equation*}
	\mathrm{D}\mathcal{J} (\beta_{\dagger}, \lambda)[v] \ge 0,~~~\forall v\in \R^d.
\end{equation*}
\end{definition} 
\begin{remark}
\emph{Any local minimizer $\beta_{\dagger}$ is directional stationary, although the reverse need not hold. }
\end{remark} 

\vspace{.5em}

In the sequel, we study the statistical properties of global minimizers $\beta_*$, local minimizers, and directional stationary points $\beta_\dagger$. While global minimizers $\beta_*$ serve as a benchmark, our focus is on whether local minimizers and stationary points $\beta_\dagger$---typically reached by iterative algorithms---also support feature learning, i.e., recover low-dimensional structure in data relevant for prediction.

\section{Global Minimizers}
\label{sec:global-minimizers}
Towards understanding the landscape of the $\mathcal{J}$, we begin by analyzing the global minimizers of $\mathcal{J}$, and focus on two foundational questions. 

\vspace{0.5em}
\noindent
$\mathsf{Q1}$ \textbf{(Existence).}
\textit{Does $\mathcal{J}(\beta, \lambda)$ attain its minimum at some $\beta \in \R^d$?}

\vspace{0.5em}
\noindent
This question ensures the global minimizers are well-defined and excludes degenerate cases where minimizers ``escape to infinity". On a positive side, we will show that a global minimizer of $\mathcal{J}$ always exists whenever $X$ is a continuous random variable. On the negative side, when $X$ is discrete, global minimizers need not be well-defined.

\vspace{1em}
\noindent
$\mathsf{Q2}$ \textbf{(Feature Recovery).}
\textit{Does the global minimizer of $\mathcal{J}(\beta, \lambda)$ select the coordinates of $X$ that are predictive of $Y$?}

This raises a fundamental statistical question: can the compositional model $f(\beta \circ X)$ uncover the relevant sparse structure in the data at its global minimizer? 

\subsection{Existence} 
In this section, we address the question $(\mathsf{Q1})$. We show that when $X$ is a continuous random variable, $\mathcal{J}$ admits a global minimizer in $\R^d$. 

\vspace{.5em} 
\begin{theorem}
\label{theorem:existence-of-global-minimizers}
Assume $X$ is a continuous random vector on $\R^d$. Then, for every $\lambda > 0$,
\begin{equation*}
	\lim_{\beta \to \infty} \mathcal{J}(\beta, \lambda) = \E[Y^2]. 
\end{equation*}
As a corollary, the function $\mathcal{J}$ admits a global minimizer $\beta_* \in \R^d$ with $\mathcal{J}(\beta_*, \lambda) = \inf_{\beta \in \R^d} \mathcal{J}(\beta, \lambda)$.
\end{theorem}
\begin{proof} 
First, by plugging a test function of all zero, we see that 
\begin{equation*}
	\mathcal{J}(\beta, \lambda) \le \mathcal{I}(0, \beta, \lambda) = \E[Y^2],~~~\forall \beta \in \R^d. 
\end{equation*}
Next, we expand the objective: 
\begin{equation*}
	\mathcal{I}(f, \beta, \lambda) = \E[Y^2] - 2\E[Y f(\beta \circ X)] + \E[f(\beta \circ X)^2] + \lambda \norm{f}_H^2.
\end{equation*}
Applying the reproducing property and Cauchy–Schwarz inequality yields
\begin{equation*}
\begin{split}
	\E[Y f(\beta \circ X)] &= \E[Y \langle \kernel(\beta \circ X, \cdot), f \rangle_H] \\
		&= \langle \E[Y  \kernel(\beta \circ X, \cdot)], f \rangle_H
		\le \norm{\E[Y  \kernel(\beta \circ X, \cdot)]}_H \norm{f}_H.
\end{split}
\end{equation*}
The squared RKHS norm admits an explicit expression: 
\begin{equation*}
	\norm{\E[Y  \kernel(\beta \circ X, \cdot)]}_H^2 
		= \langle \E[Y  \kernel(\beta \circ X, \cdot)], \E[Y'  \kernel(\beta \circ X', \cdot)]\rangle_H
		= \E[YY'\kernel(\beta \circ X, \beta \circ X')]
\end{equation*} 
where $(X', Y') \sim \P$ is an independent copy of $(X, Y)$. 
Combining the above gives a lower bound 
\begin{equation*}
\begin{split} 
	\mathcal{I}(f, \beta, \lambda) 
		&\ge \E[Y^2] - 2 \sqrt{\E[YY' \kernel(\beta \circ X, \beta \circ X')]} \norm{f}_H + \lambda \norm{f}_H^2.
\end{split} 
\end{equation*} 
Minimizing the right-hand side over $\norm{f}_H \in [0, \infty)$ gives 
\begin{equation*}
	\mathcal{J} (\beta, \lambda) = \min_{f} \mathcal{I}(f, \beta, \lambda) 
		\ge \E[Y^2] - \lambda^{-1} \E[YY'\kernel(\beta \circ X, \beta \circ X')].
\end{equation*} 
Note that $\kernel(\beta \circ x, \beta \circ x') = \kernel(\beta \circ (x-x')) \to 0$ as $\beta \to \infty$ when $x \ne x'$. 
Additionally, $\P(X \ne X') = 1$ since $X$ is assumed to be continuous. As $\kernel$ is uniformly bounded,  Lebesgue’s dominated convergence theorem then implies $\E[YY'\kernel(\beta \circ X, \beta \circ X')] \to 0$ as $\beta \to \infty$. This leads to 
\[
	\liminf_{\beta \to \infty} \mathcal{J}(\beta, \lambda) \ge \E[Y^2].
\]
Since $\mathcal{J}(\beta, \lambda) \le \E[Y^2]$ for all $\beta$, this implies the limit 
$\lim_{\beta \to \infty} \mathcal{J}(\beta, \lambda) $ exists and equals $\E[Y^2]$.

Finally, since $\mathcal{J}$ is continuous in $\beta$ (Lemma~\ref{lemma:continuity-of-J}) and bounded above by $\E[Y^2]$ (Lemma~\ref{lemma:trivial-bound-on-r-f}), it must attain its infimum over $\R^d$. That is, there exists $\beta_* \in \R^d$ such that
$
	\mathcal{J}(\beta_*, \lambda) = \inf_{\beta \in \R^d} \mathcal{J}(\beta, \lambda)$. 
\end{proof} 

Continuity of $X$ serves as a sufficient condition for $\mathcal{J}$ to admit a finite minimizer $\beta_*$. In contrast, when $X$ is discrete, the objective $\mathcal{J}$ may fail to attain its infimum (and any minimizing sequence must diverge to infinity); see Section~\ref{sec:discrete-counterexample}.

\subsection{Feature Recovery}

From a variable selection perspective, we are interested in whether the global minimizer $\beta_*$ of the population objective $\mathcal{J}(\beta, \lambda)$ captures relevant features of the input $X$---those that are predictive of the response $Y$. This leads to three guiding questions.

\begin{itemize}
\item[$\mathsf{(a)}$] \emph{Characterization of Relevant Features:} 
How should we formally define the minimal set of features in $X$ sufficient for predicting $Y$?

\item[$\mathsf{(b)}$] \emph{Irrelevant Feature Elimination:}
Does the minimizer $\beta_*$ assign zero weight to features that are unrelated to the prediction of $Y$?

\item[$\mathsf{(c)}$] \emph{Recovery of Predictive Structure:}
Does the minimizer $\beta_*$ capture at least some---or all---of the features that contribute to predicting $Y$?
\end{itemize}
We address these three questions in the subsections that follow.

%

\redblock{
\subsubsection{Minimal Sufficient Feature Set} 
\label{sec:minimal-sufficient-feature-set}
}
To formalize the notion of task-relevant features, we introduce the concept of a minimal sufficient feature set: the smallest subset of the input $X$ that retains all information needed to predict the response $Y$.

\vspace{0.5em}
\begin{definition}[Minimal Sufficient Feature Set]
\label{definition:minimal-sufficient-feature-set}
Let
\begin{equation}
\label{eqn:core-sufficient-feature-subset}
S_* := \bigcap \{S \subset \{1, 2, \ldots, d\}: \E[Y|X] = \E[Y|X_S] \}.
\end{equation} 
If $S_*$ additionally satisfies $\E[Y|X] = \E[Y|X_{S_*}]$, then it is called the \emph{minimal sufficient feature set}.
\end{definition}

\vspace{0.5em}
\begin{remark}
\emph{
This notion parallels the classical concept of the \emph{central mean subspace} for sufficient dimension reduction. The central mean subspace is the smallest linear subspace $U \subseteq \mathbb{R}^d$ such that $\E[Y|X] = \E[Y|P_U X]$, where $P_U$ is the orthogonal projection onto $U$~\cite{CookLi02, Cook09}. In contrast, the minimal sufficient feature set seeks a minimal subset of coordinate indices $S$ such that $\E[Y|X] = \E[Y|X_S]$. Both aim to identify minimal representations of the input $X$ that preserve the full predictive information about $Y$, though in different structural forms.
}
\end{remark} 

\vspace{0.5em}
\begin{remark}
\emph{
Note that a minimal sufficient feature set may not always exist. For instance, if $X_1 = X_2$ and $\E[Y|X] = \E[Y|X_1] = \E[Y|X_2]$, then both $\{1\}$ and $\{2\}$ satisfy the condition $\E[Y|X] = \E[Y|X_S]$ but their intersection $\{1\} \cap \{2\} = \emptyset$ does not satisfy $\E[Y|X] = \E[Y]$ in general. Consequently, the set $S_*$ defined in \eqref{eqn:core-sufficient-feature-subset} fails to satisfy $\E[Y|X] = \E[Y|X_{S_*}]$, and no minimal sufficient feature set exists.
}
\end{remark}

\vspace{0.5em} 
Despite the ill-conditioned case noted in the preceding remark, Proposition~\ref{proposition:existence-of-minimal-sufficient-feature-set} shows that, under the mild assumption that $X$ is a continuous random variable with support $\mathbb{R}^d$, the minimal sufficient feature set $S_*$ exists. The proof is routine; for completeness, we include it in Appendix~\ref{sec:proof-of-proposition-existence-of-minimal-sufficient-feature-set}.

\vspace{0.5em}
\begin{proposition} 
\label{proposition:existence-of-minimal-sufficient-feature-set}
Assume that $X$ has positive density with respect to Lebesgue measure on $\R^d$. 
Then the minimal sufficient feature set $S_*$ exists. 
\end{proposition}

\vspace{0.5em}
Beyond Proposition~\ref{proposition:existence-of-minimal-sufficient-feature-set}, we believe that
the minimal sufficient feature set $S_*$ exists in a wide range of practical scenarios, though a full characterization is beyond the scope of this paper. For instance, when $X$ is uniformly distributed on the hypercube, the set $S_*$ exists as a consequence of the Fourier expansion applied to $\E[Y | X]$~\cite{ODonnell14}. This belief is also supported by a large body of work in sufficient dimension reduction, where analogous constructs---such as the central mean subspace---are known to exist under remarkably mild assumptions~\cite{Cook09}.

To study feature learning rigorously, we focus on settings where a well-defined 
target structure exists. Accordingly, we assume throughout: 

\vspace{0.5em} 
\begin{assumption}
\label{assumption:existence-of-csfs}
The minimal sufficient feature set $S_*$ exists for the distribution $(X, Y) \sim \P$. 
\end{assumption} 
\vspace{0.5em} 

Under Assumption~\ref{assumption:existence-of-csfs}, the set $S_*$ provides a meaningful statistical target: 
the minimal subset of input features needed to predict $Y$. The problem of variable selection concerns whether minimizing $\mathcal{J}$ successfully recovers $S_*$. Understanding when and why this recovery succeeds is key to revealing how the compositional model $f(\beta \circ X)$ learns structure from data.


\subsubsection{Irrelevant Feature Elimination} 
We first study one aspect of feature learning: when does the global minimizer $\beta_*$ eliminate irrelevant features---that is, assign zero weight to coordinates outside the minimal sufficient feature set $S_*$? Such sparsity reflects the model’s ability to suppress irrelevant inputs automatically.

\vspace{0.5em}
\begin{assumption}
\label{assumption:independence}
Assume that $X_{S_*}$ and $X_{S_*^c}$ are probabilistically independent. Here, $S_*$ denotes the 
minimal sufficient feature set of the distribution $(X, Y) \sim \P$, and $S_*^c = \{1, 2, \ldots, d\} \backslash S_*$ denotes its complement.
\end{assumption} 

This assumption posits that the relevant features in $X$ for predicting $Y$---those in $S_*$---are independent of the rest. Such independence avoids statistical confounding, making it clear to distinguish relevant features based solely on their predictive contribution. 

\redblock{
\vspace{.5em} 
\begin{remark} 
\label{remark:independence-assumption}
\emph{
We discuss Assumption~\ref{assumption:independence} in relation to conditions commonly considered in the nonparametric feature selection literature in statistics. 
\vspace{.4em} 
\begin{itemize}
\item Assumption~\ref{assumption:independence} holds in settings where $X$ can be designed to follow a prescribed distribution with independent coordinates---for example, in computer experiments~\cite{Santner2003}---such as an isotropic Gaussian distribution $X \sim \normal(0, I_d)$. Such independence assumptions are commonly imposed in the analysis of nonparametric feature selection methods; see, e.g., the seminal work~\cite{LaffertyWasserman08}.
\item Even when $X$ does not satisfy Assumption~\ref{assumption:independence}, the analysis can be reduced to this setting when the marginal distribution of $X$ is known. In this case, a change of measure (also called sample reweighting~\cite{RobertCasella04}) allows one to work under a reference distribution (e.g., $X \sim \normal(0, I_d)$) that satisfies Assumption~\ref{assumption:independence}. Such settings---knowing the distribution of $X$---are also common in the nonparametric feature selection literature; see, for instance,~\cite{CommingesDa12, CandesFaJaLv18}.
\end{itemize}
}
\end{remark} 
}

\vspace{1em} 


For a subset $S \in \{1, 2, \ldots, d\}$, $\Pi_{S}: \R^d \to \R^d$ denote the coordinate projection onto $S$, defined by
\begin{equation*}
	(\Pi_{S} \beta)_j =
		\begin{cases} 
			\beta_j~~\text{for $j \in S$} \\
			0~~~\text{for $j \in S^c$} 
		\end{cases}.
\end{equation*}

\begin{theorem}
\label{theorem:projection-reduce-function-value}
Assume Assumptions~\ref{assumption:existence-of-csfs},~\ref{assumption:independence}. For every $\lambda > 0$
\begin{equation*}
	\mathcal{J}(\beta, \lambda) \ge \mathcal{J}(\Pi_{S_*} \beta, \lambda),~~~\forall \beta \in \R^d.
\end{equation*} 
The inequality is strict when $\Cov(\Pi_{S_*^c} \beta \circ  X) \neq 0$ and $J(\beta, \lambda) < \E[Y^2]$.
\end{theorem} 

\begin{proof} 
For every $\xi$ in $\R^d$, we define the translated function $\tau_\xi f$ by
\begin{equation*}
	(\tau_\xi f)(x):= f(x+\xi),~~~\forall x \in \R^d.
\end{equation*}
By the translation-invariance of the RKHS $H$, we have $\norm{\tau_\xi f}_H = \norm{f}_H$. By the minimizing property of $\mathcal{J}$, we deduce for every vector $\xi$ in $\R^d$, and $f \in H$ that 
\begin{equation*}
\begin{split} 
	&\E[(Y - f( \beta \circ \Pi_{S_*} X + \xi))^2] + \lambda \norm{f}_H^2 \\
	&= \E[(Y - (\tau_\xi f)( \beta \circ \Pi_{S_*} X))^2] + \lambda \norm{\tau_\xi f}_H^2  \\
	&= \E[(Y - (\tau_\xi f)( \Pi_{S_*} \beta \circ X))^2] + \lambda \norm{\tau_\xi f}_H^2 
		\ge \mathcal{J}(\Pi_{S_*} \beta, \lambda).
\end{split} 
\end{equation*}
Define $c := \E[(Y - \E[Y|X])^2]$. Since $\E[Y|X] = \E[Y|\Pi_{S_*} X]$, we can decompose the squared error as
\[
\E[(Y - f(\beta \circ \Pi_{S_*} X + \xi))^2] = \E\left[(\E[Y|\Pi_{S_*} X] - f(\beta \circ \Pi_{S_*} X + \xi))^2\right] + c.
\]
Substituting into the previous inequality gives
\[
\E\left[(\E[Y|\Pi_{S_*} X] - f(\beta \circ \Pi_{S_*} X + \xi))^2\right] + \lambda \|f\|_H^2 + c \ge \mathcal{J}(\Pi_{S_*} \beta, \lambda).
\]

By Assumption~\ref{assumption:independence}, $\Pi_{S_*} X$ and $\Pi_{S_*^c} X$ are independent. Therefore, setting $\xi := \beta \circ \Pi_{S_*^c} X$, we may condition on $\Pi_{S_*^c} X$ to obtain
\[
  \E\left[(\E[Y|\Pi_{S_*} X] - f(\beta \circ \Pi_{S_*} X + \beta \circ \Pi_{S_*^c} X))^2 \mid  \Pi_{S_*^c} X\right] + \lambda \|f\|_H^2 + c \ge \mathcal{J}(\Pi_{S_*} \beta, \lambda).
\]
Taking expectation on both sides, and using $X = \Pi_{S_*} X + \Pi_{S_*^c} X$, we get 
\begin{equation*}
\begin{split} 
	\mathcal{I}(f, \beta, \lambda) 
		&=  \E[(Y - f( \beta \circ X))^2] + \lambda \norm{f}_H^2  \\
		&= \E[(Y - f( \beta \circ \Pi_{S_*} X + \beta \circ \Pi_{S_*^c} X))^2] + \lambda \norm{f}_H^2 \\
		&=   \E\left[(\E[Y|\Pi_{S_*} X] - f(\beta \circ \Pi_{S_*} X + \beta \circ \Pi_{S_*^c} X))^2\right] + \lambda \|f\|_H^2 + c 
			\ge  \mathcal{J}(\Pi_{S_*} \beta, \lambda).
\end{split} 
\end{equation*}
By minimizing over $f \in H$, we get 
\begin{equation*}
	\mathcal{J}(\beta, \lambda)= \min_f \mathcal{I}(f, \beta, \lambda)  \ge \mathcal{J}(\Pi_{S_*} \beta, \lambda).
\end{equation*}  
This proves the desired inequality. 

Our derivation shows equality holds in 
$\mathcal{J}(\beta, \lambda) = \mathcal{J}(\Pi_{S_*} \beta, \lambda)$
only if the minimizer $f_{\beta, \lambda}$ obeys 
\begin{equation*}
(\tau_{\xi}) f_{\beta, \lambda} = (\tau_{\xi'}) f_{\beta, \lambda}~~\text{for almost all}~\xi=\beta \circ \Pi_{S_*^c} X, \xi' = \beta \circ \Pi_{S_*^c}X'
\end{equation*}
where $X, X'$ are independent copies from $\P$. 
If $\operatorname{Cov}(\Pi_{S_*^c} \beta \circ  X) \ne 0$, then with positive probability we have $\xi \ne \xi'$. 
Fix such a realization with $\xi \ne \xi'$. Then, for the 
the deterministic vector $z = \xi - \xi' \ne 0$,
\begin{equation*}
	f_{\beta, \lambda}(x + z)  = f_{\beta, \lambda}(x)~~~\forall x\in \R^d.
\end{equation*}
However, since $f_{\beta, \lambda} \in H$, it satisfies $\lim_{x \to \infty} f_{\beta, \lambda}(x) = 0$ 
according to Lemma~\ref{lemma:continuous-embedding}. In turn, this implies that $f_{\beta, \lambda}$ must be identically zero. As a consequence, it must follow that $J(\beta, \lambda) = \E[Y^2]$.
\end{proof} 

We provide sufficient and necessary conditions for which the minimum value $\mathcal{J}(\beta_*, \lambda) < \E[Y^2]$. 

\vspace{.5em} 
\begin{assumption}
\label{assumption:for-the-global-minimizer}
Assume $\E[Y|X] \neq 0$ under $\mathcal{L}_2(\P)$.
\end{assumption} 

\vspace{.5em} 
\begin{lemma}
\label{lemma:infimum-nontrivial}
Assumption~\ref{assumption:for-the-global-minimizer} holds if and only if 
\[
	\inf_{\beta \in \R^d} \mathcal{J}(\beta, \lambda) < \E[Y^2].
\] 
\end{lemma} 

\begin{proof}
Assume $\E[Y|X] \ne 0$. Let $\mathbf{1} \in \R^d$ denote the all-ones vector. We show that $\mathcal{J}(\mathbf{1}, \lambda) < \E[Y^2]$. Suppose instead that $\mathcal{J}(\mathbf{1}, \lambda) = \E[Y^2]$. Then the minimizer $f_{\mathbf{1}, \lambda}$ must be identically zero, which by Lemma~\ref{lemma:euler-lagrange-identity} implies $\E[Y g(X)] = 0$ for all $g \in H$. Since $H$ is dense in $L^2(\P_X)$ by Lemma~\ref{lemma:denseness-of-H-in-L-2}, it follows that $\E[Y g(X)] = 0$ for all $g \in L^2(\P_X)$, and hence $\E[Y|X] = 0$ almost surely---contradicting our assumption $\E[Y|X] \ne 0$.
We conclude that $\mathcal{J}(\mathbf{1}, \lambda) < \E[Y^2]$, and thus
$\inf_{\beta \in \R^d} \mathcal{J}(\beta, \lambda) < \E[Y^2]$. 

Conversely, assume $\E[Y|X] = 0$. Then the minimizer $f_{\beta, \lambda}$ is the zero function for every $\beta$. 
As a result, $\mathcal{J}(\beta, \lambda) = \E[Y^2]$ for all $\beta$, and hence 
$\inf_{\beta \in \R^d} \mathcal{J}(\beta, \lambda) = \E[Y^2]$. 
\end{proof} 

\vspace{.5em} 
\begin{assumption}
\label{assumption:mild-assumption}
Assume the irrelevant features $(X_i)_{i \in S_*^c}$ obeys $\Var(X_i) > 0,~\forall i \in S_*^c$. 
\end{assumption}

\vspace{.5em} 
We are now ready to state our main result, which combines 
Theorem~\ref{theorem:projection-reduce-function-value} and Lemma~\ref{lemma:infimum-nontrivial}.

\vspace{.5em} 
\begin{theorem}
\label{theorem:global-minimizer}
Assume Assumptions~\ref{assumption:existence-of-csfs},~\ref{assumption:independence},~\ref{assumption:for-the-global-minimizer},~\ref{assumption:mild-assumption}. 
For every regularization parameter $\lambda > 0$,  
\begin{equation*}
	\supp(\beta_*) \subset S_*,~~~\forall~\beta_*~\text{minimizing}~\mathcal{J}(\cdot, \lambda).
\end{equation*}
\end{theorem} 
\begin{remark}
\emph{
If $\mathcal{J}(\cdot, \lambda)$ does not attain its infimum in $\R^d$, we interpret the set of global minimizers as empty. In this case, all conclusions involving global minimizers should be understood as vacuously true. The same remark applies to later statements.
}
\end{remark}

\begin{proof}[Proof of Theorem~\ref{theorem:global-minimizer}]
Fix $\lambda > 0$ and let $\beta_*$ be a global minimizer. Suppose, for contradiction, that 
$\supp(\beta_*) \not  \subset S_*$. Then $\Pi_{S_*^c} \beta_* \ne 0$. 
Under Assumption~\ref{assumption:for-the-global-minimizer}, we have
$\mathcal{J}(\beta_*, \lambda) < \E[Y^2]$ by Lemma~\ref{lemma:infimum-nontrivial}.
Moreover, Assumption~\ref{assumption:mild-assumption} guarantees that
$\Cov(\Pi_{S_*^c} \beta_* \circ X) \ne 0$.
Finally, combining Assumptions~\ref{assumption:existence-of-csfs} and~\ref{assumption:independence},
Theorem~\ref{theorem:projection-reduce-function-value} implies the strict inequality $\mathcal{J}(\beta_*, \lambda) > \mathcal{J}(\Pi_{S_*} \beta_*, \lambda)$, contradicting the optimality of $\beta_*$. This completes the proof.
\end{proof}

\subsubsection{Relevant Feature Recovery} 
In contrast to the previous subsection on eliminating irrelevant features, we now ask the complementary question: when does the global minimizer $\beta_*$ retain features that are truly predictive of $Y$? That is, does $\beta_*$ assigns nonzero weights to the components of the minimal sufficient feature set $S_*$? This question is critical for understanding whether feature learning preserves all information relevant for prediction---not just filters out the noise.

\vspace{.5em} 
\begin{theorem}
\label{theorem:global-minimizer-two}
Assume Assumptions~\ref{assumption:existence-of-csfs},~\ref{assumption:independence}. 
There exists $\lambda_0 > 0$ such that for every $\lambda \in (0, \lambda_0)$: 
\begin{equation*}
	\supp(\beta_*) \supset S_*,~~~\forall~\beta_*~\text{minimizing}~\mathcal{J}(\cdot, \lambda). 
\end{equation*}
\end{theorem} 
\begin{proof} 
For any $f \in H$, we have 
\begin{equation*}
\begin{split} 
	\mathcal{I}(f, \beta, \lambda) &= \E[(Y - f(\beta \circ X))^2] + \lambda \norm{f}_H^2  \\
		&\ge \E[(Y - \E[Y| \beta \circ X])^2] = \E[(Y - \E[Y|X_{\supp(\beta)}])^2]
\end{split} 
\end{equation*} 
where the inequality follows from the optimality of conditional expectation.
Taking the infimum over $f$ yields, for any $\beta$ and any $\lambda > 0$,
\begin{equation*}
	\mathcal{J}(\beta, \lambda) \ge \E[(Y - \E[Y|X_{\supp(\beta)}])^2].
\end{equation*}
Since $S_*$ is the minimal sufficient feature set, any set $S$ with $\E[Y|X] = \E[Y|X_S]$ must obey 
$S_* \subset S$. Thus, 
\begin{equation*}
	\inf\{\E[(Y - \E[Y|X_S])^2 | S_* \not \subset S\} > \E[(Y - \E[Y|X])^2],
\end{equation*} 
which implies the existence of $\varepsilon > 0$ such that for all $\lambda > 0$,
\begin{equation*}
	\inf\{\mathcal{J}(\beta, \lambda) \mid S_* \not \subset \supp(\beta)\} \ge \E[(Y - \E[Y|X])^2] + \eps.
\end{equation*}
On the other hand, $H$ is dense in $L^2(\P_X)$ by Lemma~\ref{lemma:denseness-of-H-in-L-2}, where $\P_X$ denotes the distribution of $X$. Thus, the regression function $x \mapsto \E[Y | X = x]$ can be approximated arbitrarily well in $\mathcal{L}_2(\P_X)$ by functions in $H$. In particular, there exists $f_\varepsilon \in H$ such that
\begin{equation*}
	\E[(Y-f_\varepsilon (X))]^2 < \E[(Y - \E[Y|X])^2] + \eps/2.
\end{equation*}
Let $\mathbf{1} \in \R^d$ denote the all-ones vector. Then for sufficiently small $\lambda > 0$, 
\begin{equation*}
	\mathcal{J}(\mathbf{1}, \lambda) \le 
		\mathcal{I} (f_\varepsilon, \mathbf{1}, \lambda) = \E[(Y-f_\varepsilon (X))]^2 + \lambda \norm{f_\varepsilon}_H^2 < 
		\E[(Y - \E[Y|X])^2] + \eps,
\end{equation*}
so we have 
\begin{equation*}
	\inf\{\mathcal{J}(\beta, \lambda) \mid S_* \not \subset \supp(\beta)\}  > J(\mathbf{1}, \lambda)
		\ge \inf \{\mathcal{J}(\beta, \lambda) \mid \beta \in \R^d\}.
\end{equation*} 
It follows that, for these small $\lambda > 0$, any global minimizer $\beta_*$ of $\mathcal{J}(\cdot, \lambda)$ must satisfy $\supp(\beta_*) \supset S_*$, since otherwise its objective would exceed $\mathcal{J}(\mathbf{1}, \lambda)$.
\end{proof}

\redblock{
\subsubsection{Exact Feature Set Identification} 
\label{sec:exact-feature-identification} 
\indent\indent
We offer conditions under which any global minimizer $\beta_*$ of $\mathcal{J}$ \emph{exactly} identifies the minimal sufficient feature set $S_*$. Theorem~\ref{theorem:global-minimizer-three} follows immediately from Theorems~\ref{theorem:global-minimizer} and~\ref{theorem:global-minimizer-two}.

\vspace{.5em} 
\begin{theorem}
\label{theorem:global-minimizer-three}
Assume Assumptions~\ref{assumption:existence-of-csfs},~\ref{assumption:independence},~\ref{assumption:for-the-global-minimizer},~\ref{assumption:mild-assumption}. 
There exists $\lambda_0 > 0$ such that if $\lambda \in (0, \lambda_0)$, 
\begin{equation*}
	\supp(\beta_*) = S_*,~~~\forall~\beta_*~\text{minimizing}~\mathcal{J}(\cdot, \lambda). 
\end{equation*}
\end{theorem}

We now examine the assumptions underlying Theorem~\ref{theorem:global-minimizer-three}, with the goal of clarifying their roles and assessing their necessity.

\vspace{.8em} 
\begin{itemize}
\item Assumption~\ref{assumption:existence-of-csfs} is necessary, as it ensures the minimal sufficient feature set $S_*$ is well-defined.
\vspace{.3em} 
\item  Assumption~\ref{assumption:independence} enforces independence between relevant $X_{S_*}$ and irrelevant features $X_{S_*^c}$. 
It is comparable to standard assumptions in nonparametric feature selection problems in statistics (see Remark~\ref{remark:independence-assumption}); however, its necessity is unclear and we leave it as an open question.
\vspace{.3em} 
\item Assumption~\ref{assumption:for-the-global-minimizer} is necessary for Theorem~\ref{theorem:global-minimizer-three} to hold. If Assumption~\ref{assumption:for-the-global-minimizer} is violated, then $\E[Y| X] = 0$, which implies that $\mathcal{J}(\beta, \lambda) = \E[Y^2]$ for all $\beta$ in $\R^d$. In this case, the objective $\mathcal{J}$ is constant in $\beta$, and there is no feature learning.
\vspace{.3em} 
\item Assumption~\ref{assumption:mild-assumption} is very mild but necessary for Theorem~\ref{theorem:global-minimizer-three} to hold.
If this assumption is violated, then $\Var(X_i) = 0$ for some $i \in S_*^c$. In this case, 
$K_\beta(X, X') = K(\beta \circ (X - X'))$ is constant in $\beta_i$. By Theorem~\ref{theorem:first-variation-formula}, 
it implies $D \mathcal{J}(\beta, \lambda)[\pm e_i] = 0$ for all $\beta$, and hence $\mathcal{J}$ is constant in $\beta_i$. 
Therefore, the objective $\mathcal{J}$ cannot distinguish this variable $X_i$, and it cannot be eliminated.
\vspace{.3em} 
\item Finally, the condition $\lambda \in (0, \lambda_0)$ requires $\lambda$ to be sufficiently small, so that the regularization term $\lambda \|f\|_H^2$ does not overwhelm the data-fitting term $\E[(Y- f(\beta \circ X))^2]$ in the loss function.
\end{itemize} 

\vspace{.8em} 
To summarize, Assumptions~\ref{assumption:existence-of-csfs},~\ref{assumption:for-the-global-minimizer},~\ref{assumption:mild-assumption} are essential, and the condition that $\lambda$ be sufficiently small is also necessary. In contrast, while Assumption~\ref{assumption:independence} is comparable to standard assumptions in the literature, its necessity for Theorem~\ref{theorem:global-minimizer-three} remains unclear and we leave this as an open question.
}

\section{Stationary Points I: Irrelevant Feature Elimination} 
\label{sec:stationary-points-and-irrelevant-feature-elimination} 
The preceding results characterize the statistical behavior of global minimizers, providing conditions under which they eliminate irrelevant features and retain those that are predictive. However, in practice, such global minimizers may be computationally intractable---due to the nonconvexity of the objective, iterative algorithms are typically guaranteed to converge only to stationary points. This raises a fundamental question for feature learning: do stationary points preserve similar statistical structure, or are such guarantees unique to global optima?

Unlike global minimizers, stationary points are harder to characterize, and require more delicate analytical techniques. To proceed, we divide the analysis into two parts. This section addresses the following question:

\vspace{.5em} 

\begin{itemize}
\item[($\mathsf{Q3}$)] \emph{Do stationary points eliminate irrelevant features?}
\end{itemize} 

\vspace{.5em} 

We provide a positive result: if the irrelevant features are Gaussian distributed, then every stationary point necessarily suppresses them---that is, any nonzero coordinate of a stationary point must correspond to a relevant feature for predicting the response.

\vspace{.5em}
\begin{assumption} 
\label{assumption:Gaussian-noise}
The irrelevant features $(X_i)_{i \in S_*^c}$ are jointly Gaussian distributed. 
\end{assumption} 

\vspace{.5em} 
\begin{theorem}
\label{theorem:Gaussian-noise} 
Assume Assumptions~\ref{assumption:existence-of-csfs},~\ref{assumption:independence},~\ref{assumption:mild-assumption},~\ref{assumption:Gaussian-noise}. Then, for every $\lambda > 0$, 
\begin{equation*}
	\mathrm{D}\mathcal{J}(\beta, \lambda)[v] \le 0,~~~\text{where}~~v = -\Pi_{S_*^c} \beta. 
\end{equation*}
The inequality is strict if $\mathcal{J}(\beta, \lambda) <  \E[Y^2]$ and $\Pi_{S_*^c} \beta \ne 0$.
\end{theorem} 

\vspace{.5em} 
\begin{remark}
\emph{
Compared to Theorem~\ref{theorem:projection-reduce-function-value}, Theorem~\ref{theorem:Gaussian-noise} provides a sharper conclusion: it identifies a descent direction $v = -\Pi_{S_*^c} \beta$ toward the relevant feature subsets. This direction has a natural interpretation---it removes weight from irrelevant coordinates while leaving the relevant ones unchanged. This refinement relies on the additional Gaussian assumption on the irrelevant features. It is important to note that the Gaussian distribution is the canonical model for noise.
}
\end{remark}

\begin{proof} 
Since the problem is translation-invariant, we can without loss of generality assume 
$\E[X]=0$. 
Let $\xi \in \R^d$ be a Gaussian random vector independent of $(X, Y)$ with
\begin{equation*}
	\xi = (0, \xi'), \quad \xi' \stackrel{d}{=} X_{S_*^c} \sim \normal(0, \Cov(X_{S_*^c})).
\end{equation*} 
Here and throughout, we use $\stackrel{d}{=}$ to denote equality in distribution.

Fix $\beta \in \R^d$. For $s\in [0, 1]$, define the vector $\beta^{(s)} \in \R^d$ by 
\begin{equation*}
	\beta^{(s)} = (\beta_{S_*}, \sqrt{1-s} \,\beta_{S_*^c}).
\end{equation*} 
Writing Hadamard products blockwise, we have
\begin{equation*}
\begin{split} 
	\beta \circ X &= (\beta_{S_*} \circ X_{S_*},\; \beta_{S_*^c} \circ X_{S_*^c}) \\
	\beta^{(s)} \circ X + \sqrt{s}\, \beta\, \circ \,\xi &= (\beta_{S_*} \circ X_{S_*},\; \beta_{S_*^c} \circ (\sqrt{1-s} \, X_{S_*^c} + \sqrt{s}\, \xi'))
\end{split} .
\end{equation*}

The key step relies on a key structural property of \emph{Gaussian distributions}. Since $X_{S_*^c}$ and $\xi'$ are independent mean-zero Gaussian random vectors with the same covariance, we have
\begin{equation}
X_{S_*^c} \stackrel{d}{=} \sqrt{1-s} X_{S_*^c} + \sqrt{s} \xi'.
\end{equation} 
Moreover, since $X_{S_*}$ is independent of both $X_{S_*^c}$ and $\xi'$, it follows that
\begin{equation}
\label{eqn:equality-in-distribution}
	(X_{S_*}, \,\beta \circ X) \stackrel{d}{=} (X_{S_*}, \,\beta^{(s)} \circ X + \sqrt{s}\, \beta \,\circ\, \xi).
\end{equation} 

Let $\pi_s$ denote the probability measure of the random vector $\sqrt{s} \, \beta_{S_*^c} \,\circ\, \xi' \in \R^{|S_*^c|}$. 
For every $f \in H$ and $s > 0$, we define $P_s$ as the convolution type operator: 
\begin{equation*}
\begin{split} 
	(P_s f)(x) &= \E_\xi \big[ f(x + \sqrt{s} \, \beta \circ \,\xi) \big] \\
	&= \E_{\xi'} \big[ f(x_{S_*}, x_{S_*^c} + \sqrt{s} \, \beta_{S_*^c} \circ \xi') \big] \\
		&= \int f((x_{S_*}, x_{S_*^c}+ y)) \pi_s(dy).
\end{split} 
\end{equation*} 
For every such function $f$, we use the independence of $\xi$ and $(X, Y)$ to get 
\begin{equation*}
	Y - (P_s f)(\beta^{(s)} \circ X) = \E [Y- f(\beta^{(s)} \circ X+ \sqrt{s} \, \beta \circ \xi)|X, Y ].
\end{equation*} 
By Jensen's inequality and tower property, we further get  
\begin{equation}
\label{eqn:first-step-bounds}
\begin{split} 
	\E[(Y - (P_s f)(\beta^{(s)} \circ X))^2] &= \E [(\E [Y- f(\beta^{(s)} \circ X+ \sqrt{s} \, \beta \circ \, \xi)|X, Y ])^2] \\
				&\le \E[ \E[(Y- f(\beta^{(s)} \circ X+ \sqrt{s} \, \beta \circ \, \xi))^2 |X, Y]] \\
				&= \E[(Y- f(\beta^{(s)} \circ X+ \sqrt{s} \, \beta \circ \, \xi))^2 ].
\end{split} 
\end{equation} 
We evaluate the right-hand-side of the equation. 
By the equality in distribution in~\eqref{eqn:equality-in-distribution}, 
\begin{equation*}
	\E[(\E[Y|X_{S_*}]- f(\beta^{(s)} \circ X+ \sqrt{s} \, \beta \circ \xi))^2] =  \E[(\E[Y|X_{S_*}]- f(\beta \circ X))^2].
\end{equation*}
Using that $\E[Y|X] = \E[Y|X_{S_*}]$, and adding $\E[(Y-\E[Y|X])^2]$ on both sides, we get for all $f$, 
\begin{equation}
\label{eqn:second-step-bounds}
\begin{split} 
 \E[(Y- f(\beta^{(s)} \circ X+ \sqrt{s} \, \beta \circ \xi))^2] 	&= \E[(Y - f(\beta \circ X))^2].
\end{split}
\end{equation} 
Combining~\eqref{eqn:first-step-bounds} and~\eqref{eqn:second-step-bounds}, we have established for every $f \in H$,
\begin{equation}
\label{eqn:summary-of-first-part}
	\E[(Y - (P_s f)(\beta^{(s)} \circ X))^2]	\le  \E[(Y - f(\beta \circ X))^2].
\end{equation}

Additionally, since $P_s f$ is a convolution type operator, we can compute its Fourier transform as follows. 
We define the Fourier transform of the measure $\pi_s$ as
\begin{equation*}
	 \widehat{ \pi}_s (\eta):= \int e^{-2 \pi i\langle \eta, y\rangle} \pi_s(dy).
\end{equation*}
Then,
\begin{equation*}
\begin{split}
	\widehat {P_s f} (\omega) &= \iint f((x_{S_*}, x_{S_*^c} + y))  e^{-2 \pi i\langle x, \omega\rangle} dx\, \pi_s(dy) \\
	&= \int \pi_s(dy)  \iint f((x_{S_*}, x_{S_*^c} + y))e^{-2 \pi i\langle x, \omega\rangle} dx_{S_*} dx_{S_*^c} \\
	&= \int e^{2\pi i\langle \omega_{S_*^c}, y\rangle}  \pi_s(dy)  \cdot \widehat{f}(\omega) 
	 = \widehat{f}(\omega) \cdot \widehat{ \pi}_s (-\omega_{S_*^c}).
\end{split} 
\end{equation*}
Because $\|\widehat{\pi}_s\|_{L_\infty} \le \int \pi_s(dy) = 1$, this immediately implies 
$|\widehat {P_s f}|(\omega) \le |\widehat f|(\omega)$ for all $\omega$. Thus
\begin{equation}
\label{eqn:summary-of-second-part}
	\norm{P_s f}_H^2 = \int \frac{|\widehat {P_s f}|(\omega)^2}{k(\omega)} d\omega
		\le \int \frac{|\widehat {f} |(\omega)^2}{k(\omega)} d\omega = \norm{f}_H^2.
\end{equation}  
Hence, $P_s f \in H$. 
Combining~\eqref{eqn:summary-of-first-part} and~\eqref{eqn:summary-of-second-part}, we have shown for every $f \in H$, 
\begin{equation*}
	\E[(Y - (P_s f)(\beta^{(s)} \circ X))^2] + \lambda 	\norm{P_s f}_H^2
		\le  \E[(Y - f(\beta \circ X))^2] + \lambda \norm{f}_H^2.
\end{equation*}
Thus, taking infimum over $f \in H$ on both sides yields 
\begin{equation}
\label{eqn:J-decrease}
	\mathcal{J}(\beta^{(s)}, \lambda) \le \mathcal{J}(\beta, \lambda).
\end{equation} 
Let $v := -\Pi_{S_*^c} \beta \in \R^d$. Since $\beta^{(s)} = \beta + \half s v + o(|s|)$ as $s \to 0^+$, this yields 
\begin{equation*}
	\mathrm{D}\mathcal{J}(\beta, \lambda)[v] =  \lim_{s \to 0^+} \frac{2}{s} ( \mathcal{J}(\beta^{(s)}, \lambda) - \mathcal{J}(\beta, \lambda)) \le 0
\end{equation*} 
as desired. 
 
We show the above inequality is strict when $\mathcal{J}(\beta, \lambda) < \E[Y^2]$ and $\Pi_{S_*^c} \beta \ne 0$. Let $\Sigma := \Cov(X_{S_*^c})$. We compute the Fourier transform of the Gaussian measure $\pi_s$ and get 
\begin{equation*}
	\widehat \pi_s (\omega_{S_*^c}) = e^{-2\pi^2  s\|(\beta \circ \omega)_{S_*^c}\|^2_\Sigma}\,\,~~~\text{where}~~~\|\eta\|^2_\Sigma := \eta^\top \Sigma \eta.
\end{equation*}
Following the above derivation, and using Fatou's lemma, we get for all $f \in H$,
\begin{equation*}
\begin{split} 
	\liminf_{s \to 0^+} \frac{1}{s} (\norm{f}_H^2 - \norm{P_s f}_H^2) 
		&\ge \liminf_{s \to 0^+} \int \frac{|\widehat {f} |(\omega)^2}{k(\omega)} \cdot  \frac{1}{s} (1-|\widehat \pi_s (-\omega_{S_*^c})|^2) d\omega \\
		&\ge  2\pi^2 \cdot \int \frac{|\widehat {f} |(\omega)^2}{k(\omega)} \cdot  \|(\beta \circ \omega)_{S_*^c}\|^2_\Sigma ~d\omega.
\end{split} 
\end{equation*} 
Since 
\begin{equation*}
\begin{split}
	\mathcal{J}(\beta^{(s)}, \lambda) \le \mathcal{I}(\beta^{(s)}, P_s f_{\beta, \lambda}, \lambda) 
		&= \E[(Y - (P_s f_{\beta, \lambda})(\beta^{(s)} \circ X))^2] + \lambda \norm{P_s f_{\beta, \lambda}}_H^2 \\
	\mathcal{J}(\beta, \lambda) = \mathcal{I}(\beta,  f_{\beta, \lambda}, \lambda) 
		&= \E[(Y - f_{\beta, \lambda}(\beta \circ X))^2] + \lambda \norm{f_{\beta, \lambda} }_H^2
\end{split} 
\end{equation*} 
and since by~\eqref{eqn:summary-of-first-part}
\begin{equation*}
	\E[(Y - (P_s f_{\beta, \lambda})(\beta^{(s)} \circ X))^2]	\le  \E[(Y - f_{\beta, \lambda}(\beta \circ X))^2]
\end{equation*} 
we get 
\begin{equation*}
\begin{split} 
	\mathrm{D}\mathcal{J}(\beta, \lambda)[v] &= 
		\lim_{s \to 0^+} \frac{2}{s} ( \mathcal{J}(\beta^{(s)}, \lambda) - \mathcal{J}(\beta, \lambda))) \\
		&\le \lambda \cdot \limsup_{s \to 0^+} \frac{2}{s} ( \norm{P_s f_{\beta, \lambda}}_H^2 - \norm{f_{\beta, \lambda}}_H^2) \\
		&\le -4\pi^2 \lambda \cdot  \int \frac{|\widehat {f}_{\beta, \lambda} |(\omega)^2}{k(\omega)} \cdot  \|(\beta \circ \omega)_{S_*^c}\|^2_\Sigma d\omega.
\end{split} 
\end{equation*} 
The right-hand-side vanishes if and only if
\begin{equation*}
	\frac{|\widehat {f}_{\beta, \lambda} |(\omega)^2}{k(\omega)} \cdot  \|(\beta \circ \omega)_{S_*^c}\|^2_\Sigma = 0~~~\text{a.e. $\omega$}.
\end{equation*} 
Since $\Sigma $ has positive diagonal entries, this happens if and only if: (i) either $\beta_{S_*^c} = 0$, so that $\Pi_{S_*^c} \beta = 0$; or (ii) 
$\widehat{f}_{\beta, \lambda}(\omega) = 0$ for almost every $\omega$, in which case $f_{\beta, \lambda} \equiv 0$ and hence $\mathcal{J}(\beta, \lambda) = \mathbb{E}[Y^2]$.
\end{proof} 

\redblock{
\vspace{.5em} 
\begin{remark}\label{remark:gaussian-noise-identity} \emph{
We discuss the role of the Gaussian noise assumption (Assumption~\ref{assumption:Gaussian-noise}) in Theorem~\ref{theorem:Gaussian-noise}. 
The identity~\eqref{eqn:equality-in-distribution} states that increasing the weight on the irrelevant features $S_*^c$ (i.e. moving from $\beta^{(s)}$ to $\beta$) is equivalent, in distribution, to injecting \emph{independent} Gaussian noise $\xi$ into $X$; in other words, $\beta \circ X$ has the same distribution as $\beta^{(s)} \circ X + \sqrt{s}\,\beta \circ \xi$.
Since $Y$ depends only on $X$, this introduces only noise and degrades the signal-to-noise ratio. 
The proof formalizes this intuition, and shows $\mathcal{J}(\beta^{(s)}, \lambda) \le \mathcal{J}(\beta, \lambda)$ in~\eqref{eqn:J-decrease}, with conditions for strict inequality discussed below~\eqref{eqn:J-decrease}. }
\end{remark}

\vspace{.5em} 
\begin{remark}\label{remark:gaussian-extension-open}\emph{
We note that the Gaussian assumption (Assumption~\ref{assumption:Gaussian-noise}) is not essential for Theorem~\ref{theorem:Gaussian-noise} to hold. 
In fact, one can construct specific examples showing that the conclusion continues to hold in certain non-Gaussian settings.  However, we currently lack a general extension of Theorem~\ref{theorem:Gaussian-noise} to broader classes of distributions, and such an extension is left open.
}
\end{remark} 
}

\vspace{.5em} 
As an immediate consequence of Theorem~\ref{theorem:Gaussian-noise}, we get
\vspace{.5em} 
\begin{corollary}
\label{corollary:elimination-at-stationary-points}
Assume Assumptions~\ref{assumption:existence-of-csfs},~\ref{assumption:independence},~\ref{assumption:mild-assumption},~\ref{assumption:Gaussian-noise}. 
Let $\lambda > 0$.
If $\beta_\dagger$ is a directional stationary point of $\mathcal{J}(\cdot, \lambda)$, then one of the following must hold: 
\begin{equation*}
	(i)~~\Pi_{S_*^c}  \beta_\dagger = 0~~\qquad \text{or} \qquad 
	(ii)~~\mathcal{J}(\beta_\dagger, \lambda) = \E[Y^2]. 
\end{equation*}
\end{corollary} 

We remark that $\mathcal{J}(\beta_\dagger, \lambda) = \E[Y^2]$ is equivalent to 
$\E[Y|\beta_\dagger \circ X]= 0$ (Lemma~\ref{lemma:trivial-bound-on-r-f}).

\vspace{1em} 
\begin{remark}
\emph{
Compared to Theorem~\ref{theorem:global-minimizer}, which concerns global minimizers of $\mathcal{J}$, Corollary~\ref{corollary:elimination-at-stationary-points} shows that a similar feature elimination property holds more broadly at directional stationary points---under an additional Gaussian assumption on the irrelevant features $X_{S_*^c}$. In particular, any stationary point that achieves nontrivial prediction error must lie entirely within the relevant feature subspace. This extends the statistical interpretation of feature sparsity beyond the global minimizers.
}
\end{remark}

\section{Stationary Points II: Relevant Feature Recovery} 
\label{sec:stationary-points-and-relevant-feature-recovery} 
In this section, we continue our analysis of stationary points of $\mathcal{J}$, turning to the feature recovery question complementary to $(\mathsf{Q3})$:
\vspace{.5em}
\begin{itemize}
\item[($\mathsf{Q4}$)] Can stationary points capture at least some---or all features---in $S_*$? 
\end{itemize} 

\vspace{.5em}
The answer to $(\mathsf{Q4})$ is subtle and depends not only on the data distribution but, more crucially, on the choice of kernel $\kernel$. 
Our analysis shows that $\ell_1$-type RKHS (Example~\ref{example:ell-1-type-RKHS}) are particularly effective for this purpose, enabling recovery of relevant features under milder distributional assumptions than commonly used radial kernel RKHS (Example~\ref{example:radial-kernel-RKHS}) such as the Gaussian RKHS. We outline the underlying intuition in Section~\ref{sec:laplace-kernel-why}, and briefly recall the kernel class here.

Specifically, we focus on the following class of \emph{$\ell_1$-type} translation-invariant kernels:
\begin{equation}
\label{eqn:choice-of-ell-one-kernels}
	\kernel(x, x') = \profile( \norm{x-x'}_1),~~~\text{where}~~~\profile(z) = \int_0^\infty e^{-tz} \mu(dt)
\end{equation} 
where $\mu$ is a probability measure whose support is compact, and contained in $(0, \infty)$. The Laplace kernel arises from the Dirac measure at $z=1$, i.e. $\mu = \delta_1$, in which case $\profile(t) = e^{-t}$. This will serve as our canonical example. 
In contrast, the Gaussian kernel does not belong to this class.

All main results in this section focus on the $\ell_1$-type kernel class, addressing $(\mathsf{Q4})$ under various distributional assumptions. We assume $\E[Y] = 0$ throughout this section.

\vspace{.5em} 
\begin{remark} 
\emph{The assumption $\E[Y] = 0$ is without loss of generality. Otherwise, one may subtract the mean from $Y$ before regression or, equivalently, reformulate the problem with an intercept term $\gamma$: 
\begin{equation*}
	\min_\beta \min_f \min_\gamma \E[(Y -f(\beta \circ X) - \gamma)^2] + \lambda \norm{f}_H^2.
\end{equation*}
The objective $\mathcal{J}(\beta, \lambda)$ can then be defined accordingly, and all subsequent results continue to hold.
}
\end{remark} 

\subsection{Why the Laplace Kernel? A First Indication} 
\label{sec:laplace-kernel-why}
We aim to understand when stationary points of $\mathcal{J}$ recover the relevant features as posed in $(\mathsf{Q4})$.  As it turns out, the answer depends crucially on the choice of kernel $\kernel$. 

To build intuition, we consider a simpler diagnostic question:

\vspace{0.5em}
\begin{itemize}
\item[$\mathsf{(Q5)}$]  Suppose the true feature set $S_* \neq \emptyset$. (i) When is $\beta = 0$ not stationary? (ii) When does every stationary point $\beta$ of $\mathcal{J}$ satisfy $\supp(\beta) \cap S_* \neq \emptyset$?
\end{itemize} 

\vspace{0.5em}
This question $(\mathsf{Q5})$ serves as a basic test of whether the kernel supports feature detection. Question (i) is the most basic: it asks whether the trivial solution $\beta = 0$ can be ruled out a stationary point. Question (ii) is stronger---it asks whether all stationary points capture at least one relevant feature.  The two questions in $(\mathsf{Q5})$ are in fact equivalent under the additional assumptions of Theorem~\ref{theorem:Gaussian-noise}, since Theorem~\ref{theorem:Gaussian-noise} ensures that every stationary point eliminates irrelevant features.

Perhaps surprisingly, the Laplace kernel---and more broadly, the class of $\ell_1$-type kernels---passes this test under much weaker assumptions than the commonly used Gaussian kernel. 
We start by presenting the positive guarantees shared by all the $\ell_1$ type kernels. 

\vspace{0.5em} 
\begin{theorem}
\label{proposition:why-Laplace-kernel-is-good}
Suppose the $\ell_1$-type kernel is used in the objective $\mathcal{J}$. Then the directional derivative 
at $\beta = 0$ along the direction $e_i$ for every $1 \le i \le d$ satisfies 
\begin{equation*}
	\mathrm{D}\mathcal{J}(0, \lambda)[e_i]
		= \frac{1}{\lambda} \cdot \profile^\prime(0) \cdot \int_\R \left|\E[Y e^{-2\pi i \zeta X_i}]\right|^2 \frac{1}{2\pi^2 \zeta^2} d\zeta. 
\end{equation*}
As a result, $\mathrm{D}\mathcal{J}(0, \lambda)[e_i] < 0$ if and only if $\E[Y|X_i] \neq 0$. 
\end{theorem} 

\vspace{.7em} 
The key to Theorem~\ref{proposition:why-Laplace-kernel-is-good} lies in the structure of the $\ell_1$-type kernel, which induces a bivariate map $(x, x') \mapsto |x - x'|$ on $\R \times \R$. This map is classically known to be \emph{conditionally negative definite}, a fact we make precise in Lemma~\ref{lemma:absolute-negative-definite} via a Fourier representation. The general definition of conditional negative definite kernel can be found in the monograph~\cite{Wendland04}, though this definition is not needed for our purpose; the lemma provides all that is required for our analysis.

\vspace{.5em} 
\begin{lemma}
\label{lemma:absolute-negative-definite} 
Let $\nu$ be a probability measure on $\mathbb{R}$, and let $f \in L^2(\nu)$ be complex-valued, satisfying the mean-zero condition
\[
\int_{\mathbb{R}} f(x) \, d\nu(x) = 0.
\]
Suppose $\int x^2 \nu(dx) < \infty$. Then 
\[
    \iint_{\mathbb{R}^2} f(x) \overline{f(x')} \, |x - x'| \, d\nu(x) \, d\nu(x')
    = - \int_{\mathbb{R}} \left| \int_{\mathbb{R}} f(x) e^{-2 \pi i \zeta x} \, d\nu(x) \right|^2 \frac{1}{2\pi^2 \zeta^2} \, d \zeta.
\]
As a result, the left-hand side is always nonpositive, and is strictly negative whenever $f \not\equiv 0$ $\nu$-a.e.
\end{lemma}

\begin{proof}
For each $s > 0$, the function $z \mapsto e^{-s|z|}$ has the Fourier inversion representation 
\[
    e^{-s|z|} = \int_{\mathbb{R}} e^{-2\pi i \zeta z} m_s(\zeta) \, d\zeta, \quad 
    m_s(\zeta) := \frac{2s}{4\pi^2 \zeta^2 + s^2}.
\]
Applying this to the kernel $e^{-s|x - x'|}$ and using Fubini's theorem, we get
\begin{equation*}
\begin{split} 
    \iint f(x) \overline{f(x')} e^{-s|x - x'|} \, d\nu(x) d\nu(x') 
    &= \int \left| \int f(x) e^{-2\pi i \zeta x} \, d\nu(x) \right|^2 m_s(\zeta) \, d\zeta.
\end{split} 
\end{equation*}
Since $|x-x'| = \lim_{s \to 0^+} \frac{1}{s} (1-e^{-s|x-x'|})$, with the uniform bound $|\frac{1}{s} (1-e^{-s|x-x'|})| \le |x-x'|$
for all $s > 0$,  the Lebesgue's dominated convergence theorem applies:
\begin{equation*}
\begin{split} 
    \iint f(x) \overline{f(x')} |x - x'| \, d\nu(x) d\nu(x')
   & = \lim_{s \to 0^+} \frac{1}{s} \iint f(x) \overline{f(x')} \big(1 - e^{-s|x - x'|} \big) \, d\nu(x) d\nu(x').
\end{split} 
\end{equation*}
Since $\int f \, d\nu = 0$, the integral of the constant term vanishes, so we have
\begin{equation*}
\begin{split} 
  \iint f(x) \overline{f(x')} |x - x'| \, d\nu(x) d\nu(x') &= -\lim_{s \to 0^+} \frac{1}{s} \int \left| \int f(x) e^{-2\pi i \zeta x} \, d\nu(x) \right|^2 m_s(\zeta) \, d\zeta \\
&= -\int \left| \int f(x) e^{-2\pi i \zeta x} \, d\nu(x) \right|^2 \frac{1}{2\pi^2 \zeta^2} \, d\zeta
\end{split} 
\end{equation*} 
where the last equality follows from monotone convergence, since $m_s(\zeta)/s \uparrow \frac{1}{2\pi^2 \zeta^2}$ as $s \to 0^+$. 

This completes the proof. 
\end{proof}

\begin{proof}[Proof of Theorem~\ref{proposition:why-Laplace-kernel-is-good}] 
We apply the first variation formula of $\mathcal{J}$. By definition 
\begin{equation*}
\begin{split} 
	\mathrm{D} \kernel_0(x, x')[e_i] &= \lim_{s \to 0^+} \frac{1}{s} (\kernel_{se_i} - \kernel_{0})(x, x') \\
		&= \lim_{s \to 0^+} \frac{1}{s} (\profile(s |x_i - x_i^\prime|) - \profile(0)) = \profile^\prime(0)|x_i - x_i'|.
\end{split} 
\end{equation*}
Since $\E[Y] = 0$, $f_{0, \lambda}$ is the zero function. Thus, $r_{0, \lambda}(x, y) = y - f_{0, \lambda}(0) = y$. Therefore, 
\begin{equation*}
\begin{split} 
	\mathrm{D}\mathcal{J}(0, \lambda)[e_i] &= -\frac{1}{\lambda}
		\E[r_{0, \lambda}(X, Y) r_{0, \lambda}(X', Y') \mathrm{D} \kernel_0(X, X')[e_i] ] = -\frac{1}{\lambda} \profile^\prime(0) \cdot \E[YY' |X_i - X_i'|]. 
\end{split}
\end{equation*}
Let $f_i(x_i) = \E[Y|X_i = x_i]$. Then $\E[f_i(X_i)] = \E[Y] = 0$.  Applying Lemma~\ref{lemma:absolute-negative-definite}, we get  
\begin{equation*}
\begin{split} 
	\E[YY'|X_i - X_i'|] &= \E[f_i(X_i) f_i(X_i') |X_i - X_i'|] \\
		&= - \int_\R \left|\E[f_i(X_i) e^{-2\pi i\zeta X_i}]\right|^2 \frac{1}{2\pi^2 \zeta^2} d\zeta \\
		&=  -\int_\R \left|\E[Y e^{-2\pi i\zeta X_i}]\right|^2 \frac{1}{2\pi^2 \zeta^2} d\zeta 
\end{split} 
\end{equation*} 
In particular, $\E[YY'|X_i - X_i'|] < 0$ since $f_i \neq 0$ under $\P$. Finally, we note $
	\profile^\prime(0) = -\int_0^\infty t \mu(dt)$.   
Since $\mu$ is not concentrated at $0$, we get  $\profile^\prime(0) < 0$.  As a result, we obtain 
\[
	\mathrm{D}\mathcal{J}(0, \lambda)[e_i] = \frac{1}{\lambda} \profile'(0) \cdot \int_\R \left|\E[Y e^{-2\pi i \zeta X_i}]\right|^2 \frac{1}{2\pi^2 \zeta^2} d\zeta < 0.
\]
This completes the proof.
\end{proof} 

An immediate consequence of Theorem~\ref{proposition:why-Laplace-kernel-is-good} is the following corollary.

\vspace{.5em}  
\begin{corollary}
\label{corollary:beta=zero-not-stationary-ell-one}
For $\ell_1$ type kernel, $\beta = 0$ is stationary for $\mathcal{J}$ if and only if $\E[Y|X_i] =0$ for all $i$. 
\end{corollary} 

\vspace{.5em}  

Our analysis shows that $\ell_1$-type kernels are sensitive to marginal signals of \emph{any kind}: at $\beta = 0$, the objective is not stationary whenever some feature has a nonzero \emph{main effect}, i.e., $\E[Y|X_i] \ne 0$. At a fundamental level, this behavior is driven by the \emph{conditional negative definiteness} of the map $(x, x') \mapsto |x-x'|$ on $\R^2$ (Lemma~\ref{lemma:absolute-negative-definite}), which appears in the derivative of $\mathcal{J}$ at $\beta = 0$ (see Theorem~\ref{proposition:why-Laplace-kernel-is-good}). By contrast, radial basis kernels (e.g., Gaussian kernel) are far more restrictive. As we will see in Section~\ref{sec:comparison-to-radial-basis-kernels}, they only respond to marginal signals of \emph{linear type}, i.e., when $\E[Y X_i] \ne 0$. 

On the downside, when the signal is nontrivial but lacks main effects---meaning $\E[Y|X_i] = 0$ for all $i$ but $\E[Y|X] \ne 0$---the objective $\mathcal{J}(\beta, \lambda)$ has vanishing directional derivatives at $\beta = 0$, and thus provides no first-order information for identifying relevant features. In such cases, even $\ell_1$-type kernels are powerless at $\beta = 0$. This highlights a  limitation of first-order analysis: detecting such higher-order signals likely requires exploiting higher-order variation in $\mathcal{J}$, a problem we leave open.

\subsubsection{Comparison to Radial Basis Kernels} 
\label{sec:comparison-to-radial-basis-kernels}
Our analysis focuses on $\ell_1$-type kernels because they are sensitive to marginal signals of all type. To place this in context, we contrast them with radial basis kernels (e.g., Gaussian), which exhibit more selective behavior at $\beta = 0$: they respond only to marginal signals of \emph{linear type}, as we show below. 

Consider 
\begin{equation*}
	\kernel(x, x') = \profile( \norm{x-x'}_2^2),~~~\text{where}~~~\profile(z) = \int_0^\infty e^{-tz} \mu(dt)
\end{equation*} 
for some nonnegative probability measure $\mu$ with compact support in $(0, \infty)$. This class includes the Gaussian kernel as a canonical example. 

We assume $X$ has finite fourth moments in this subsection~\ref{sec:comparison-to-radial-basis-kernels}.
\vspace{.5em} 
\begin{lemma}
\label{lemma:limitation-of-Gaussian}
Suppose the radial basis kernel is used in the objective $\mathcal{J}$. Then the directional derivative 
at $\beta = 0$ along the direction $e_i$ for every $1 \le i \le d$ is zero, i.e., $\mathrm{D} \mathcal{J}(0, \lambda)[e_i] = 0$ for all $i$. Moreover,
\begin{equation*}
	\lim_{s \to 0^+} \frac{1}{s^2} \left(\mathcal{J}(s e_i, \lambda) - \mathcal{J}(0, \lambda)\right) 
		=  \frac{1}{\lambda} \cdot \profile^\prime(0) \cdot (\E[YX_i])^2 \le 0. 
\end{equation*}
The right-hand side is strictly negative if and only if $\mathbb{E}[Y X_i] \ne 0$.
\end{lemma} 

\begin{proof} 
Fix $i$ and define 
\begin{equation*}
G_i(s):=-\frac{1}{\lambda} \cdot \E[r_{s e_i, \lambda}(X, Y)r_{s e_i, \lambda}(X', Y') \profile^\prime(s^2|X_i - X_i'|^2)|X_i -X_i'|^2].
\end{equation*} 
By Theorem~\ref{theorem:first-variation-formula}, for every $s$, the directional derivative at $s e_i$ along $e_i$ exists and satisfies
\begin{equation*}
\mathrm{D}\mathcal{J}(s e_i, \lambda)[e_i] = s G_i(s).
\end{equation*}
In particular, $\mathrm{D}\mathcal{J}(0, \lambda)[e_i] = 0$ for all $1\le i \le d$. 

By Lemma~\ref{lemma:continuity-of-J}, the map $s \mapsto r_{s e_i, \lambda}(X,Y)$ is continuous in $L^2(\P)$. Since $\profile'$ is continuous 
and bounded, and $X$ has finite fourth moments, it follows that $G_i$ is continuous by the dominated convergence theorem. Therefore, we have
\begin{equation*}
	\mathcal{J}(s e_i, \lambda) - \mathcal{J}(0, \lambda) = \int_0^s \mathrm{D}\mathcal{J}(t e_i, \lambda)[e_i] dt  
		= \int_0^s t G_i(t) dt = \frac{s^2}{2} G_i(0) + o(s^2)~~ (s \to 0^+).
\end{equation*} 
This yields
\begin{equation*}
	\lim_{s \to 0^+} \frac{1}{s^2} \left(\mathcal{J}(s e_i, \lambda) - \mathcal{J}(0, \lambda)\right)  = \half G_i(0).
\end{equation*} 
It remains to compute $G_i(0)$.
Since $\E[Y] = 0$, the minimizing function at $\beta = 0$ is $f_{0, \lambda} \equiv 0$, the zero function, and hence $r_{0, \lambda}(X, Y) = Y$.
Substituting into the definition of $G_i(t)$ gives
\begin{equation*}
	G_i(0) = -\frac{2}{\lambda} \profile^\prime(0) \cdot  \E[YY'|X_i - X_i'|^2] = \frac{2}{\lambda} \profile^\prime(0) \cdot (\E[YX_i])^2.
\end{equation*}
The claim follows by combining the last two identities.
\end{proof} 

Radial basis kernels suffer from a fundamental limitation: at $\beta = 0$, they detect only marginal signals of \emph{linear type} (i.e., when $\E[YX_i] \ne 0$) and ignore other nonlinear main effects. The lemma formalizes this: $\mathcal{J}(\beta,\lambda)$ is second-order flat at $\beta = 0$, with all first-order directional derivatives vanishing, and the leading second-order term depending only on $\E[YX_i]$. This restriction persists under reparameterization. For example, setting $\beta_i = \sqrt{|\theta_i|}$ and defining $\tilde{\mathcal{J}}(\theta, \lambda) := \mathcal{J}(\sqrt{|\theta|}, \lambda)$ yields
\begin{equation*}
	\tilde{J}(se_i, \lambda) -\tilde{J}(0, \lambda) = -\frac{s}{\lambda} \profile^\prime(0) \E[YX_i]^2 + o(|s|)~~~(s \to 0^+)
\end{equation*}
so the objective varies at first order but still only through $\E[YX_i]$.

Thus the dependence on linear correlations is intrinsic, and radial kernels cannot reveal nonlinear main effects through its leading-order behavior near zero. This motivates our focus on $\ell_1$-type kernels.

%
%
%
%
%
%

\subsection{A Decomposition of Directional Derivative} 
\label{sec:decomposition-of-derivative}
In this subsection, we analyze the directional derivative
\begin{equation*}
	\mathrm{D}\mathcal{J}(\beta, \lambda)[e_i]
\end{equation*} 
at a general $\beta$ along the coordinate directions $e_i$. We restrict attention to $\ell_1$-type kernels and assume $X$ has finite second moment, which guarantees the existence of the derivative (see Example~\ref{example:first-variation-for-ell-one-type-RKHS}).

\subsubsection{An Integral Representation} 
\label{sec:an-integral-representation}
\vspace{0.5em} 
\begin{lemma}
\label{lemma:first-step}
Let $\beta \in \R^d$ satisfy $\beta_i \ge 0$. Then 
\begin{equation*}
	\mathrm{D}\mathcal{J}(\beta, \lambda)[e_i] = 
		-\frac{1}{\lambda} \cdot \E[r_{\beta, \lambda}(X, Y)r_{\beta, \lambda}(X', Y') \profile^\prime(\norm{\beta\circ(X-X')}_{1})|X_i -X_i'|].
\end{equation*} 
\end{lemma} 
\begin{proof} 
With $\beta_i \ge 0$, we get 
\begin{equation*}
\begin{split} 
	\mathrm{D}\kernel_\beta(x,x')[e_i] &= \lim_{s \to 0^+} \frac{1}{s} (\kernel_{\beta + s e_i} (x, x') - \kernel_\beta(x, x')) 
		= \profile^\prime(\norm{\beta \circ (x-x')}_{1}) |x_i - x_i'|.
\end{split} 
\end{equation*}
Substituting this into the first variation formula from Theorem~\ref{theorem:first-variation-formula} gives the result.
\end{proof} 

We recall the definition of~$\profile$ from~\eqref{eqn:choice-of-ell-one-kernels}.
The function $\profile$ and its derivative $\profile^\prime$ satisfy
\begin{equation*}
	\profile(z) = \int_0^\infty e^{-t z} \, \mu(dt),~~~\quad~~~\profile^\prime(z) =  -\int_0^\infty t\, e^{-t z} \, \mu(dt).
\end{equation*}
For each $t > 0$, using the Fourier inversion formula, we get 
\begin{equation*}
e^{-t\|x\|_1} = \int_{\R^d} e^{-2\pi i \langle \omega, x \rangle} \, q_t(\omega) \, d\omega,
\quad \text{where} \quad 
q_t(\omega) := \prod_{j=1}^d \frac{2t}{4\pi^2 \omega_j^2 + t^2}.
\end{equation*} 
Combining the two, we obtain for every $x$ in $\R^d$:
\begin{equation}
\label{eqn:computation-of-psi-psi-prime}
\begin{split} 
\profile \big(\|x\|_1\big)
&= \int_0^\infty \int_{\R^d} 
e^{-2\pi i \langle \omega, x \rangle} 
\, q_t(\omega) \, d\omega \, \mu(dt) \\
\profile^\prime\big(\|x\|_1\big)
&= -\int_0^\infty \int_{\R^d} 
e^{-2\pi i \langle \omega, x \rangle} 
\, tq_t(\omega) \, d\omega \, \mu(dt).
\end{split} 
\end{equation} 
These formulas yield an integral representation for the directional derivative $\mathrm{D}\mathcal{J}(\cdot)[e_i]$.

\vspace{.5em} 
\begin{lemma}
\label{lemma:integral-representation-of-derivative}
Let $\beta \in \R^d$ with $\beta_i \ge 0$. Then
\[
\mathrm{D}\mathcal{J}(\beta, \lambda)[e_i] = 
\frac{1}{\lambda} \int_0^\infty \int_{\R^d} 
\E\left[ r_{\beta, \lambda}(X, Y) r_{\beta, \lambda}(X', Y') \, 
e^{-2\pi i \langle \omega, \beta \circ (X - X') \rangle} 
\, |X_i - X_i'| \right] 
\cdot t q_t(\omega) \, d\omega  \mu(dt).
\]
\end{lemma}

\begin{proof}
By Lemma~\ref{lemma:first-step}, $\mathrm{D}\mathcal{J}(\beta,\lambda)[e_i]$ is expressed in terms of $\profile'(\norm{\beta \circ (X-X')}_{1})$. By~\eqref{eqn:computation-of-psi-psi-prime}, we have
\begin{equation*}
\profile^\prime\big(\|\beta \circ (x-x')\|_1\big)
= -\int_0^\infty \int_{\R^d} 
e^{-2\pi i \langle \omega, \beta \circ (x-x') \rangle} 
\, tq_t(\omega) \, d\omega \, \mu(dt).
\end{equation*}
Substituting this representation into the expression for $\mathrm{D}\mathcal{J}(\beta,\lambda)[e_i]$ in Lemma~\ref{lemma:first-step} and applying Fubini’s theorem to exchange the order of expectation and integration yields the desired formula.
\end{proof}

\subsubsection{A Proxy for Directional Derivative} 
\label{sec:proxy-for-directional-derivative}
In Section~\ref{sec:laplace-kernel-why}, we show that $\ell_1$-type kernels prevent $\beta = 0$ from being stationary whenever there is a nontrivial main effect feature. This conclusion relies on a key fact: the bivariate function $(x, x') \mapsto |x - x'|$ is \emph{conditionally negative definite}, which forces the directional derivative $\mathrm{D}\mathcal{J}(0, \lambda)[e_i]$ to be strictly negative whenever $Y$ depends marginally on some coordinate $X_i$. This is made precise via the Fourier representation of $|x - x'|$ in Lemma~\ref{lemma:absolute-negative-definite}.

This raises a natural question: does the same mechanism extend to general $\beta \ne 0$?
To pursue this, we aim to expand the conditionally negative definite kernel $(x, x') \mapsto |x - x'|$ 
within the directional derivative formula. Let us define  
\begin{equation*}
R_{\beta, \omega, \lambda} (X, Y) := r_{\beta, \lambda}(X, Y) e^{-2\pi i \langle \omega, \beta \circ X\rangle}.
\end{equation*}
Our earlier formula of directional derivative (Lemma~\ref{lemma:integral-representation-of-derivative}) shows
\begin{equation*}
\begin{split} 
\mathrm{D}\mathcal{J}(\beta, \lambda)[e_i] 
&=
\frac{1}{\lambda} \int_0^\infty \int_{\R^d} 
\E\left[ R_{\beta, \omega, \lambda}(X, Y) \overline{R_{\beta, \omega, \lambda}(X', Y')} \,
\, |X_i - X_i'| \right] 
\cdot t q_t(\omega) \, d\omega  \mu(dt).
\end{split} 
\end{equation*} 
To pursue a Fourier-based representation, we are led to consider the quadratic form
\begin{equation*}
	\E\left[ R_{\beta, \omega, \lambda}(X, Y) \overline{R_{\beta, \omega, \lambda}(X', Y')}|X_i - X_i'|\right]
\end{equation*}
with the aim of applying Lemma~\ref{lemma:absolute-negative-definite}. However, a key difficulty 
arises: the lemma crucially relies on the mean-zero condition, which does not hold in 
our setting---$\E[ R_{\beta, \omega, \lambda}(X, Y) ] \ne 0$ in general.

To overcome this obstacle, we work with a centered integrand:
\begin{equation*}
	S_{\beta, \omega, \lambda}(x, y) = R_{\beta, \omega, \lambda}(x, y) - \E[R_{\beta, \omega, \lambda}(X, Y)].
\end{equation*}
This leads to the following quantity, which we view as a \emph{proxy} for the directional derivative:
\begin{equation*}
	\mathcal{S}_i (\beta, \lambda) = \frac{1}{\lambda} \int_0^\infty \int_{\R^d} 
\E\left[ S_{\beta, \omega, \lambda}(X, Y) \overline{S_{\beta, \omega, \lambda}(X', Y')}
\, |X_i - X_i'| \right] 
\cdot t q_t(\omega) \, d\omega  \mu(dt).
\end{equation*} 
After centering, this proxy quantity admits a Fourier-type representation.
 
 \vspace{.5em} 
\begin{lemma}
\begin{equation*}
	\mathcal{S}_i (\beta, \lambda) =  -\frac{1}{\lambda} \int_0^\infty \int_{\R^d}
		\int_\R \left|\E[S_{\beta, \omega, \lambda} (X, Y) e^{-2\pi i \zeta X_i }] \right|^2 \frac{1}{2\pi^2 \zeta^2} d\zeta \cdot t q_t(\omega) \, d\omega  \mu(dt).
\end{equation*} 
\end{lemma} 
\begin{proof} 
Let us define
$
T_{\beta, \omega, \lambda}(x_i) := \E[S_{\beta, \omega, \lambda}(X, Y) \mid X_i = x_i]
$.
Then 
\[
	\E[T_{\beta, \omega, \lambda}(X_i)] = \E[S_{\beta, \omega, \lambda}(X, Y)] = 0.
\]
We now compute
\begin{equation*}
\begin{split} 
	\E\left[ S_{\beta, \omega, \lambda}(X, Y) \overline{S_{\beta, \omega, \lambda}(X', Y')}
\, |X_i - X_i'| \right] 
&= \E\left[ T_{\beta, \omega, \lambda}(X_i) \overline{T_{\beta, \omega, \lambda}(X_i')}
\, |X_i - X_i'| \right]  \\
&= -\int_\R \left|\E[T_{\beta, \omega, \lambda} (X_i) e^{-2\pi i \zeta X_i }] \right|^2 \frac{1}{2\pi^2 \zeta^2} d\zeta \\
&= -\int_\R \left|\E[S_{\beta, \omega, \lambda} (X, Y) e^{-2\pi i \zeta X_i }] \right|^2 \frac{1}{2\pi^2 \zeta^2} d\zeta.
\end{split} 
\end{equation*} 
The first and third equality follow from the tower property of conditional expectation, and the second from Lemma~\ref{lemma:absolute-negative-definite}, noting that $T$ satisfies the mean-zero condition.
\end{proof} 
Finally, using the identity
\begin{equation*}
	\E[S_{\beta, \omega, \lambda} (X, Y) e^{-2\pi i \zeta X_i }] = 
		\Cov(r_{\beta, \lambda} (X, Y) e^{-2\pi i \langle \omega, \beta \circ X\rangle},  e^{2\pi i \zeta X_i })
\end{equation*}
we arrive at the following formula, which will be useful for future reference. 
\begin{lemma}
\label{lemma:major-term}
\begin{equation*}
	\mathcal{S}_i (\beta, \lambda) =  -\frac{1}{\lambda} \int_0^\infty \int_{\R^d}
		\int_\R \left|\Cov(r_{\beta, \lambda} (X, Y) e^{-2\pi i \langle \omega, \beta \circ X\rangle},  e^{2\pi i \zeta X_i }) \right|^2 \frac{1}{2\pi^2 \zeta^2} d\zeta \cdot t q_t(\omega) \, d\omega  \mu(dt).
\end{equation*} 
\end{lemma}

\subsubsection{Error Control}
\label{sec:error-control}
We now quantify the error between the directional derivative $\mathrm{D}\mathcal{J}(\beta, \lambda)[e_i]$ and 
its proxy $\mathcal{S}_i(\beta, \lambda)$. The difference between the two arises from the lack of centering 
in the original integrand, specifically the presence of the nonzero mean term $\E[R_{\beta, \omega, \lambda}(X, Y)]$.

The following lemmas provide bounds to control this discrepancy. Let 
\begin{equation*}
	M_\mu := \sup \{t: t\in \supp(\mu)\}
\end{equation*}  
denote the upper endpoint of the support of the measure $\mu$. 
\begin{lemma}
\label{lemma:residual-bound-one}
For every $\beta, \lambda$, 
\begin{equation*}
	\int_0^\infty \int_{\R^d} |\E[R_{\beta, \omega, \lambda}(X, Y)]|^2 t q_t(\omega) d\omega \mu(dt) 
		\le \lambda M_\mu \E[Y^2]. 
\end{equation*}
\end{lemma} 

\begin{proof}
We begin by observing that $t \le M_\mu$ for all $t$ in the support of $\mu$, and so
\begin{equation*}
	\int_0^\infty \int_{\R^d} |\E[R_{\beta, \omega, \lambda}(X, Y)]|^2 t q_t(\omega) d\omega \mu(dt)
		\le M_\mu \cdot \int_0^\infty \int_{\R^d} |\E[R_{\beta, \omega, \lambda}(X, Y)]|^2 q_t(\omega) d\omega \mu(dt). 
\end{equation*} 

To evaluate the right-hand side, we expand:
\begin{equation*}
\begin{split} 
	|\E[R_{\beta, \omega, \lambda}(X, Y)]|^2 &= \E[R_{\beta, \omega, \lambda}(X, Y) \overline{R_{\beta, \omega, \lambda}(X', Y')}] \\
		&=  \E[r_{\beta, \lambda}(X, Y) r_{\beta, \lambda}(X', Y') e^{-2\pi i \langle \omega, \beta \circ (X-X')\rangle}].
\end{split} 
\end{equation*}
where $(X', Y')$ is an independent copy of $(X, Y)$.
By~\eqref{eqn:computation-of-psi-psi-prime}, we have 
\[
\kernel_\beta(x, x') 
= \profile(\|\beta \circ (x - x')\|_1)
= \int_0^\infty \int_{\R^d} e^{-2\pi i \langle \omega, \beta \circ (x - x') \rangle} q_t(\omega) \, d\omega \, \mu(dt).
\]
Applying Fubini’s theorem, and using the representation above, we obtain:
\begin{equation*} 
\begin{split}
&\int_0^\infty \int_{\R^d} |\E[R_{\beta, \omega, \lambda}(X, Y)]|^2 q_t(\omega) \, d\omega \, \mu(dt)  \\
&= \E\left[ r_{\beta, \lambda}(X, Y) r_{\beta, \lambda}(X', Y') \cdot \int_0^\infty \int_{\R^d} 
	e^{-2\pi i \langle \omega, \beta \circ (X - X') \rangle} q_t(\omega) \, d\omega \, \mu(dt) \right] \\
&= \E\left[ r_{\beta, \lambda}(X, Y) r_{\beta, \lambda}(X', Y') \kernel(\beta \circ X, \beta \circ X') \right].
\end{split}
\end{equation*}
By Lemma~\ref{lemma:r-r'-K-bound}, this expectation equals $\lambda^2 \norm{f_{\beta, \lambda}}_H^2$, and 
is at most $\lambda \E[Y^2]$ by Lemma~\ref{lemma:trivial-bound-on-r-f}. 

This completes the proof. 
\end{proof} 

\begin{lemma}
\label{lemma:residual-bound-two}
For every $\beta, i, \lambda$, 
\begin{equation*}
	\int_0^\infty \int_{\R^d} |\E[R_{\beta, \omega, \lambda}(X, Y)|X_i - X_i'|] |^2 t q_t(\omega) d\omega \mu(dt) 
		\le 2 M_\mu\E[Y^2] \E[|X_i|^2]. 
\end{equation*}
\end{lemma}
\begin{proof}
By definition $|R_{\beta, \omega, \lambda}|(x, y) \le |r_{\beta, \lambda}|(x, y)$ for all $\beta, \omega, x, y$ and thus
\begin{equation*}
	\E[|R_{\beta, \omega, \lambda}(X, Y)|^2] \le \E[|r_{\beta, \lambda}(X, Y)|^2] \le \E[Y^2]. 
\end{equation*}
By Cauchy-Schwartz inequality, this implies 
\begin{equation*}
	 |\E[R_{\beta, \omega, \lambda}(X, Y)|X_i - X_i'|] |^2 \le \E[|R_{\beta, \omega, \lambda}(X, Y)|^2] \E[|X_i - X_i'|^2]
	 	\le 2\E[Y^2] \E[|X_i|^2].
\end{equation*} 
The estimate then follows from $\iint tq_t(\omega) d\omega \mu(dt) = \int t \mu(dt) \le M_\mu \int \mu(dt) = M_\mu$. 
\end{proof} 

With these lemmas in preparation, we reach the final conclusion. Let 
\begin{equation*}
	M_X:= \max_i \E[X_i^2],~~~\text{and}~~~M_Y:=\E[Y^2].
\end{equation*}
\begin{lemma}
\label{lemma:error-term}
For every $\beta, i, \lambda$, 
\begin{equation*}
	|\mathrm{D}\mathcal{J}(\beta, \lambda)[e_i] - \mathcal{S}_i(\beta, \lambda)| \le C(\frac{1}{\sqrt{\lambda}} + 1).
\end{equation*}
The constant $C  < \infty$ depends only on $M_X$, $M_Y$ and $M_\mu$. 
\end{lemma} 

\begin{proof} 
Fix $\beta, i, \lambda$. Note 
\begin{equation*}
	\lambda \cdot (\mathrm{D}\mathcal{J}(\beta, \lambda)[e_i] -  \mathcal{S}_i(\beta, \lambda)) = \Delta_1 + \Delta_2 - \Delta_3
\end{equation*}
where 
\begin{equation*}
\begin{split} 
	\Delta_1&= \int_0^\infty \int_{\R^d} \E[R_{\beta, \omega, \lambda}(X, Y)] \E[\overline{R_{\beta, \omega, \lambda}(X', Y')}|X_i - X_i'|] ~t q_t(\omega) d\omega \mu(dt) \\
	\Delta_2&= \int_0^\infty \int_{\R^d} \E[\overline{R_{\beta, \omega, \lambda}(X, Y)}] \E[R_{\beta, \omega, \lambda}(X', Y')|X_i - X_i'|] ~t q_t(\omega) d\omega \mu(dt) \\
	\Delta_3&= \int_0^\infty \int_{\R^d} \E[R_{\beta, \omega, \lambda}(X, Y)] \E[\overline{R_{\beta, \omega, \lambda}(X', Y')}] \E[|X_i - X_i'|] ~t q_t(\omega) d\omega \mu(dt) 
\end{split}. 
\end{equation*}
The measure $t q_t(\omega) d\omega \mu(dt)$ is finite with $\iint  t q_t(\omega) d\omega \mu(dt) \le M_\mu \iint q_t(\omega) d\omega \mu(dt) = M_\mu$.
Applying the Cauchy Schwarz inequality and invoking Lemmas~\ref{lemma:residual-bound-one} and~\ref{lemma:residual-bound-two}, we obtain that the error terms $|\Delta_1|$ and $|\Delta_2|$ are bounded by $C\sqrt{\lambda}$. Applying Lemma~\ref{lemma:residual-bound-one}, we get $|\Delta_3|$ is bounded by $C\lambda$. These 
constants $C < \infty$ depend only on $M_X, M_Y, M_\mu$.

\end{proof} 

\begin{remark}
\emph{
While both Sections~\ref{sec:an-integral-representation} and~\ref{sec:error-control} involve Fourier analysis---one exploits the forward transform of $\psi'$ (e.g., Lemma~\ref{lemma:integral-representation-of-derivative}), the other through the inverse transform of $\psi$ (e.g., Lemma~\ref{lemma:residual-bound-one}) ---this repetition is necessary. The key is that error control requires expansion in the frequency domain, where bounds involving $M_\mu$ emerge (Lemma~\ref{lemma:residual-bound-one}). We do not know how to carry out error control without working in the frequency domain.
}
\end{remark}

\subsubsection{Conclusion} 
\label{sec:derivative-of-J-with-respect-to-beta-i}
We conclude by presenting the decomposition of the directional derivative $\mathrm{D}\mathcal{J}(\beta, \lambda)[e_i]$.
For clarity of presentation, we define $ \mathcal{M}_i(\beta, \lambda) := -\lambda \mathcal{S}_i(\beta, \lambda)$. 

\vspace{.5em}
\begin{theorem}
\label{theorem:gradient-expression}
Let $\beta \in \R^d$ satisfy $\beta_i \ge 0$, and let $\lambda \in (0, 1)$. Then
\begin{equation*}
	\mathrm{D}\mathcal{J}(\beta, \lambda)[e_i] = - \frac{1}{\lambda} \mathcal{M}_i(\beta, \lambda) + \frac{1}{\sqrt{\lambda}}\mathcal{R}_i(\beta, \lambda)
\end{equation*}
where $|\mathcal{R}_i(\beta, \lambda)| \le C$ 
for some constant $C  < \infty$ depending only on $M_X$, $M_Y$, and $M_\mu$, and where 
\begin{equation*}
	 \mathcal{M}_i(\beta, \lambda) = \iiint \left|\Cov(r_{\beta, \lambda} (X, Y) e^{-2\pi i \langle \omega, \beta\circ X \rangle},  e^{2\pi i \zeta X_i}) \right|^2 \frac{1}{2\pi^2 \zeta^2} d\zeta \cdot t q_t(\omega) \, d\omega  \mu(dt)
\end{equation*} 
\end{theorem} 

\begin{proof}
This follows from Lemma~\ref{lemma:major-term} and~\ref{lemma:error-term}, and $ \mathcal{M}_i(\beta, \lambda) = -\lambda \mathcal{S}_i(\beta, \lambda)$
\end{proof} 

Theorem~\ref{theorem:gradient-expression} decomposes the directional derivative $\mathrm{D}\mathcal{J}(\beta, \lambda)[e_i]$ into two terms: $\mathcal{M}_i(\beta, \lambda)$ scaled by $1/\lambda$, and a remainder term $\mathcal{R}_i(\beta, \lambda)$ scaled by $1/\sqrt{\lambda}$. Crucially, the remainder term $\mathcal{R}_i(\beta, \lambda)$  is uniformly bounded.
The term  $\mathcal{M}_i(\beta, \lambda)$ aggregates the squared covariance between the modulated residual 
$r_{\beta, \lambda}(X, Y)\, e^{-2\pi i \langle \omega, \beta\circ X \rangle}$  and the coordinate function $e^{2\pi i \zeta X_i}$ evaluated at the current $\beta$, so intuitively it quantifies the extent to which feature $X_i$ explains what remains in the residual $r_{\beta, \lambda}(X, Y)$. This decomposition holds for all $\beta$ and any data distribution with finite second moments.
 
We note $\mathcal{M}_i(\beta, \lambda)$ depends on $\lambda$ since the residual $r_{\beta, \lambda}$ depends on $\lambda$. 
As a result, we are unsure whether it is the leading term. 
In the next subsection, we will evaluate $\mathcal{M}_i(\beta, \lambda)$ under the functional ANOVA model and identify regimes where it is \emph{uniformly bounded away from zero}. In these regimes, $-\mathcal{M}_i(\beta, \lambda)/\lambda$ is the dominating term in $\mathrm{D}\mathcal{J}(\beta, \lambda)[e_i]$ for small $\lambda$, and the interpretation is that the feature $X_i$ explains the residual and are detectable through the directional derivative $\mathrm{D}\mathcal{J}(\beta, \lambda)[e_i]$.



\subsection{Functional ANOVA Model} 
\label{sec:functional-anova-model}
By definition of the minimal sufficient feature set $S_*$, we have
\begin{equation*}
	\E[Y|X] = \E[Y|X_{S_*}]
\end{equation*}
so all predictive information in $X$ is contained in $X_{S_*}$.  
To quantify how each coordinate $X_j$ within $S_*$ contributes---either individually 
or through interactions---we impose the following assumption.

\vspace{.5em}

\begin{assumption}
\label{assumption:feature-variable}
The variables $(X_j)_{j \in S_*}$ in $S_*$ are mutually independent.
\end{assumption} 


Under this independence assumption~\ref{assumption:feature-variable}, the regression function $f(x) = \E[Y|X=x]$ admits a unique functional ANOVA decomposition~\cite{Stone94}: 
\[
    f(x) = f_0 + \sum_{\emptyset \neq S \subseteq S_*} f_S(x_S),
\]
where $x_S = (x_j)_{j \in S}$ and each $f_S$ depends only on the coordinates in $S$, 
and the component functions $f_S$ satisfy the orthogonality 
conditions: 
\[
    \mathbb{E}[f_S(X_S)|X_T] = 0, ~~~\forall T \subsetneq S.
\]
In this decomposition, the constant term is $f_0 = \E[Y] = 0$, the first-order component is $f_i(X_i) = \E[Y|X_i]$, 
and the second-order componet is $f_{i, j}(X_i, X_j) = \E[Y|X_i, X_j] - \E[Y|X_i] -\E[Y|X_j]$, and so on for higher-order terms.
Each $f_S$ captures a distinct contribution to the regression function $f$---whether a main effect
from a single feature ($|S| = 1$) or from interactions among multiple features ($|S| > 1$). The decomposition makes explicit the contributions of individual features and their interactions. 

In what follows, we use the functional ANOVA model as a stylized framework to understand feature learning. 
Although some of the results can be extended to slightly weaker conditions---such as replacing mutual independence with pairwise or approximate independence---we do not pursue that direction here, as it would be a distraction from the main ideas.

\subsubsection{Decomposition of Directional Derivative under Functional ANOVA} 
Under the functional ANOVA model, the directional derivative at those $\beta$ with $\beta_i = 0$
\begin{equation*}
	 D \mathcal{J}(\beta, \lambda)[e_i] 
\end{equation*}
admits a particularly simple decomposition. The perhaps surprise fact is that the term $\mathcal{M}_i(\beta, \lambda)$ in the decomposition
(Theorem~\ref{theorem:gradient-expression}) becomes independent of $\lambda$. Lemma~\ref{lemma:stupid-term-in-S} offers the key insight. 

\begin{lemma}
\label{lemma:stupid-term-in-S}
Suppose  \( X_i \) is independent of \( (X_j)_{j \ne i} \). Then, at every $\beta$ with $\beta_i = 0$, 
\begin{equation*}
	\Cov\big(r_{\beta, \lambda}(X, Y) e^{-2\pi i \langle \omega, \beta \circ X \rangle}, e^{2\pi i \zeta X_i}\big) = 
		\E[(Y - \E[Y|X_{\supp(\beta)}]) e^{-2\pi i \langle \omega, \beta \circ X \rangle}  e^{-2\pi i \zeta X_i}].
\end{equation*}
In particular, the covariance term on the left-hand-side is independent of $\lambda$.
\end{lemma}
\begin{proof} 
Let $\beta_i = 0$. By definition, 
\begin{equation*}
\begin{split} 
&r_{\beta, \lambda}(X, Y) e^{-2\pi i \langle \omega, \beta \circ X \rangle} \\
& = (Y-  \E[Y|X_{\supp(\beta)}]) e^{-2\pi i \langle \omega, \beta \circ X \rangle} + 
			(\E[Y|X_{\supp(\beta)}] -  f_{\beta, \lambda}(\beta \circ X)) e^{-2\pi i \langle \omega, \beta \circ X \rangle}.
\end{split}
\end{equation*} 
Since $X_i$ is independent of $X_{\supp(\beta)}$, the second term above is independent of  \( e^{2\pi i \zeta X_i} \). Therefore,
\[
\Cov\big(r_{\beta, \lambda}(X, Y) e^{-2\pi i \langle \omega, \beta \circ X \rangle}, \, e^{2\pi i \zeta X_i}\big)
    = \Cov\big((Y-\E[Y|X_{\supp(\beta)}]) e^{-2\pi i \langle \omega, \beta \circ X \rangle}, \, e^{2\pi i \zeta X_i}\big).
\]
Since the first factor in the covariance is mean-zero, the conclusion follows. 
\end{proof}

\begin{theorem} 
\label{theorem:gradient-expression-ANOVA}
Let $\beta \in \R^d$ satisfy $\beta_i = 0$, and let $\lambda \in (0, 1)$. Assume Assumptions~\ref{assumption:existence-of-csfs},~\ref{assumption:independence},~\ref{assumption:feature-variable}. Let 
\begin{equation*} 
	 \mathcal{M}_i(\beta): =  \iiint \left|\mathbb{E}\left[(Y-\E[Y|X_{\supp(\beta)}]) e^{-2\pi i \langle \omega, \beta \circ X \rangle}  e^{-2\pi i \zeta X_i}\right]\right|^2 \frac{1}{2\pi^2 \zeta^2} d\zeta \cdot t q_t(\omega) \, d\omega  \mu(dt).
\end{equation*}
Then
\begin{equation*}
	\mathrm{D}\mathcal{J}(\beta, \lambda)[e_i] = - \frac{1}{\lambda} \mathcal{M}_i(\beta) + \frac{1}{\sqrt{\lambda}}\mathcal{R}_i(\beta, \lambda)
\end{equation*}
where $|\mathcal{R}_i(\beta, \lambda)| \le C$ 
for some constant $C < \infty$ depending only on $M_X$, $M_Y$, and $M_\mu$.
\end{theorem} 

\begin{proof} 
This is a consequence of Theorem~\ref{theorem:gradient-expression} and Lemma~\ref{lemma:stupid-term-in-S}.
\end{proof} 

Theorem~\ref{theorem:gradient-expression-ANOVA} refines the earlier decomposition in Theorem~\ref{theorem:gradient-expression} by showing that 
the directional derivative $\mathrm{D}\mathcal{J}(\beta, \lambda)[e_i]$ splits into a dominant term 
$-\mathcal{M}_i(\beta)/\lambda$, and a smaller term of order $1/\sqrt{\lambda}$, which is negligible when 
$\mathcal{M}_i(\beta) > 0$. Indeed, Theorem~\ref{theorem:gradient-expression-ANOVA} not only shows that for any $\beta$ with $\beta_i = 0$,  
\[
\lim_{\lambda \to 0^+} \lambda \, \mathrm{D}\mathcal{J}(\beta, \lambda)[e_i] = -\mathcal{M}_i(\beta),
\]
but also, via the uniform bound $|\mathcal{R}_i(\beta, \lambda)| \le C$, yields the stronger statement  
\[
\lim_{\lambda \to 0^+} 
\sup_{\beta \in \R^d: \, \beta_i = 0} 
\left| \lambda \, \mathrm{D}\mathcal{J}(\beta, \lambda)[e_i] + \mathcal{M}_i(\beta) \right| = 0.
\]
It is therefore important to understand when the leading term $\mathcal{M}_i(\beta)$ is strictly postive. 

\vspace{.8em} 
\begin{lemma}
\label{lemma:S-I-positive} 
$\mathcal{M}_i(\beta) \ge 0$ with equality holds if and only if $\E[Y|X_{\supp(\beta) \cup \{i\}}] = \E[Y|X_{\supp(\beta)}]$.
\end{lemma}

\begin{proof} 
We investigate when equality $\mathcal{M}_i(\beta) = 0$ holds. Let us denote  
$Q(\omega) = \int_0^\infty t q_t(\omega) \mu(dt)$. 
Since $q_t$ is 
strictly positive and $\mu$ is not concentrated at zero, we get $Q(\omega) > 0$ for all $\omega$. Thus, 
\begin{equation*}
	 \mathcal{M}_i(\beta) = \iint \left|\mathbb{E}[(Y-\E[Y|X_{\supp(\beta)}]) e^{-2\pi i \langle \omega, \beta \circ X \rangle}  e^{-2\pi i \zeta X_i}]\right|^2 \frac{Q(\omega)}{2\pi^2 \zeta^2} d\zeta  \, d\omega.
\end{equation*}
The weights $Q(\omega)/(2\pi^2 \zeta^2)$ is strictly positive. Thus, $\mathcal{M}_i (\beta) = 0$ if and only if 
\begin{equation*}
	\mathbb{E}[(Y-\E[Y|X_{\supp(\beta)}]) e^{-2\pi i \langle \omega, \beta \circ X \rangle}  e^{-2\pi i \zeta X_i}] = 0~~a.e.~\omega, \zeta.
\end{equation*}
Then 
\begin{equation*}
	\E\left[\E[Y|X_{\supp(\beta) \cup \{i\}}]-\E[Y|X_{\supp(\beta)}]e^{-2\pi i \langle \omega, \beta \circ X \rangle}  e^{-2\pi i \zeta X_i}\right] = 0~~a.e.~\omega, \zeta.
\end{equation*} 
This implies the random variable 
\begin{equation*}
\E[Y|X_{\supp(\beta) \cup \{i\}}]-\E[Y|X_{\supp(\beta)}]
\end{equation*} 
 must be zero almost surely. 
\end{proof} 

Therefore, the leading term $\mathcal{M}_i(\beta) > 0$  if and only if $\E[Y | X_{\supp(\beta) \cup \{i\}}] \ne  \E[Y|X_{\supp(\beta)}]$; that is, if $X_i$ helps explain the target $Y$ beyond what is already captured by $X_{\supp(\beta)}$. 



\vspace{.5em} 
\begin{remark}
\emph{
Theorem~\ref{theorem:gradient-expression-ANOVA} and Lemma~\ref{lemma:S-I-positive} together provide intuitions how the objective $\mathcal{J}$ learns relevant features. They show that when $X_i$ helps explain the target $Y$ beyond what is already captured by $X_{\supp(\beta)}$, then the term $\mathcal{M}_i(\beta)$ is strictly positive, making the derivative $\mathrm{D} \mathcal{J}(\beta, \lambda)[e_i]$ strictly negative for small $\lambda$, thereby pushing the model to activate that coordinate $i$.
}
\end{remark}

\subsection{Relevant Feature Recovery} 
\label{sec:relevant-feature-recovery-at-stationary-points}
In this section, we provide answers to ($\mathsf{Q4}$)---namely, understand when stationary points of 
$\mathcal{J}$ recover the relevant features. To this end, we assume independence of the feature variables
(Assumption~\ref{assumption:feature-variable}),  and work under the functional ANOVA model 
(section~\ref{sec:functional-anova-model})
\begin{equation*}
\E[Y | X] = \sum_{\emptyset \neq S \subseteq S_*} f_S(X_S).
\end{equation*}
A feature $i \in S_*$ may contribute to prediction through one of two types of signals: 
(i) a \emph{main effect},  meaning the univariate component $f_{{i}}$ in the ANOVA expansion is nonzero, i.e., 
$f_{i}(X_i) \not\equiv 0$ in $L^2(\P)$; or (ii) an \emph{interaction}, meaning $i$ appears in some higher-order 
component $f_S$ with $|S| \ge 2$ and $f_S \not\equiv 0$ for some $S \ni i$. 
In the subsections that follow, we investigate regimes under which stationary points of $\mathcal{J}$ recover the corresponding feature.

\subsubsection{Recovery of Main Effects} 
We begin with the case where a relevant feature $i \in S_*$ contributes through a main effect---that is, $\E[Y | X_i] \ne 0$. In this case, we show that the directional derivative $\mathrm{D}\mathcal{J}(\beta, \lambda)[e_i]$ is uniformly negative over a neighborhood of $\beta$ where $\beta_i = 0$, ensuring the model is driven to activate coordinate $i$.
\vspace{.5em} 

\begin{theorem}
\label{theorem:uniform-lower-bound-on-M-i-beta}
Assume Assumptions~\ref{assumption:existence-of-csfs},~\ref{assumption:independence},~\ref{assumption:feature-variable}. Suppose $i \in S_*$ has a main effect: $\E[Y|X_i] \ne 0$. Then, for every compact set $\mathsf{C} \subset \{\beta \in \R^d : \beta_i = 0\}$, we have
	\begin{equation*}
		\inf_{\beta \in \mathsf{C}} \mathcal{M}_i(\beta) > 0.
	\end{equation*}
Consequently, there exist constants $\varepsilon > 0$ and $\lambda_0 > 0$ such that for all $\lambda \in (0, \lambda_0)$ and all $\beta \in \mathsf{C}$,
\begin{equation*}
	\lambda \cdot \mathrm{D}\mathcal{J}(\beta, \lambda)[e_i] \le -\varepsilon.
\end{equation*}
\end{theorem}

\begin{remark}
\emph{
Theorem~\ref{theorem:uniform-lower-bound-on-M-i-beta} shows that any feature $i$ with a nontrivial main effect induces a uniformly negative directional derivative across compact sets where it is inactive (i.e, $\beta_i = 0$). This guarantees a descent direction toward activating $X_i$, reinforcing the idea that features of main effects are reliably detectable by the objective when $\lambda$ is small.
}
\end{remark}

\begin{proof}[Proof of Theorem~\ref{theorem:uniform-lower-bound-on-M-i-beta}]
We note $\mathcal{M}_i(\beta)$ is not continuous in $\beta$ due to the discontinuity of $\supp(\beta)$. 
To remedy this issue, we define for every $i$, subset $A \subset \{1, 2, \ldots, d\}$, and $\beta \in \R^d$
\begin{equation*}
	 \mathcal{N}_{i, A}(\beta) = \iiint \left|\E[(Y-\E[Y|X_A]) e^{-2\pi i \langle \omega, \beta\circ X \rangle} e^{-2\pi i \zeta X_i}] \right|^2 \frac{1}{2\pi^2 \zeta^2} d\zeta \cdot t q_t(\omega) \, d\omega  \mu(dt).
\end{equation*} 
By definition, $\mathcal{M}_i(\beta)  = \mathcal{N}_{i, A}(\beta)$ when $\supp(\beta) = A$. 

We claim that $\mathcal{N}_{i, A}(\beta)$ is lower semicontinuous. Fix any sequence $\beta^{(n)} \to \beta$. For each fixed $(\omega, \zeta, t)$, the inner expectation in the integrand is continuous in $\beta$, so the integrand converges pointwise. Moreover, the integrand is nonnegative. By Fatou’s lemma, we conclude
\[
\liminf_{n \to \infty} \mathcal{N}_{i, A}(\beta^{(n)}) \ge \mathcal{N}_{i, A}(\beta),
\]
so $\mathcal{N}_{i, A}$ is lower semicontinuous on $\R^d$. 

Next, we show $\mathcal{N}_{i, A}(\beta)  > 0$ when $i \not \in A$. To see this, 
suppose $ \mathcal{N}_{i, A}(\beta) = 0$. Then the integrand vanishes almost everywhere, 
\begin{equation*}
	\mathbb{E}[(Y-\E[Y|X_A]) e^{-2\pi i \langle \omega, \beta \circ X \rangle}  e^{-2\pi i \zeta X_i}] = 0~~a.e.~\omega, \zeta.
\end{equation*}
Let $B:=\supp(\beta)$. Then
\begin{equation*}
	\E\left[\E[(Y-\E[Y|X_A])|X_{B \cup \{i\}}]e^{-2\pi i \langle \omega, \beta \circ X \rangle}  e^{-2\pi i \zeta X_i}\right] = 0~a.e.~\omega, \zeta
\end{equation*} 
which implies the random variable $\E[(Y-\E[Y|X_A])|X_{B \cup \{i\}}]$
must be zero almost surely. However, consider its
inner product with the main effect component $f_{{i}}(X_i)$. If $i \not\in A$, then 
\begin{equation*}
\begin{split}
\mathbb{E}\left[\mathbb{E}[Y - \mathbb{E}[Y | X_A] | X_{B \cup \{i\}}] \cdot f_{{i}}(X_i)\right]
&= \mathbb{E}[(Y - \mathbb{E}[Y | X_A]) \cdot f_{{i}}(X_i)] \\
&= \mathbb{E}[Y \cdot f_{{i}}(X_i)] - \mathbb{E}[\mathbb{E}[Y| X_A] \cdot f_{{i}}(X_i)] \\
&= \mathbb{E}[f_{{i}}^2(X_i)],
\end{split}
\end{equation*}
where the last equality uses that $\E[f_i(X_i)] = 0$ and $X_i$ is independent of $X_A$ whenever $i \notin A$. Since $f_i(X_i) = \E[Y | X_i] \not\equiv 0$ by assumption, we have $\E[f_i^2(X_i)] > 0$, contradicting the claim that the random variable is zero. Thus, 
we have $\mathcal{N}_{i, A}(\beta)  > 0$ when $i \not \in A$.


We now return to the proof of Theorem~\ref{theorem:uniform-lower-bound-on-M-i-beta}. Fix any subset 
$A \subset \{1, \dots, d\} \setminus \{i\}$.  The function $\mathcal{N}_{i, A}(\beta)$ is lower semicontinuous 
and  strictly positive on the compact set $\mathsf{C}$. By the extreme value theorem, it attains its minimum
on $\mathsf{C}$. Thus, there exists a constant $\varepsilon_A > 0$ such that
\[
\mathcal{N}_{i, A}(\beta) \ge \varepsilon_A \quad \forall \beta \in \mathsf{C}.
\]
Let $\varepsilon > 0$ be the minimum of all such $\varepsilon_A$ over subsets $A \subset \{1, \dots, d\} \setminus \{i\}$. 

For any $\beta \in \mathsf{C}$, we have $i \notin \supp(\beta)$ by assumption, so
\begin{equation*}
	\mathcal{M}_i(\beta) = \mathcal{N}_{i, \supp(\beta)}(\beta) \ge \eps_{\supp(\beta)} \ge \eps > 0.
\end{equation*}
This shows $\inf_{\beta \in \mathsf{C}} \mathcal{M}_i(\beta)  > 0$, as desired.
%
%
%

Finally, by Theorem~\ref{theorem:gradient-expression-ANOVA}, there exists $C < \infty$ such that for all $\lambda \in (0, 1)$,
and for all $\beta \in \mathsf{C}$
\begin{equation*}
	\lambda \cdot \mathrm{D}\mathcal{J}(\beta, \lambda)[e_i] \le -\mathcal{M}_i(\beta) + C \sqrt{\lambda}.
\end{equation*}
It follows that for sufficiently small \( \lambda > 0 \), we have
$\lambda \cdot \mathrm{D}\mathcal{J}(\beta, \lambda)[e_i] \le -\eps$ must hold for all $\beta \in \mathsf{C}$.
\end{proof}

\vspace{.5em} 
\begin{corollary}
\label{corollary:extraction-of-main-effect}
Assume Assumptions~\ref{assumption:existence-of-csfs},~\ref{assumption:independence},~\ref{assumption:feature-variable}. Suppose $i \in S_*$ has a main effect: $\E[Y|X_i] \ne 0$.
 
For every compact subset $\mathsf{C}$ in $\R^d$, there is $\lambda_0 > 0$ such that for all 
$\lambda \in (0, \lambda_0)$, if $\beta$ is a directional stationary point of $\mathcal{J}(\cdot, \lambda)$, 
then one of the following must hold: 
\begin{equation*}
	(i)~~i \in \supp(\beta_{\dag})~~\qquad \text{or} \qquad (ii)~~\beta_{\dag} \not \in \mathsf{C}.
\end{equation*} 
\end{corollary} 

Corollary~\ref{corollary:extraction-of-main-effect} strengthens Corollary~\ref{corollary:beta=zero-not-stationary-ell-one} by extending the signal-extraction property from the origin $\beta = 0$ to arbitrary compact subsets of $\R^d$. Once a main effect exists for a feature $X_i$, any directional stationary point that remains within a bounded region must include $X_i$ for small $\lambda$.

\subsubsection{Recovery of Interactions} 
Next, we consider features $i \in S_*$ that contribute only through interaction terms---that is, $f_{{i}} \equiv 0$ but $f_S \not\equiv 0$ for some $S \ni i$ with $|S| \ge 2$. This implies in particular $\E[Y|X_S] \ne \E[Y|X_{S \backslash \{i\}}]$. At the origin $\beta = 0$, the directional derivative $\mathrm{D}\mathcal{J}(0, \lambda)[e_i]$ vanishes (Corollary~\ref{corollary:beta=zero-not-stationary-ell-one}), reflecting that $X_i$ does not influence the objective at first order when considered independently of other variables. Consequently, learning such features requires moving away from the origin to parameter configurations of $\beta$ that capture their joint contribution.

We show that once the other variables in an interaction set $S$ are sufficiently active, the objective induces a descent direction along coordinate $i$. Specifically, we show that the directional derivative $\mathrm{D}\mathcal{J}(\beta, \lambda)[e_i]$ is uniformly negative over a neighborhood of $\beta$ where $\beta_i = 0$, and $\beta_j$ remains bounded away from zero for all $j \in S \setminus \{i\}$. The interpretation is that the model is driven to activate coordinate $i$ once its interaction partners are already represented in the current parameter configuration.

\begin{theorem}
\label{theorem:uniform-lower-bound-on-M-i-beta-interaction}
Assume Assumptions~\ref{assumption:existence-of-csfs},~\ref{assumption:independence},~\ref{assumption:feature-variable}. Let $i \in S \subset S_*$ where $\E[Y|X_{S \backslash \{i\}}] \ne \E[Y|X_S]$.

Let $\mathsf{C}$ be a compact set in $\R^d$ with $\mathsf{C} \subset \left\{ \beta \in \mathbb{R}^d : \beta_i = 0 \text{ and } \beta_j > 0 \ \text{for all } j \in S \setminus \{i\} \right\}$. Then,
	\begin{equation*}
		\inf_{\beta \in \mathsf{C}} \mathcal{M}_i(\beta) > 0.
	\end{equation*}
Consequently, there exist constants $\varepsilon > 0$ and $\lambda_0 > 0$ such that for all $\lambda \in (0, \lambda_0)$ and all $\beta \in \mathsf{C}$,
\begin{equation*}
	\lambda \cdot \mathrm{D}\mathcal{J}(\beta, \lambda)[e_i] \le -\varepsilon.
\end{equation*}
\end{theorem}

\begin{remark}
\emph{
Theorem~\ref{theorem:uniform-lower-bound-on-M-i-beta-interaction} shows that a feature $i$ involved interactions with other variables $X_{S\backslash \{i\}}$ induces a uniformly negative directional derivative over compact sets where $\beta_i = 0$ and the other coordinates in $S \setminus \{i\}$ are active. This ensures a descent direction along $e_i$, and such features can be learned through coordinated activation of their interaction partners when $\lambda$ is small.  \\ \indent 
Mathematically, Theorem~\ref{theorem:uniform-lower-bound-on-M-i-beta-interaction} generalizes Theorem~\ref{theorem:uniform-lower-bound-on-M-i-beta}: when $S = \{i\}$, the interaction reduces to a main effect and the condition on $S \setminus \{i\}$ is vacuous. Conceptually, the key distinction is that main effects are identifiable from first-order variation already at $\beta = 0$, whereas interaction effects become identifiable once the other variables in $S$ are sufficiently active. 
}
\end{remark}
\begin{proof}[Proof of Theorem~\ref{theorem:uniform-lower-bound-on-M-i-beta-interaction}]
 We define 
for every $i$, subset $A \subset \{1, 2, \ldots, d\}$, and $\beta \in \R^d$
\begin{equation*}
	 \mathcal{N}_{i, A}(\beta) = \iiint \left|\E[(Y-\E[Y|X_A]) e^{-2\pi i \langle \omega, \beta\circ X \rangle} e^{-2\pi i \zeta X_i}] \right|^2 \frac{1}{2\pi^2 \zeta^2} d\zeta \cdot t q_t(\omega) \, d\omega  \mu(dt).
\end{equation*} 
By definition, $\mathcal{M}_i(\beta)  = \mathcal{N}_{i, A}(\beta)$ when $\supp(\beta) = A$. Also, 
 $\mathcal{N}_{i, A}(\beta)$ is lower semicontinuous in $\R^d$. 

We show $\mathcal{N}_{i, A}(\beta)  > 0$ when $i \not \in A$ and $S \backslash \{i\} \subset \supp(\beta)$. To see this, 
suppose $ \mathcal{N}_{i, A}(\beta) = 0$. Then 
\begin{equation*}
	\mathbb{E}[(Y-\E[Y|X_A]) e^{-2\pi i \langle \omega, \beta \circ X \rangle}  e^{-2\pi i \zeta X_i}] = 0~~a.e.~\omega, \zeta.
\end{equation*}
This implies that the characteristic function of the random variable
\begin{equation*}
	\E[(Y-\E[Y|X_A])|X_{B \cup \{i\}}]~~~~\text{where}~~~B:=\supp(\beta)
\end{equation*} 
vanishes almost everywhere, so the random variable must be zero almost surely. However, consider its
inner product with $g(X_S):= \E[Y|X_S] - \E[Y|X_{S \backslash \{i\}}]$. If $i \not\in A$ and $S \backslash \{i\} \subset B$, then 
$S \subset B \cup \{i\}$, and 
\begin{equation*}
\begin{split}
\mathbb{E}\left[\mathbb{E}[Y - \mathbb{E}[Y | X_A] | X_{B \cup \{i\}}] \cdot g(X_S)\right]
&= \mathbb{E}[(Y - \mathbb{E}[Y | X_A]) \cdot g(X_S)] \\
&= \mathbb{E}[Y \cdot g(X_S)] - \mathbb{E}[\mathbb{E}[Y| X_A] \cdot g(X_S)] \\
&= \mathbb{E}[g^2(X_S)],
\end{split}
\end{equation*}
where the last equality follows from the orthogonality of ANOVA components: 
\begin{equation*}
	\mathbb{E}[\mathbb{E}[Y| X_A] \cdot g(X_S)]
		= \mathbb{E} [\sum_{T \subset A} f_{T}(X_T) \cdot \sum_{i \in W, W \subset S} f_W(X_W)] = 0.
\end{equation*}
This yields a contradiction, since $\mathbb{E}[g^2(X_S)] > 0$ by assumption. Therefore, $\mathcal{N}_{i, A}(\beta) > 0$ when $i \not \in A$ and $S \backslash \{i\} \subset \supp(\beta)$.


We now return to the proof of Theorem~\ref{theorem:uniform-lower-bound-on-M-i-beta-interaction}. Fix any subset $A \subset \{1, \dots, d\} \setminus \{i\}$. By construction of $\mathsf{C}$, we have $\beta_i = 0$ and $S \setminus \{i\} \subset \supp(\beta)$ for all $\beta \in \mathsf{C}$. Therefore, the preceding argument apply, and we conclude that $\mathcal{N}_{i,A}(\beta) > 0$ for all $\beta \in \mathsf{C}$. Moreover, since $\mathcal{N}_{i,A}$ is lower semicontinuous, and $\mathsf{C}$ is compact, it attains its minimum on $\mathsf{C}$. Thus, there exists a constant $\varepsilon_A > 0$ such that
\[
\mathcal{N}_{i, A}(\beta) \ge \varepsilon_A \quad \forall \beta \in \mathsf{C}.
\]
Let $\varepsilon > 0$ be the minimum of all such $\varepsilon_A$ over subsets $A \subset \{1, \dots, d\} \setminus \{i\}$. 

For any $\beta \in \mathsf{C}$, we have $i \notin \supp(\beta)$ by assumption, so
\begin{equation*}
	\mathcal{M}_i(\beta) = \mathcal{N}_{i, \supp(\beta)}(\beta) \ge \eps_{\supp(\beta)} \ge \eps > 0.
\end{equation*}
This shows $\inf_{\beta \in \mathsf{C}} \mathcal{M}_i(\beta)  > 0$, as desired.
%
%
%

Finally, by Theorem~\ref{theorem:gradient-expression-ANOVA}, there exists $C < \infty$ such that for all $\lambda \in (0, 1)$,
and for all $\beta \in \mathsf{C}$
\begin{equation*}
	\lambda \cdot \mathrm{D}\mathcal{J}(\beta, \lambda)[e_i] \le -\mathcal{M}_i(\beta) + C \sqrt{\lambda}.
\end{equation*}
It follows that for sufficiently small \( \lambda > 0 \), we have
$\lambda \cdot \mathrm{D}\mathcal{J}(\beta, \lambda)[e_i] \le -\eps$ must hold for all $\beta \in \mathsf{C}$.
\end{proof}

\vspace{.5em} 
\begin{corollary}
\label{corollary:extraction-of-interaction-effect}
Assume Assumptions~\ref{assumption:existence-of-csfs},~\ref{assumption:independence}, and~\ref{assumption:feature-variable}.  
Let $i \in S \subset S_*$ with $\E[Y|X_{S \backslash \{i\}}] \ne \E[Y|X_S]$, and let $\mathsf{C} \subset \mathbb{R}^d$ be a compact set such that
\[
\beta_j > 0 \quad \text{for all } j \in S \setminus \{i\}, \quad \text{for all } \beta \in \mathsf{C}.
\]
Then there exists $\lambda_0 > 0$ such that for all $\lambda \in (0, \lambda_0)$, if $\beta_\dagger$ is a directional stationary point of $\mathcal{J}(\cdot, \lambda)$, then one of the following holds:
\begin{equation*}
	(i)~~i \in \supp(\beta_{\dag})~~\qquad \text{or} \qquad (ii)~~\beta_{\dag} \not \in \mathsf{C}.
\end{equation*} 
\end{corollary} 

Corollary~\ref{corollary:extraction-of-interaction-effect} complements Corollary~\ref{corollary:extraction-of-main-effect} by extending the signal-extraction guarantee to nonlinear interaction effects. Specifically, once an interaction involving $X_i$ is present---that is, $\E[Y|X_S] \ne \E[Y|X_{S \backslash \{i\}}]$ for some $S \ni i$---any directional stationary point must include $X_i$ as long as the remaining variables in $S \setminus \{i\}$ are sufficiently active (i.e., their corresponding $\beta_j$ are bounded away from zero) and $\lambda$ is sufficiently small. In this sense, interaction effects can still be learned, provided the other interacting variables are already active.

\section{Open Problems} 
\label{sec:open-questions}
We list some open questions which are important for theoretical or numerical development of 
the compositional kernel model.

\vspace{.5em}
\begin{itemize}
\item (Dynamical problem)
This paper has focused on the statics of the variational problem, but in practice one needs iterative schemes to minimize the nonconvex objective $\mathcal{J}$. Important questions concern the design of initialization and the choice of update dynamics. For example, our analysis shows that when predictive features contribute only through interactions, $\beta = 0$ is always a stationary point. This indicates the importance of carefully chosen initializations to ensure recovery of the true signals.

\vspace{.5em}
\item (Finite sample rates)   
Our analysis is at the population level; in practice we only have access to finite sample data. How many samples suffice to guarantee that the same variable selection results hold with high probability?

\vspace{.5em}
\item (Generalization error) 
In the regimes where the coordinates of $X$ are independent, selecting the right features is a necessary prerequisite for good generalization. How do our population-level \emph{variable selection} results translate into guarantees on finite sample \emph{generalization error}? Does an $\ell_1$ norm penalization on $\beta$ help reduce sample complexity and improve generalization?   

\vspace{.5em}
\item (Dependence) 
In the context of predictive feature recovery, our results assume independence between features. As explained earlier, this independence is mainly used for attributing prediction signals to individual features and feature subsets. Extending the analysis to dependent designs is particularly interesting.

\vspace{.5em}
\item (Scale)  
This paper discusses the problem of \emph{feature learning} through the lens of \emph{variable selection}. However, we do not analyze what the magnitude of $\beta$ means. Scale plays a crucial role in trading-off approximation and regularization, and the real problem is to learn all the intrinsic scales of the data, a point being raised in the related work on $f(UX)$ model~\cite{LiRu25}. Can the $\beta$-parameterized model learn such scales automatically?

\vspace{.5em} 
\item (Nonsmooth kernels and the $f(UX)$ context) This paper has demonstrated the benefit of using coordinate-wise $\ell_1$-type kernels, such as the Laplace kernel $e^{-\|x-x'\|_1}$, for sparse feature recovery. A natural next question is whether nonsmooth but rotationally invariant kernels, such as the rotational version of the Laplace kernel $e^{-\|x-x'\|_2}$, can aid recovery of low-dimensional subspaces in the related $f(UX)$ context. 

\end{itemize}

\section{Acknowledgements} 
Feng Ruan would like to thank Dmitriy Drusvyatskiy and Damek Davis for conversations that motivated the revision of this work, whose first version appeared online about four years ago~\cite{RuanLiJo21}. This revision makes significant efforts to clarify the structure of the arguments and incorporate new results on denoising for stationary points, with the aim of presenting the framework in a more systematic and transparent manner. An independent motivation of this revision is to provide a self-contained exposition (with some generalization) of some results in one author’s related work~\cite{LiRu25, LiRub25}---which develops a geometric foundation for compositional kernel models $f(UX)$---in a more accessible style to broader audience. Feng Ruan would also like to thank Yang Li for carefully reading the manuscript and for his encouragement.  This work was partially supported by the National Science Foundation under Grant DMS-2540678.

\newpage
\bibliography{sn-bibliography}

\newpage
\appendix

\section{Proofs} 
\subsection{Proof of Lemma~\ref{lemma:continuous-embedding}}
\label{sec:proof-lemma-continuous-embedding}
By Cauchy-Schwartz, we have 
\begin{equation*}
	\norm{f}_{L_\infty}^2 \le \|\hat{f}\|_{L_1}^2 \le 
		 \int_{\R^d} \frac{|\hat{f}|^2(\omega)}{k(\omega)} d\omega \int_{\R^d} k(\omega) d\omega = \norm{f}_H^2.
\end{equation*} 
Since $\hat{f} \in L_1$, the continuity of $f$ follows from Lebesgue's dominated convergence theorem. 
By the Riemann-Lebesgue lemma, we get $\lim_{x \to \infty} f(x) = 0$.

\subsection{Proof of Lemma~\ref{lemma:denseness-of-H-in-L-2}}
\label{sec:proof-of-lemma-denseness-of-H-in-L-2}
Suppose this is not true. There is nonzero $f \in L^2(\nu)$ such that $\int f h d\nu = 0$ 
for all $h \in H$. In particular, taking $h(\cdot) = \kernel(\cdot - z) \in H$, we obtain
$\int f(x) \kernel(x-z)\nu(dx) = 0$ for all $z \in \R^d$.  Using the Fourier inversion formula, 
$\kernel(x-z) =  \int k(\omega) e^{2\pi i \langle x-z, \omega\rangle} d\omega$, we get
\begin{equation*}
\begin{split} 
	\iint f(x)  k(\omega) e^{2\pi i \langle x-z, \omega\rangle} d\omega \nu(dx) = 0~~~\forall z \in \R^d.
\end{split} 
\end{equation*} 
Define the Fourier transform of $f$ under $\nu$ as $F(\omega) := \int f(x) e^{-2\pi i \langle x, \omega \rangle} \nu(dx)$. Then the above equation becomes
\begin{equation*}
	\int F(-\omega) k(\omega) e^{-2\pi i\langle z, \omega\rangle}  d\omega = 0~~~\forall z \in \R^d.
\end{equation*} 
That is, the Fourier transform of $F(-\omega) k (\omega)$ is identically zero. 
Since $k(\omega) > 0$, this implies $F(\omega) = 0$ almost everywhere $\omega$, so $f = 0$ in $L^2(\nu)$---a contradiction.

\subsection{Proof of Lemma~\ref{lemma:euler-lagrange-identity}}
\label{sec:proof-of-lemma-euler-lagrange-identity}
By setting the first variation of $f \mapsto \mathcal{I}(f,\beta,\lambda)$ equal to zero, we get 
\begin{equation*}
	\E[(Y- f_{\beta, \lambda}(\beta \circ X)) g(\beta \circ X)] = \lambda \langle f_{\beta, \lambda}, g\rangle_H,~~~\forall g\in H.
\end{equation*}
For every $z \in \R^d$, define $g_{z} := \kernel(\cdot, z) \in H$. Applying Lemma~\ref{lemma:euler-lagrange-identity} with $g_{z}$, we get 
\begin{equation*}
	\E[r_{\beta, \lambda}(X, Y) g_{z}(\beta \circ X)]
		= \lambda \langle f_{\beta, \lambda}, g_{z}\rangle_H.	
\end{equation*}
By the reproducing property of the RKHS $H$, $\langle f_{\beta, \lambda}, g_{z}\rangle_H = \langle f_{\beta, \lambda}, \kernel(\cdot, z)\rangle_H = f_{\beta, \lambda}(z)$. Substituting back completes the proof.

\subsection{Proof of Lemma~\ref{lemma:trivial-bound-on-r-f}}
\label{sec:proof-of-trivial-bound-on-r-f}
Plug in the test function $f=0$, we get 
\begin{equation*}
	\E[Y^2] = \mathcal{I}(0, \beta, \lambda) \ge \mathcal{J}(\beta, \lambda).
\end{equation*} 
By definition, $\mathcal{J}(\beta, \lambda) = \E[r_{\beta, \lambda}(X, Y)^2] + \lambda \norm{f_{\beta, \lambda}}_H^2$. We note that equality holds only if and only if the unique minimizer $f_{\beta, \lambda} \equiv 0$. By Lemma~\ref{lemma:euler-lagrange-identity}, this happens if and only if $\E[Yg(\beta\circ X)] =0$ for all $g \in H$. The universality of $H$ (Lemma~\ref{lemma:denseness-of-H-in-L-2}) then implies that this holds if and only if $\E[Y | \beta \circ X] = 0$.

\subsection{Proof of Lemma~\ref{lemma:r-r'-K-bound}}
\label{sec:proof-of-lemma-r-r'-K-bound}
By Lemma~\ref{lemma:euler-lagrange-identity}, for any $z \in \R^d$, we have
\begin{equation*}
	\E[r_{\beta, \lambda}(X, Y) \kernel(\beta \circ X, z)] = \lambda f_{\beta, \lambda}(z).
\end{equation*}
Setting $z = \beta \circ X'$ for an independent copy $(X', Y')$ in Lemma~\ref{lemma:euler-lagrange-identity}, we obtain
\begin{equation*}
	\E[r_{\beta, \lambda}(X, Y) \kernel(\beta \circ X, \beta \circ X')|X'] = \lambda f_{\beta, \lambda}(\beta \circ X').
\end{equation*}
Multiplying both sides by $r_{\beta, \lambda}(X', Y')$ and taking expectation gives
\begin{equation*}
\begin{split} 
	&\E\left[r_{\beta, \lambda}(X, Y) r_{\beta, \lambda}(X', Y') \kernel(\beta \circ X, \beta \circ X')\right]
		= \lambda \E[ r_{\beta, \lambda}(X', Y') f_{\beta, \lambda}(\beta \circ X') ].
\end{split}
\end{equation*}
Finally, applying Lemma~\ref{lemma:euler-lagrange-identity} with $g = f_{\beta, \lambda}$ gives
\begin{equation*}
	\E[r_{\beta, \lambda}(X, Y) f_{\beta, \lambda}(\beta \circ X)] = \lambda \norm{f_{\beta, \lambda}}_H^2.
\end{equation*}
Combining the last two identities establishes the desired relation.

\subsection{Proof of Lemma~\ref{lemma:continuity-of-J}}
\label{sec:proof-of-lemma-continuity-of-J}
By Lemma~\ref{lemma:euler-lagrange-identity}, we get 
\begin{equation*}
	\E[r_{\beta, \lambda}(X, Y) \kernel(\beta \circ X, \beta \circ X')] = \lambda f_{\beta, \lambda}(\beta \circ X').
\end{equation*}
The same identity holds by replacing $\beta$ by $\beta'$. Let 
$\kernel_\beta(x, x') = \kernel(\beta \circ x, \beta \circ x')$. Thus, the difference 
$\Delta_{\beta, \beta'}(x) = f_{\beta, \lambda}(\beta \odot x) - f_{\beta', \lambda}(\beta' \circ x)$ satisfies 
\begin{equation*}
	\E[r_{\beta, \lambda}(X, Y) \cdot (\kernel_\beta - \kernel_{\beta'})(X,X')] -\E[\Delta_{\beta, \beta'}(X) \kernel_{\beta'}(X, X')] = \lambda \Delta_{\beta, \beta'}(X')
\end{equation*} 
We multiply both sides by $\Delta_{\beta, \beta'}(X')$ and take expectation. Note that 
\begin{equation*}
\begin{split}
	\E[\Delta_{\beta, \beta'}(X) \Delta_{\beta, \beta'}(X') \kernel_\beta(X, X')] &
		= \E[\Delta_{\beta, \beta'}(X)\Delta_{\beta, \beta'}(X') \int e^{-2\pi i\langle \beta \circ (X- X')\rangle}
		k(\omega) d\omega] \\
		&= \int \left|\E[\Delta_{\beta, \beta'}(X) e^{-2\pi i \langle \beta \circ X\rangle}]\right|^2 k(\omega) d\omega \ge 0.
\end{split} 
\end{equation*}
We thus deduce the inequality
\begin{equation*}
	\E[r_{\beta, \lambda}(X, Y) \cdot (\kernel_\beta - \kernel_{\beta'})(X,X') \cdot \Delta_{\beta, \beta'}(X')] \ge \lambda \E[\Delta_{\beta, \beta'}(X)^2].
\end{equation*}
Applyig Cauchy-Schwartz to the left-hand-side yields
\begin{equation*}
	\lambda \E[\Delta_{\beta, \beta'}(X)^2] \le \sqrt{\E[r_{\beta, \lambda}(X, Y)^2 \Delta_{\beta, \beta'}(X')^2] \cdot \E[(\kernel_\beta - \kernel_{\beta'})^2(X,X')]}.
\end{equation*}
By independence of $(X, Y)$ and $(X', Y')$, we may further simplify the first factor:
\begin{equation*}
\E[r_{\beta, \lambda}(X, Y)^2 \cdot \Delta_{\beta, \beta'}(X')^2]
= \E[r_{\beta, \lambda}(X, Y)^2] \cdot \E[\Delta_{\beta, \beta'}(X)^2].
\end{equation*}
Note $\E[r_{\beta, \lambda}(X, Y)^2]\le \E[Y^2]$ by Lemma~\ref{lemma:trivial-bound-on-r-f}. Substituting these bounds back gives
\begin{equation*}
\E[\Delta_{\beta, \beta'}(X)^2]
\le \frac{1}{\lambda^2} \cdot \E[Y^2] \cdot \E[(\kernel_\beta - \kernel_{\beta'})^2(X,X')] .
\end{equation*}
By the Lebesgue's dominated convergence theorem, we get $\E[(\kernel_\beta - \kernel_{\beta'})^2(X,X')] \to 0$ 
as $\beta' \to \beta$. This implies $\E[\Delta^2_{\beta, \beta'}(X)] \to 0$ as $\beta' \to \beta$. In other words, we have shown 
\begin{equation*}
	\lim_{\beta' \to \beta} \E[(f_{\beta', \lambda}( \beta' \circ X) - f_{\beta, \lambda} (\beta \circ X))^2] = 0
\end{equation*}
as desired. 

We then prove $\mathcal{J}$ is continuous in $\beta$.  By definition,
\begin{equation*}
	\mathcal{J}(\beta, \lambda) = \E[r_{\beta, \lambda}(X, Y)^2] + \lambda \norm{f_{\beta, \lambda}}_H^2.
\end{equation*}
Applying Lemma~\ref{lemma:euler-lagrange-identity} with $g = f_{\beta, \lambda}$, we get 
$\lambda \norm{f_{\beta, \lambda}}_H^2 = \E[r_{\beta, \lambda}(X, Y) f_{\beta, \lambda}(X)]$. Substituting this into the 
above expression yields
\begin{equation*}
	\mathcal{J}(\beta, \lambda) = \E[r_{\beta, \lambda}(X, Y)^2] + \E[r_{\beta, \lambda}(X, Y) f_{\beta, \lambda}(X)]
		= \E[Y^2] - \E[Y f_{\beta, \lambda}(\beta \circ X)].
\end{equation*}
Since $f_{\beta, \lambda}(\beta \circ X)$ is continuous in $\beta$ in $L^2(\P)$ and $Y \in L^2(\P)$, the map $\beta \mapsto \mathcal{J}(\beta, \lambda)$ is continuous.

\subsection{Proof of Theorem~\ref{theorem:first-variation-formula}}
\label{sec:proof-of-theorem-first-variation-formula}
\redblock{\subsubsection{Proof Architecture} 
\label{sec:proof-architecture}
Let $L^2(\P_X):= \{g: \R^d \to \R \mid \E[g(X)^2] < \infty\}$. Define
\begin{equation*}
	\Phi_\beta: L^2(\P_X) \mapsto C(\R^d),~~~~ g \mapsto \E[g(X) \kernel(\beta \circ X, \cdot)].
\end{equation*}
Define $g_{\beta, \lambda}(x) := \frac{1}{\lambda} \E[r_{\beta, \lambda}(X, Y) |X = x]$. Let 
$\mathfrak{L}(g, \beta, \lambda) := \mathcal{I}(\Phi_\beta(g), \beta, \lambda)$.  The proof has three steps. 

\vspace{.8em}
\noindent\noindent
\textbf{Step 1.}
In Section~\ref{sec:representation-of-the-minimizer-and-the-minimum-value}, we 
first show that the objective function admits the representation
\begin{equation}
\label{eqn:representer-theorem-population-level}
\mathcal{J}(\beta,\lambda)
= \min_{g \in L^2(\P_X)} \mathfrak{L}(g, \beta, \lambda)
= \mathfrak{L}(g_{\beta, \lambda}, \beta, \lambda).
\end{equation} 
This can be viewed as a population-level version of the representer theorem for kernel ridge regression, which is commonly formulated in the finite-sample setting in the machine learning literature~\cite{Bach24}.

\vspace{.8em}
\noindent\noindent
\textbf{Step 2.}
Next, we denote the directional derivative
$\mathrm{D} \mathfrak{L}(g, \beta, \lambda)[v] := \lim_{s \to 0^+} \frac{1}{s}
	(\mathfrak{L}(g, \beta + s v, \lambda) - \mathfrak{L}(g, \beta, \lambda))$ whenever the limit exists. 
In Section~\ref{sec:the-functional-mathfrak-L-and-its-differentiability}, we prove the identity 
\begin{equation}	
\label{eqn:direction-derivative-explicit-formula}
	\mathrm{D} \mathfrak{L}(g_{\beta, \lambda}, \beta, \lambda)[v] = 
		- \frac{1}{\lambda} \E[r_{\beta, \lambda}(X, Y) r_{\beta, \lambda}(X', Y') \mathrm{D} \kernel_\beta(X, X')[v]]. 
\end{equation}

\vspace{.8em}
\noindent\noindent
\textbf{Step 3.}
In Section~\ref{sec:danskin-type-formula}, 
we use the variational characterization of $\mathcal{J}$ in~\eqref{eqn:representer-theorem-population-level} to derive the following formula in the spirit of Danskin's theorem~\cite{Danskin67}:
\begin{equation}
\label{eqn:Danskin-type-formula}
\mathrm{D}\mathcal{J}(\beta,\lambda)[v] 
= \mathrm{D} \mathfrak{L}(g_{\beta, \lambda}, \beta, \lambda)[v].
\end{equation}
While this identity resembles the classical Danskin theorem~\cite{Danskin67}, it does not follow directly from the standard formulations, since the minimization is over the infinite-dimensional space $L^2(\P_X)$.

Substituting the explicit expression from~\eqref{eqn:direction-derivative-explicit-formula} to the right-hand-side 
of~\eqref{eqn:Danskin-type-formula}
yields Theorem~\ref{theorem:first-variation-formula}.
}

\subsubsection{Representation of the Minimizer and the Minimum Value}
\label{sec:representation-of-the-minimizer-and-the-minimum-value} 

We start with a basic property of $\Phi_\beta$, whose definition is given in Section~\ref{sec:proof-architecture}.
\vspace{.5em} 

\begin{lemma}
\label{lemma:Phi-beta-map-into-H}
For every $g\in L^2(\P_X)$, the function $\Phi_\beta (g) \in H$, and its RKHS norm is given by 
\begin{equation*}
	\norm{\Phi_\beta(g)}_H^2 = \E[g(X) g(X') \kernel_\beta(X, X')].
\end{equation*} 
In particular, $\Phi_\beta$ maps into $H$.
\end{lemma}
\begin{proof} 
By direct computation, the Fourier transform of $\Phi_\beta(g)$ is given by
\begin{equation*}
	\widehat{\Phi_\beta}(g) = k(\omega) \cdot \E[g(X) e^{-2\pi i \langle \omega, \beta \circ X\rangle}].
\end{equation*}
By definition of the RKHS norm, we have 
\begin{equation*}
\begin{split} 
	\norm{\Phi_\beta(g)}_H^2 &= \int \frac{|\widehat{\Phi_\beta}(g)|^2(\omega)}{k(\omega)} d \omega
		= \int |\E[g(X) e^{-2\pi i \langle \omega, \beta \circ X\rangle}]|^2 k(\omega) d\omega.
\end{split} 
\end{equation*}
Expanding the squared modulus as an expectation over independent copies $(X, X')$, we obtain
\begin{equation*}
	\norm{\Phi_\beta(g)}_H^2 = 
		\int \E[g(X)g(X') e^{-2\pi i \langle \omega, \beta \circ (X-X')\rangle}] k(\omega) d\omega.
\end{equation*}
Finally, since $\kernel$ and $k$ form a Fourier transform pair, we conclude
\begin{equation*}
	\norm{\Phi_\beta(g)}_H^2 =  \E[g(X) g(X')  \kernel(\beta \circ X, \beta \circ X')].
\end{equation*}
\end{proof} 

Recall from Section~\ref{sec:proof-architecture} that
$g_{\beta, \lambda}(x) = \frac{1}{\lambda} \E[r_{\beta, \lambda}(X, Y) \mid X = x]$,
and 
$\mathfrak{L}(g, \beta, \lambda) = \mathcal{I}(\Phi_\beta(g), \beta, \lambda)$. 
Lemma~\ref{lemma:population-level-representer-theorem} can be viewed as a population-level version of 
the representer theorem~\cite{Bach24}. 

\vspace{1em} 
\begin{lemma}
\label{lemma:population-level-representer-theorem}
At every $\beta$, the minimizer $f_{\beta, \lambda}$ admits the representation 
\begin{equation*}
	f_{\beta, \lambda} = \Phi_\beta(g_{\beta, \lambda}).
\end{equation*}
Additionally, 
\begin{equation*} 
	\mathcal{J}(\beta, \lambda) = \mathfrak{L}(g_{\beta, \lambda}, \beta, \lambda)
		= \min_{g: g \in L^2(\P_X)} \mathfrak{L}(g, \beta, \lambda).
\end{equation*}
\end{lemma} 
\begin{proof} 
Note that 
$g_{\beta, \lambda} \in L_2(\P_X)$ since $\E[g_{\beta, \lambda}(X)^2] \le \E[Y^2] < \infty$.
The representation $f_{\beta, \lambda} = \Phi_\beta(g_{\beta, \lambda})$ follows from Lemma~\ref{lemma:euler-lagrange-identity}. 

For the second identity, first observe that
\begin{equation*}
    \mathcal{J}(\beta, \lambda) = \mathcal{I}(f_{\beta, \lambda}, \beta, \lambda) = \mathcal{I}(\Phi_\beta(g_{\beta, \lambda}), \beta, \lambda).
\end{equation*}
Next, observe that for any $g \in L^2(\P_X)$, we have $\Phi_\beta(g) \in H$ by Lemma~\ref{lemma:Phi-beta-map-into-H}, so
\begin{equation*}
    \mathcal{I}(\Phi_\beta(g), \beta, \lambda) \ge \inf_{f \in H} \mathcal{I}(f, \beta, \lambda) = \mathcal{J}(\beta, \lambda).
\end{equation*}
Hence, $\mathcal{J}(\beta, \lambda) = \min_{g \in L^2(\P_X)} \mathcal{I}(\Phi_\beta(g), \beta, \lambda)$, with equality attained at \( g = g_{\beta, \lambda} \). The proof is then complete, since $\mathfrak{L}(g, \beta, \lambda) = \mathcal{I}(\Phi_\beta(g), \beta, \lambda)$ by definition. 
\end{proof} 

\subsubsection{The Directional Derivative of $\mathfrak{L}$}
\label{sec:the-functional-mathfrak-L-and-its-differentiability}
By Lemma~\ref{lemma:Phi-beta-map-into-H}, we have 
\begin{equation*}
\begin{split} 
	\mathfrak{L}(g, \beta, \lambda) &= 
		\E[(Y- \Phi_\beta(g)(\beta \circ X))^2] + \lambda \norm{\Phi_\beta(g)}_H^2 \\
		&= 
	\E[(Y- \E[g(X') \kernel_\beta(X, X')|X])^2] + \lambda \E[g(X) g(X') \kernel_\beta(X, X')].
\end{split} 
\end{equation*}
Recall that 
$\mathrm{D} \mathfrak{L}(g, \beta, \lambda)[v] = \lim_{s \to 0^+} \frac{1}{s}
	(\mathfrak{L}(g, \beta + s v, \lambda) - \mathfrak{L}(g, \beta, \lambda))$ whenever the limit exists. 
The following two lemmas provide basic properties of the directional derivative $\mathrm{D} \mathfrak{L}(g, \beta, \lambda)[v]$.

\vspace{1em} 
\begin{lemma}
\label{lemma:identity-for-V}
Let $\beta \in \R^d$ and $v \in \R^d$ satisfy the assumptions of Theorem~\ref{theorem:first-variation-formula}. 
For $g\in L^2(\P_X)$, we have
\begin{equation*}
	\mathrm{D} \mathfrak{L}(g, \beta, \lambda)[v] = 
		-2\E[(Y- \Phi_\beta(g)(\beta \circ X))g(X') \mathrm{D}\kernel_\beta(X, X')[v]]
			+ \lambda \E[g(X) g(X') \mathrm{D}\kernel_\beta(X, X')[v]].
\end{equation*}
In particular, 
\begin{equation*}	
	\mathrm{D} \mathfrak{L}(g_{\beta, \lambda}, \beta, \lambda)[v] = 
		- \frac{1}{\lambda} \E[r_{\beta, \lambda}(X, Y) r_{\beta, \lambda}(X', Y') \mathrm{D} \kernel_\beta(X, X')[v]]. 
\end{equation*}
\end{lemma}

\begin{proof}
Under the stated assumptions, the map
$\beta \mapsto \kernel_{\beta}(X, X')$ is differentiable along $v$ at every point $\beta + tv$, $t \in [0,\eps)$,
with directional derivative $\mathrm{D}\kernel_{\beta + t v}(X, X')[v]$ continuous in $t$ satisfying
\[
  \E\Big[\sup_{t \in [0, \eps)} 
     \big(\mathrm{D} \kernel_{\beta + t v}(X, X')[v]\big)^2\Big] < \infty.
\]
Therefore, Lebesgue's dominated convergence theorem allows us to interchange 
differentiation with respect to $\beta$ (along the direction $v$) and expectation in the definition of 
$\mathfrak{L}(g, \beta, \lambda)$. Differentiating the terms in 
$\mathfrak{L}(g, \beta, \lambda)$ then yields the first identity.

We specialize $g = g_{\beta, \lambda}$ in the first identity. Since
$f_{\beta, \lambda} = \Phi_\beta(g_{\beta, \lambda})$, we get that 
\begin{equation*}
	r_{\beta, \lambda}(x, y) = y - \Phi_\beta(g_{\beta, \lambda})(\beta \circ x).
\end{equation*}
Substituting this expression into the formula  
and simplifying the two terms yields the second identity. 
\end{proof} 

\vspace{.5em} 
\begin{lemma}
\label{lemma:continuity-of-V}
Let $\beta \in \R^d$ and $v \in \R^d$ satisfy the assumptions of Theorem~\ref{theorem:first-variation-formula}. Then
the mapping 
\begin{equation*}
	(t_1, t_2) \mapsto \mathrm{D} \mathfrak{L}(g_{\beta + t_1 v, \lambda}, \beta + t_2 v, \lambda)[v]
\end{equation*}  
is continuous on $[0, \eps) \times [0, \eps)$.
\end{lemma}

\begin{proof}
Applying Lemma~\ref{lemma:identity-for-V} gives
\begin{equation*}
\begin{split} 
	&\mathrm{D} \mathfrak{L}(g_{\beta + t_1 v, \lambda}, \beta + t_2 v, \lambda)[v] \\ &=
		-2\E[(Y-\E[g_{\beta + t_1 v, \lambda}(X') \kernel_{\beta + t_2 v} (X, X')|X]) g_{\beta + t_1 v, \lambda} (X') \mathrm{D} \kernel_{\beta + t_2 v}(X, X')[v]] \\
			&~~~~~~~+ \lambda \E[g_{\beta + t_1 v, \lambda}(X) g_{\beta + t_1 v, \lambda}(X')
				\mathrm{D} \kernel_{\beta + t_2 v}(X, X')[v]].
\end{split} 
\end{equation*}
Next we check continuity of the terms. 
By Lemma~\ref{lemma:continuity-of-J}, $r_{\beta, \lambda}(X, Y)$ is continuous in $\beta$ with respect to 
the $L^2(\P)$ norm. As a result, $g_{\beta + t v, \lambda}(X)$ is continuous in $t$ with respect to the $L^2(\P)$-norm. 
In addition, our assumptions 
on the kernel function $\kernel$ ensure that $\mathrm{D} \kernel_{\beta + tv}(X, X')[v]$ and 
$\kernel_{\beta + tv}(X, X')$ depend continuously on $t \in [0, \eps)$ in the $L^2(\P)$-norm.
Following  Lebesgue's dominated convergence theorem, one shows the map 
$(t_1, t_2) \mapsto \mathrm{D} \mathfrak{L}(g_{\beta + t_1 v, \lambda}, \beta + t_2 v, \lambda)[v]$
is continuous on $[0, \eps) \times [0, \eps)$.
\end{proof} 

\subsubsection{A Formula for $\mathcal{J}$}
\label{sec:danskin-type-formula}
Lemma~\ref{lemma:danskin-type-lemma} provides a formula in the spirit of Danskin's theorem~\cite{Danskin67}, showing that the directional derivative of $\mathcal{J}$ can be computed by differentiating $\mathfrak{L}$ at the minimizer $g_{\beta,\lambda}$.

\vspace{1em} 
\begin{lemma}
\label{lemma:danskin-type-lemma}
Let $\beta \in \R^d$ and $v \in \R^d$ satisfy the assumptions of Theorem~\ref{theorem:first-variation-formula}. Then
\begin{equation}
\mathrm{D}\mathcal{J}(\beta,\lambda)[v] 
= \mathrm{D} \mathfrak{L}(g_{\beta, \lambda}, \beta, \lambda)[v].
\end{equation}
\end{lemma} 
\begin{proof} 
Define, for $s > 0$, the difference quotient: 
\begin{equation*}
	\Delta_{s, v} \mathcal{J}(\beta, \lambda):= 
		\frac{1}{s} (\mathcal{J}(\beta + sv, \lambda) - \mathcal{J}(\beta, \lambda)).
\end{equation*}
By the minimizing property of $\mathcal{J}$, we have
$
	\mathcal{J}(\beta, \lambda) \le \mathfrak{L}(g_{\beta + sv, \lambda}, \beta, \lambda)$. Hence, 
\begin{equation*}
\begin{split} 
	\Delta_{s, v} \mathcal{J}(\beta, \lambda) &\ge \frac{1}{s} \left(\mathfrak{L}(g_{\beta+sv, \lambda}, \beta+sv, \lambda)
		- \mathfrak{L}(g_{\beta+sv, \lambda}, \beta, \lambda)\right) \\
		&= \frac{1}{s} \int_0^s \mathrm{D}\mathfrak{L}(g_{\beta+sv, \lambda}, \beta + \tilde{s} v, \lambda)[v] \, d\tilde{s}.
\end{split} 
\end{equation*}
By the continuity of
$\mathrm{D}\mathfrak{L}$ established in Lemma~\ref{lemma:continuity-of-V}, we get
\begin{equation*}
	\liminf_{s \to 0^+} \Delta_{s, v} \mathcal{J}(\beta, \lambda)  \ge \mathrm{D}\mathfrak{L}(g_{\beta, \lambda}, \beta, \lambda)[v].
\end{equation*}
Reversing the roles of $\beta$ and $\beta+sv$, we similarly have $\mathcal{J}(\beta+sv, \lambda) \le \mathfrak{L}(g_{\beta, \lambda}, \beta+sv, \lambda)$. Hence,
\begin{equation*}
\begin{split} 
	\Delta_{s, v} \mathcal{J}(\beta, \lambda) &\le \frac{1}{s} \left(\mathfrak{L}(g_{\beta, \lambda}, \beta+sv, \lambda)
		- \mathfrak{L}(g_{\beta, \lambda}, \beta, \lambda)\right).
\end{split} 
\end{equation*}
Taking limits yields 
\begin{equation*}
	\limsup_{s \to 0^+} \Delta_{s, v} \mathcal{J}(\beta, \lambda)  \le \mathrm{D}\mathfrak{L}(g_{\beta, \lambda}, \beta, \lambda)[v].
\end{equation*} 
This shows that $\mathrm{D}\mathcal{J}(\beta, \lambda)[v] =\lim_{s \to 0^+} \Delta_{s, v} \mathcal{J}(\beta, \lambda)$ exists, and satisfies 
\begin{equation*}
	 \lim_{s \to 0^+} \Delta_{s, v} \mathcal{J}(\beta, \lambda)= \mathrm{D}\mathfrak{L}(g_{\beta, \lambda}, \beta, \lambda)[v].
\end{equation*}
\end{proof} 

\subsubsection{Finalizing the Proof of Theorem~\ref{theorem:first-variation-formula}} 
By Lemma~\ref{lemma:identity-for-V} and Lemma~\ref{lemma:danskin-type-lemma}, 
\begin{equation*}
\begin{split} 
	\mathrm{D}\mathcal{J}(\beta,\lambda)[v] 
		&= \mathrm{D} \mathfrak{L}(g_{\beta, \lambda}, \beta, \lambda)[v] = - \frac{1}{\lambda} \E[r_{\beta, \lambda}(X, Y) r_{\beta, \lambda}(X', Y') \mathrm{D} \kernel_\beta(X, X')[v]]. 
\end{split} 
\end{equation*} 
This completes the proof of Theorem~\ref{theorem:first-variation-formula}.

\subsection{Failure of minimizer existence when $X$ is discrete}
\label{sec:discrete-counterexample}
\indent\indent
We illustrate that the continuity assumption on $X$ in Theorem~\ref{theorem:existence-of-global-minimizers} is necessary for the existence of a minimizer of $J(\cdot,\lambda)$.
Let $d=1$ and consider the discrete distribution
\[
\P\big((X,Y)=(+1,+2)\big)
=
\P\big((X,Y)=(-1,-2)\big)
=
1/2.
\]
We consider either the Laplace $K_{{\rm Lap}}(x, x') = e^{-|x-x'|}$ or Gaussian kernel $K_{{\rm Gauss}}(x,x')=e^{-(x-x')^2}$.
A direct calculation yields (e.g., by using the representer theorem~\cite{Bach24})
\[
J_{{\rm Lap}}(\beta,\lambda) = 
\frac{8\lambda}{(1+2\lambda)-e^{-2|\beta|}}, \qquad 
J_{{\rm Gauss}}(\beta,\lambda) = 
\frac{8\lambda}{(1+2\lambda)-e^{-4\beta^2}}.
\]
In both cases, $J(\beta, \lambda)$ is strictly decreasing in $|\beta|$, and therefore does not attain its infimum over $\R$.

\subsection{Proof of Proposition~\ref{proposition:existence-of-minimal-sufficient-feature-set}}
\label{sec:proof-of-proposition-existence-of-minimal-sufficient-feature-set}
We call a subset $S \subset \{1, 2, \ldots, d\}$ a \emph{sufficient feature set} if 
\begin{equation*}
	\E[Y|X] = \E[Y|X_S]. 
\end{equation*}
Among all such sets, we consider those with the smallest cardinality. Let $S'$ denote one such set.

Below we show that $S' \subset S$ for any sufficient feature set $S$. Suppose, for the sake of contradiction, that this is not true for some sufficient feature set $S$. Since $S' \not\subset S$ and, by definition, $S'$ has cardinality less than or equal to that of $S$, it must be that $S' \setminus S \neq \emptyset$ and $S \setminus S' \neq \emptyset$. We can thus write 
\begin{equation*}
	S = A \cup B~~~\text{and}~~~S' = A\cup C,
\end{equation*}
where $A = S \cap S'$, and $B, C \subset \{1, 2, \ldots, d\}$ are nonempty, disjoint from each other and from $A$.
By the definition of sufficiency of $S, S'$, we have:
\begin{equation*}
	\E[Y|X_A, X_B] =  \E[Y|X_A, X_C].
\end{equation*}
\redblock{
Let $f:\mathbb{R}^{|A|+|B|} \to \mathbb{R}$ and $g:\mathbb{R}^{|A|+|C|} \to \mathbb{R}$ be Lebesgue measurable functions such that
$f(X_A, X_B) = \E[Y | X_A, X_B]$, and 
$g(X_A, X_C) = \E[Y | X_A, X_C]$ almost surely under $\P$.  
Then 
\begin{equation*}
	f(X_A,X_B)=g(X_A,X_C)~~\text{a.s under $\P$}.
\end{equation*}
Given that $X$ has positive probability density with respect to Lebesgue measure on $\R^d$, this implies that 
$f(x_A, x_B) = g(x_A, x_C) ~\text{a.e. under the Lebesgue measure}$. 
By a standard measure-theoretic argument (see Lemma~\ref{lemma:measure-theoretic-lemma} below), one can show that
there exists a Lebesgue measurable function 
$h:\mathbb{R}^{|A|} \to \mathbb{R}$ such that 
$
f(x_A,x_B) = g(x_A,x_C) = h(x_A) ~\text{a.e. under the Lebesgue measure}.
$
As a result, $f(X_A, X_B) = g(X_A,X_C) = h(X_A)$ holds almost surely under $\P$. This then implies 
\begin{equation*}
	\E[Y|X_A, X_B] = \E[Y|X_A] = \E[Y|X_A, X_C].
\end{equation*}
Thus, $A$ is itself a sufficient feature set. But $A \subsetneq S'$ by construction, which contradicts the minimality of $S'$. This proves that $S'$ is a subset of every sufficient feature set.}

With this property, we know that 
\[
S' \subset S \quad \text{for every sufficient feature set } S,
\]
so the intersection over all sufficient sets equals $S'$. Since $S'$ is itself sufficient by construction, it satisfies the definition of the minimal sufficient feature set. This proves existence.

\vspace{.5em} 

\redblock{
\begin{lemma}
\label{lemma:measure-theoretic-lemma}
Let $f: \R^{|A|+|B|} \to \R$ and $g: \R^{|A|+|C|} \to \R$ be measurable functions such that 
\begin{equation*}
	f(x_A, x_B) = g(x_A, x_C)~~\text{a.e. $(x_A,x_B,x_C)\in \R^{|A|+|B|+|C|}$}.
\end{equation*}
Then there exists a Lebesgue measurable function $h: \R^{|A|} \to \R$ such that 
\begin{equation*}
	f(x_A, x_B) = g(x_A, x_C) = h(x_A)~~\text{a.e. $(x_A,x_B,x_C)\in \R^{|A|+|B|+|C|}$}.
\end{equation*}
Throughout, all statements involving “a.e.” are understood with respect to the Lebesgue measure.
\end{lemma}
\begin{proof}
The proof exploits the product structure of Lebesgue measure. Define
\[
F(x_A,x_B,x_C):=f(x_A,x_B)-g(x_A,x_C).
\]
Then $F=0$ a.e. By Fubini's theorem, there exists 
$E\subset \R^{|A|}$ of full measure such that for all $x_A\in E$,
\[
F(x_A,x_B,x_C)=0
\quad \text{for a.e. }(x_B,x_C).
\]
Fix $x_A\in E$. Applying Fubini again, we get a full-measure set 
$B_{x_A}\subset \R^{|B|}$ so that for each $x_B\in B_{x_A}$,
\[
f(x_A,x_B)=g(x_A,x_C)
\quad \text{for a.e. }x_C.
\]
Hence if $x_B^1,x_B^2\in B_{x_A}$, then for some $x_C$,
\[
f(x_A,x_B^1)=g(x_A,x_C)=f(x_A,x_B^2).
\]
This implies that, for every $x_A \in E$, $f(x_A,\cdot)$ is constant on $B_{x_A}$, and thus constant a.e. on \(\R^{|B|}\). 
Denote this value by $h(x_A)$, and define $h$ arbitrarily on $\R^{|A|}\setminus E$.
Combining this with $f(x_A,x_B)=g(x_A,x_C)~\text{for a.e. }(x_B,x_C)$, 
we obtain for every $x_A \in E$,
\[
f(x_A,x_B)=g(x_A,x_C)=h(x_A)
\quad \text{for a.e. }(x_B,x_C).
\]
Since $E \subset \R^{|A|}$ has full measure, it follows that
\begin{equation*}
	f(x_A, x_B) = g(x_A, x_C) = h(x_A)~~\text{for a.e. $(x_A,x_B,x_C)$}
\end{equation*}
as desired. 
\end{proof}
}

\end{document}